%% file: main.tex
\RequirePackage{fix-cm}
\documentclass[a4paper,fleqn]{cas-sc}

\usepackage[numbers,sort&compress]{natbib}
\usepackage{fontspec}
\usepackage{amsmath,amssymb,mathtools}
\usepackage{graphicx}
\usepackage{xcolor}
\usepackage{booktabs,longtable,array,calc}
\usepackage{caption}
\usepackage{algorithm}
\usepackage{algpseudocode}
\usepackage{placeins}
\usepackage{enumitem}
\usepackage{titlesec}
\usepackage{microtype}
\usepackage{geometry}
\usepackage{hyperref}
\usepackage{xurl}
\usepackage{chngcntr}
\usepackage{pdflscape}
\usepackage{needspace}

\hypersetup{colorlinks=true,linkcolor=black,citecolor=black,urlcolor=blue,pdfborder={0 0 0}}
\titleformat{\section}{\bfseries\fontsize{12pt}{15pt}\selectfont}{\thesection}{0.65em}{}
\titleformat{\subsection}{\bfseries\fontsize{10.5pt}{14pt}\selectfont}{\thesubsection}{0.65em}{}
\titlespacing*{\section}{0pt}{18pt}{8pt}
\titlespacing*{\subsection}{0pt}{13pt}{5pt}

\floatname{algorithm}{Algorithm}
\algrenewcommand\algorithmicrequire{\textbf{Input:}}
\algrenewcommand\algorithmicensure{\textbf{Output:}}
\algrenewcommand\algorithmicreturn{\textbf{return}}

\providecommand{\real}[1]{#1}

\shorttitle{SALI-FP}
\shortauthors{}
\ExplSyntaxOn
\cs_set:Npn \__first_footerline:
  { \group_begin: \small \sffamily \rmfamily \itshape
    Preprint~submitted~to~Elsevier \group_end: }
\ExplSyntaxOff

\begin{document}
\fontsize{10.5pt}{14pt}\selectfont

\begin{center}
{\fontsize{14pt}{18pt}\selectfont\bfseries Evidence-gated multimodal parsing and vectorization of architectural floor plans\par}
\end{center}
\hypersetup{pdfauthor={Hongxuan Chen, Wenda Wang, Jiachen Lu, Qirui Shen, Zilong Huang, Lei He, Xinyue Dong, Weixin Huang}}
\begin{center}
{\fontsize{10.5pt}{14pt}\selectfont Hongxuan Chen\textsuperscript{\(\dagger\)}, Wenda Wang\textsuperscript{\(\dagger\)}, Jiachen Lu, Qirui Shen, Zilong Huang, Lei He, Xinyue Dong, Weixin Huang\textsuperscript{\(*\)}\par}
\vspace{4pt}
School of Architecture, Tsinghua University\par
\vspace{3pt}
{\fontsize{9pt}{12pt}\selectfont
\textsuperscript{\(\dagger\)} These authors contributed equally.\par
\textsuperscript{*} Corresponding author: Weixin Huang; \href{mailto:3512232835@qq.com}{3512232835@qq.com}\par}
\end{center}
\vspace{6pt}

\noindent\textbf{Abstract}\par
Architectural floor plans remain a high-friction barrier to archive digitization and early design-model preparation because heterogeneous graphics encode spatial semantics and editable geometry together. We introduce SALI-FP, an evidence-gated multimodal pipeline that converts a plan into reviewable semantic maps, objects, vectors, and relation records while constraining local revisions by image evidence. In a full production audit of 11,534 heterogeneous plans, SALI-FP produced structured outputs for every plan, including 752,510 valid polygon-bearing objects. The same output form has supported initial drawing digitization and design-model preparation in practical design work. Public-benchmark calibration is paired with a 30-case matched visual evidence set in Appendix F, where room-scale coverage, openings, oblique boundaries, and circulation continuity can be inspected directly. SALI-FP offers an engineering-oriented interpretation-to-geometry workflow for reviewed CAD/BIM preparation and existing-building information recovery.

\noindent\textbf{Keywords:} Architectural floor plan parsing; Semantic vectorization; Vision-language models; Evidence-gated agents; Large language models; Geometric reconstruction.
\input{highlights.tex}
\vspace{10pt}

\section{Introduction}\label{introduction}

Floor plans encode geometry, connectivity, and function in the same drawing. When editable CAD or building information modeling (BIM) sources are unavailable, raster archives become an important starting point for existing-building documentation, renovation surveys, and operations information. Recovering objects and their coordinates can reduce the need to reconstruct every geometric entity from linework, while retaining the source for dimensional and functional verification.

Rule-based approaches exploit line patterns and drawing conventions \cite{ref01}. CubiCasa\allowbreak 5K links annotated floor plans to multi-task learning \cite{ref02}; Raster-to-Vector predicts junctions and structured geometry \cite{ref03}, while Floor-SP reconstructs room polygons \cite{ref04}. Room-boundary-guided recognition couples room and boundary tasks \cite{ref05}. VectorFloorSeg and self-constructing graph networks incorporate geometric relationships \cite{ref06,ref07}. These methods establish recognition and structural recovery as connected but distinct problems.

Heterogeneous drawings contain oblique boundaries, furniture, dimensions, and occupancy types that differ from training examples. Object detection \cite{ref08}, multi-unit corpora \cite{ref09}, and sparse point representations \cite{ref10} address complementary aspects of this challenge. Beyond assigning pixel labels, geometric model initialization requires identifiable boundaries, recoverable coordinates, and explicit failure states.

\begin{figure}[!htbp]
  \centering
  \includegraphics[width=\textwidth,height=0.82\textheight,keepaspectratio]{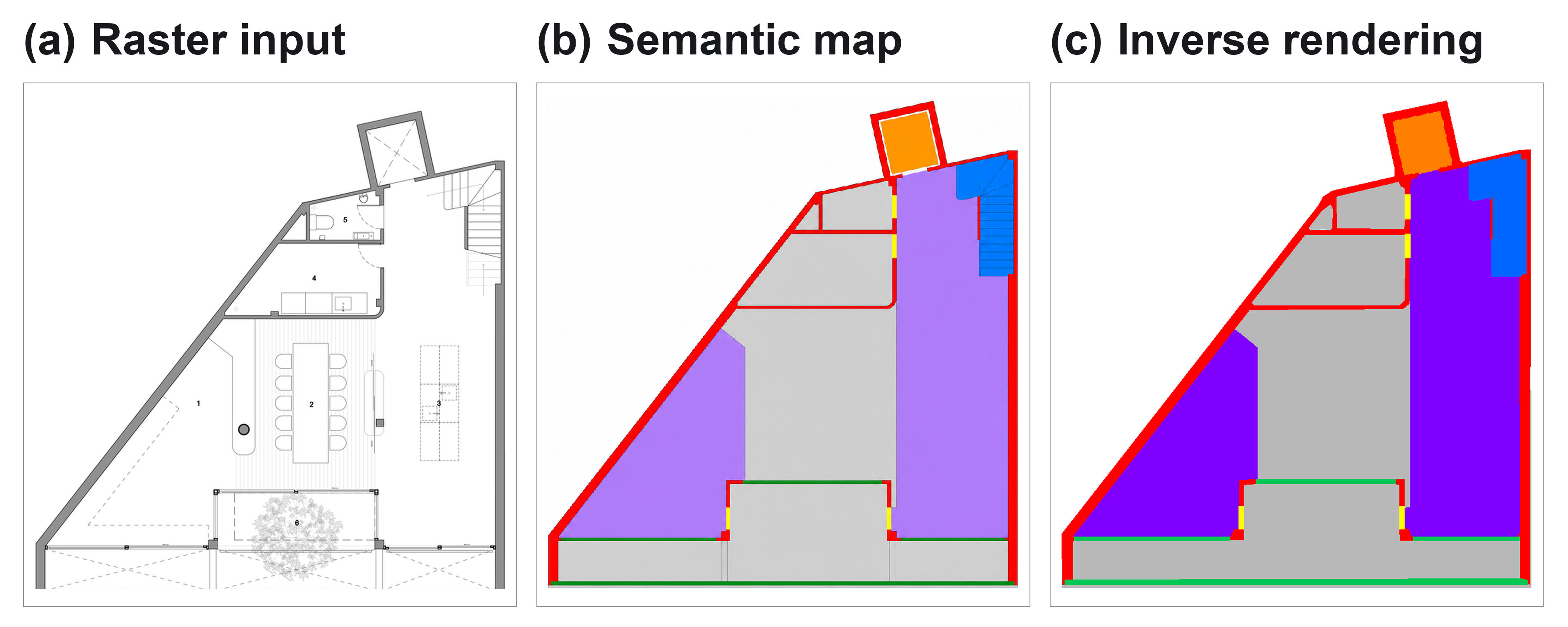}
  \caption{Floor-plan recognition and vectorization: (a) source raster; (b) semantic output; (c) sparse inverse rendering.}
  \label{fig:1}
\end{figure}

We propose SALI-FP as an evidence-gated multimodal parsing method. Its central design separates contextual interpretation from authorization to change the drawing: a model proposes semantic edits, deterministic rules screen the evidence, and a protected state connects accepted changes to structured output. Fig.~\ref{fig:1} illustrates the task.

\FloatBarrier

We contribute an integrated parsing method, a corpus-scale execution study on Architecture Plans 10k (ArchP10k), and a multi-layer protocol linking recognition, geometry, and execution evidence. The evaluation connects shared-annotation calibration with corpus-scale delivery and direct visual inspection of complex drawings. Appendix F provides the latter evidence through 30 fixed, same-coordinate comparisons with localized observations of room coverage, openings, oblique boundaries, and circulation continuity.

Three research questions structure the evaluation: RQ1, how does the method recover semantic and vector representations for reviewed engineering use? RQ2, what recorded changes occur across audit, repair, and representation stages? RQ3, what structured outputs and residual failures emerge on complex heterogeneous plans? Sections 2--4 describe related work, the method, and evaluation settings; Sections 5--7 answer these questions through results, discussion, and conclusions.

\section{Related research}\label{related-research}

\subsection{Semantic segmentation and structural recovery}\label{semantic-segmentation-and-structural-recovery}

Existing methods span rule-based interpretation, supervised segmentation, junction prediction, and vector or graph decoding \cite{ref01,ref02,ref03,ref04,ref05,ref06,ref07}. Pixel losses provide dense supervision, while explicit geometric recovery supplies valid polygon topology. Junction-based methods encode geometric relationships directly, and room-wise reconstruction makes spatial closure central to decoding. Architectural conventions and source-domain distributions influence all of these routes. Recent reviews emphasize the range of datasets and target tasks rather than a single interchangeable definition of floor-plan recognition \cite{ref11}.

\Needspace{9\baselineskip}
\noindent\begin{minipage}{\linewidth}\captionof{table}{Research routes and the evidence required for comparison.}\label{tab:1}\end{minipage}\addtocounter{table}{-1}

\begingroup
\small
\begin{longtable}[]{@{}
  >{\raggedright\arraybackslash}p{(\linewidth - 4\tabcolsep) * \real{0.3333}}
  >{\raggedright\arraybackslash}p{(\linewidth - 4\tabcolsep) * \real{0.3333}}
  >{\raggedright\arraybackslash}p{(\linewidth - 4\tabcolsep) * \real{0.3333}}@{}}
\toprule\noalign{}
\begin{minipage}[b]{\linewidth}\raggedright
Route
\end{minipage} & \begin{minipage}[b]{\linewidth}\raggedright
Main output
\end{minipage} & \begin{minipage}[b]{\linewidth}\raggedright
Comparison boundary
\end{minipage} \\
\midrule\noalign{}
\endhead
\bottomrule\noalign{}
\endlastfoot
Supervised segmentation \cite{ref02,ref05} & Pixel classes & Shared labels, canvas, and test split \\
Junction or polygon recovery \cite{ref03,ref04,ref10} & Structured geometry & Native decoder and topology must be identified \\
Graph-based interpretation \cite{ref06,ref07} & Objects and relations & Align object and relation scope \\
Vision-language understanding \cite{ref12,ref13,ref14,ref15} & Semantic or vector predictions & Evaluate localization explicitly \\
Tool-using agents \cite{ref16,ref17,ref18} & Plans, actions, and state & Report execution and drawing evidence separately \\
\end{longtable}
\endgroup

\subsection{Multimodal evidence and evidence-gated agents}\label{multimodal-evidence-and-evidence-gated-agents}

Text recognition supplies room names and technical labels \cite{ref19}, while multimodal symbol spotting combines heterogeneous evidence \cite{ref20}. LLM-based semantic layering \cite{ref12} and WAFFLE \cite{ref13} extend floor-plan understanding beyond a fixed pixel taxonomy. Visual language models can identify graphics semantically \cite{ref14}, and FloorplanVLM explores structured vector output \cite{ref15}. SALI-FP uses these capabilities to connect semantic interpretation with localized, coordinate-aware action.

Agent schemas for building analysis separate planning, action, memory, and tools \cite{ref16}. BIM coordination \cite{ref17} and Text2BIM \cite{ref18} show how language-based interaction can drive structured software operations. SALI-FP adapts this division of responsibility to floor-plan parsing: the audit model proposes a localized action, deterministic gates authorize the action, and state records retain its outcome.

\subsection{Datasets and quantitative evaluation}\label{datasets-and-quantitative-evaluation}

CubiCasa\allowbreak 5K provides a reproducible public raster/SVG evaluation setting \cite{ref02}. Multi-unit plans \cite{ref09} and FloorPlanCAD \cite{ref21} cover different building scales or symbol tasks. The present quantitative comparison uses the official CubiCasa\allowbreak 5K test plans, a shared target mapping, and frozen preprocessing so that recognition and representation evidence can be read on common coordinates.

\subsection{Geometric interfaces and recent reproducible methods}\label{geometric-interfaces-and-recent-reproducible-methods}

BIM reconstruction from drawings \cite{ref22}, vector-to-energy-model editing \cite{ref23}, and tool-augmented IFC reasoning \cite{ref24} require geometric and semantic attributes beyond a visually coherent mask. MiT-UNet \cite{ref25} provides a recent released wall-segmentation checkpoint, Boundary IoU \cite{ref26} measures contour agreement separately from region overlap, and Raster2Seq \cite{ref27} releases polygon-sequence code and weights. FloorPlanFormer \cite{ref28} contributes a related training formulation. The reproducible rerun set used here combines the released CubiCasa\allowbreak 5K, MiT-UNet, and Raster2Seq implementations.

HEAT uses holistic edge attention for planar graph reconstruction \cite{ref29}, and RoomFormer predicts room polygons with two-level queries \cite{ref30}. Their indoor benchmarks use projected 3D observations, whereas Raster2Seq \cite{ref27} directly models raster-conditioned polygon sequences and is therefore included in our room-geometry comparison. We retain each method's native output form when constructing the shared evidence tables.

Generative polygon refinement offers a complementary route. PolyDiffuse \cite{ref31} uses guided set diffusion to refine polygon proposals; its indoor experiments condition on projected scan evidence rather than architectural raster semantics. Unlike proposal refinement with learned geometric priors, SALI-FP authorizes image-space changes using local evidence gates before constructing polygons.

\section{SALI-FP method}\label{sali-fp-method}

\begin{figure}[p]
  \centering
  \includegraphics[width=\textwidth,height=0.84\textheight,keepaspectratio]{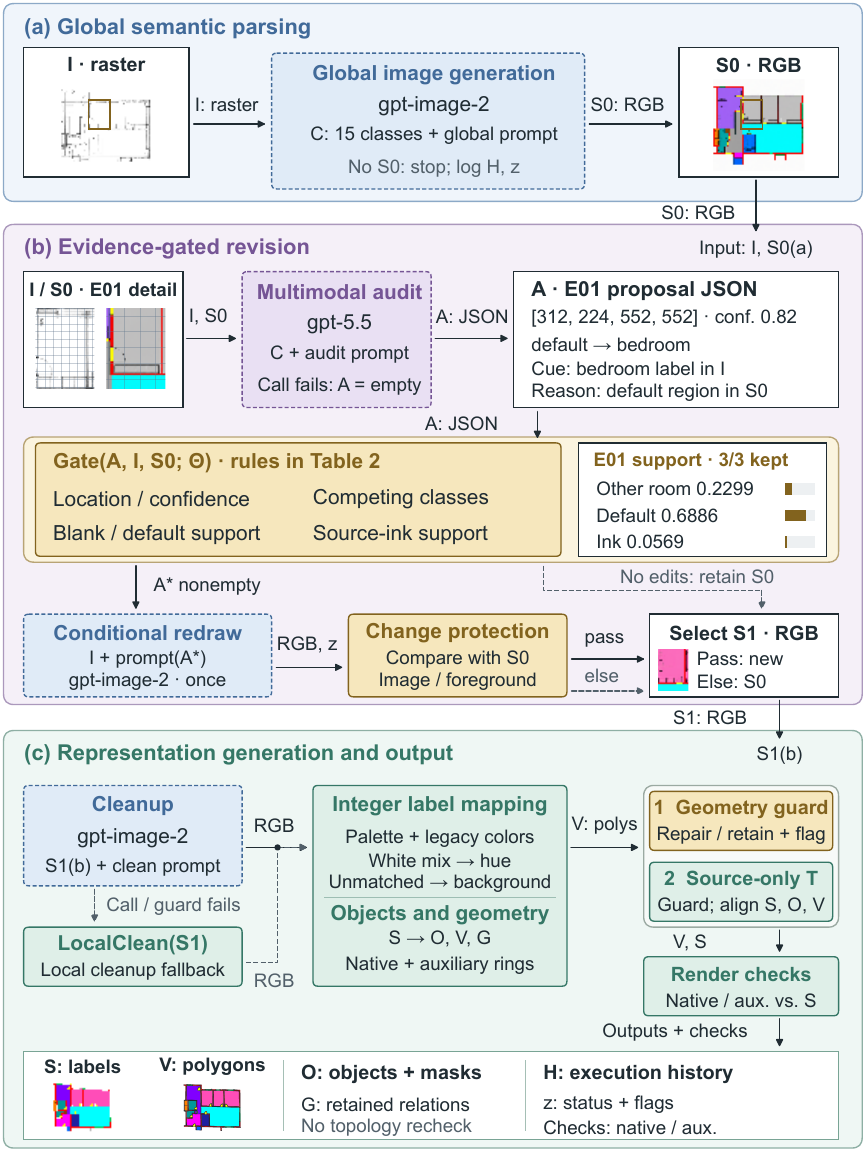}
  \caption{SALI-FP architecture and evidence flow: (a) global semantic parsing; (b) evidence-gated revision and protected candidate selection; (c) representation generation and a connected output package. Rounded containers group operations; rectangular frames contain data and recorded evidence. Arrows identify data or selection conditions. The frozen CubiCasa\allowbreak 5K test 0002 example supplies the images and E01 record; three accepted proposals illustrate the recorded evidence-gated execution. G retains generated relations. The local geometric extension is evaluated as a subsequent representation stage.}
  \label{fig:2}
\end{figure}

SALI-FP integrates global parsing, local audit and gating, optional repair, and representation generation. Models interpret the drawing and propose changes; local operations decide admissibility, preserve a recoverable state, and construct geometric outputs. Fig.~\ref{fig:2} links the complete data flow to the proposal gates, protected state selection, and representation operations.

\clearpage
\clearpage

\subsection{Problem definition and semantic representation}\label{problem-definition-and-semantic-representation}

Let I be the input raster, \(\mathcal C\) the fixed class set, and \(\Theta\) the gate and representation configuration. Eq.~\eqref{eq:1} defines the outputs: integer labels S, objects O, vector geometry V, relation records G, history H, and status z. In G, \(\mathcal R\) contains identified spatial objects and \(E_G\) contains generated relations. The intermediate S0 and S1 are RGB candidate images, S is the integer-label output, and H records execution history and retained states.

\begin{equation}
\mathcal F(I;\mathcal C,\Theta)=(S,O,V,G,H,z),\qquad G=(\mathcal R,E_G).
\label{eq:1}
\end{equation}

The palette contains five component types, nine named spatial types, and one fallback type. RGB assignment first applies thresholded current/legacy palette matching, then white-mixture transition matching, and finally ordered hue rules; unassigned pixels become background. This order resolves anti-aliased and tinted outputs before geometric processing. Appendix A defines labels and Appendix B gives the executed predicates and archived settings.

\subsection{Audit, gated repair, and failure handling}\label{audit-gated-repair-and-failure-handling}

The initial candidate \(S_0\) retains drawing context. The audit model reads I and \(S_0\) with coordinate grids and returns proposals A containing a bounding box or polygon, current and expected classes, source evidence, a described candidate error, repair rationale, and confidence. Gate maps A to \(A^*\) by testing these fields against source pixels and the current labels. A nonempty accepted set triggers one source-conditioned redraw: the image model receives I and a repair prompt assembled from \(A^*\), while \(S_0\) serves as the audit reference and protected comparison. A passed guard selects the redraw as \(S_1\); all other branches retain \(S_0\). This design gives every local action explicit spatial, class, and image-change support before it enters the structured output chain.

\Needspace{9\baselineskip}
\noindent\begin{minipage}{\linewidth}\captionof{table}{Inspected proposal gates and image-change guards.}\label{tab:2}\end{minipage}\addtocounter{table}{-1}

\begingroup
\small
\begin{longtable}[]{@{}
  >{\raggedright\arraybackslash}p{(\linewidth - 2\tabcolsep) * \real{0.5000}}
  >{\raggedright\arraybackslash}p{(\linewidth - 2\tabcolsep) * \real{0.5000}}@{}}
\toprule\noalign{}
\begin{minipage}[b]{\linewidth}\raggedright
Condition
\end{minipage} & \begin{minipage}[b]{\linewidth}\raggedright
Setting and action
\end{minipage} \\
\midrule\noalign{}
\endhead
\bottomrule\noalign{}
\endlastfoot
Confidence / target support & Reject below 0.80 / below 16 target pixels \\
Existing target label & Reject at 0.70 coverage; construction targets at 0.15 \\
Other named room / blank or fallback & Reject at 0.30 other-room coverage / below 0.35 support \\
Source structure & Intensity below 170; require 0.02 dark support for non-door construction \\
Image-change fraction & Repair at most 0.55; cleanup at most 0.35 \\
Cleanup foreground change & Non-white fraction changes by at most 0.15 \\
\end{longtable}
\endgroup

Table~\ref{tab:2} reports the frozen implementation settings that operationalize action authorization. Missing evidence, invalid locations, and incompatible class changes are screened before repair. Appendix B records predicate order, prompt provenance, and the local threshold replay.

SafeCall is the bounded request wrapper. The resume path checks stored success status, output size, and the generation-prompt hash; the evidence evaluations additionally freeze input and code hashes. H records response identifiers, attempts, exceptions, and retained state, while LocalClean applies deterministic palette assignment and local cleanup to the last valid image. Algorithm 1 makes semantic, geometric, and execution states explicit.

\begin{algorithm}[!htbp]
\caption{Evidence-gated multimodal floor-plan parsing}
\label{alg:sali-fp}
\begin{algorithmic}[1]
\Require \parbox[t]{0.88\linewidth}{\raggedright I, class set C, gate configuration \(\Theta\), image and audit models, frozen prompts}
\Ensure \parbox[t]{0.88\linewidth}{\raggedright S, O, V, G, H, z}
\State \parbox[t]{0.88\linewidth}{\raggedright \(S_0\) \(\gets\) SafeCall(image model, I, global prompt)}
\State \parbox[t]{0.88\linewidth}{\raggedright If \(S_0\) is absent: record failure in H, z; return}
\State \parbox[t]{0.88\linewidth}{\raggedright A \(\gets\) SafeCall(audit model, I, \(S_0\), audit prompt)}
\State \parbox[t]{0.88\linewidth}{\raggedright If audit fails: A \(\gets\) empty; record audit degradation}
\State \parbox[t]{0.88\linewidth}{\raggedright \(A^*\) \(\gets\) Gate(A, I, \(S_0\), \(\Theta\)); \(S_1\) \(\gets\) \(S_0\)}
\State \parbox[t]{0.88\linewidth}{\raggedright If \(A^*\) is nonempty:}
\State \parbox[t]{0.88\linewidth}{\raggedright Candidate \(\gets\) SafeCall(image model, I, repair prompt assembled from \(A^*\))}
\State \parbox[t]{0.88\linewidth}{\raggedright If comparison with \(S_0\) passes protection: \(S_1\) \(\gets\) Candidate}
\State \parbox[t]{0.88\linewidth}{\raggedright Else: retain \(S_0\); record protected fallback}
\State \parbox[t]{0.88\linewidth}{\raggedright Candidate \(\gets\) SafeCall(image model, \(S_1\), cleanup prompt)}
\State \parbox[t]{0.88\linewidth}{\raggedright If cleanup fails or violates protection: Candidate \(\gets\) LocalClean(\(S_1\))}
\State \parbox[t]{0.88\linewidth}{\raggedright S \(\gets\) integer labels; extract O, V and relation records G}
\State \parbox[t]{0.88\linewidth}{\raggedright Validate V; accept guarded repair or retain object failure status}
\State \parbox[t]{0.88\linewidth}{\raggedright T \(\gets\) source-only anchoring(I, S); use identity if inadmissible}
\State \parbox[t]{0.88\linewidth}{\raggedright Apply T to S, O masks and V; restore state if geometry protection fails}
\State \parbox[t]{0.88\linewidth}{\raggedright Inverse-render V and auxiliary rings; record their separate RCR values}
\State \parbox[t]{0.88\linewidth}{\raggedright Preserve G as the generated relation record; record H and z}
\State \parbox[t]{0.88\linewidth}{\raggedright Return S, O, V, G, H, z}
\end{algorithmic}
\end{algorithm}
\FloatBarrier

\begin{figure}[!htbp]
  \centering
  \includegraphics[width=\textwidth,height=0.82\textheight,keepaspectratio]{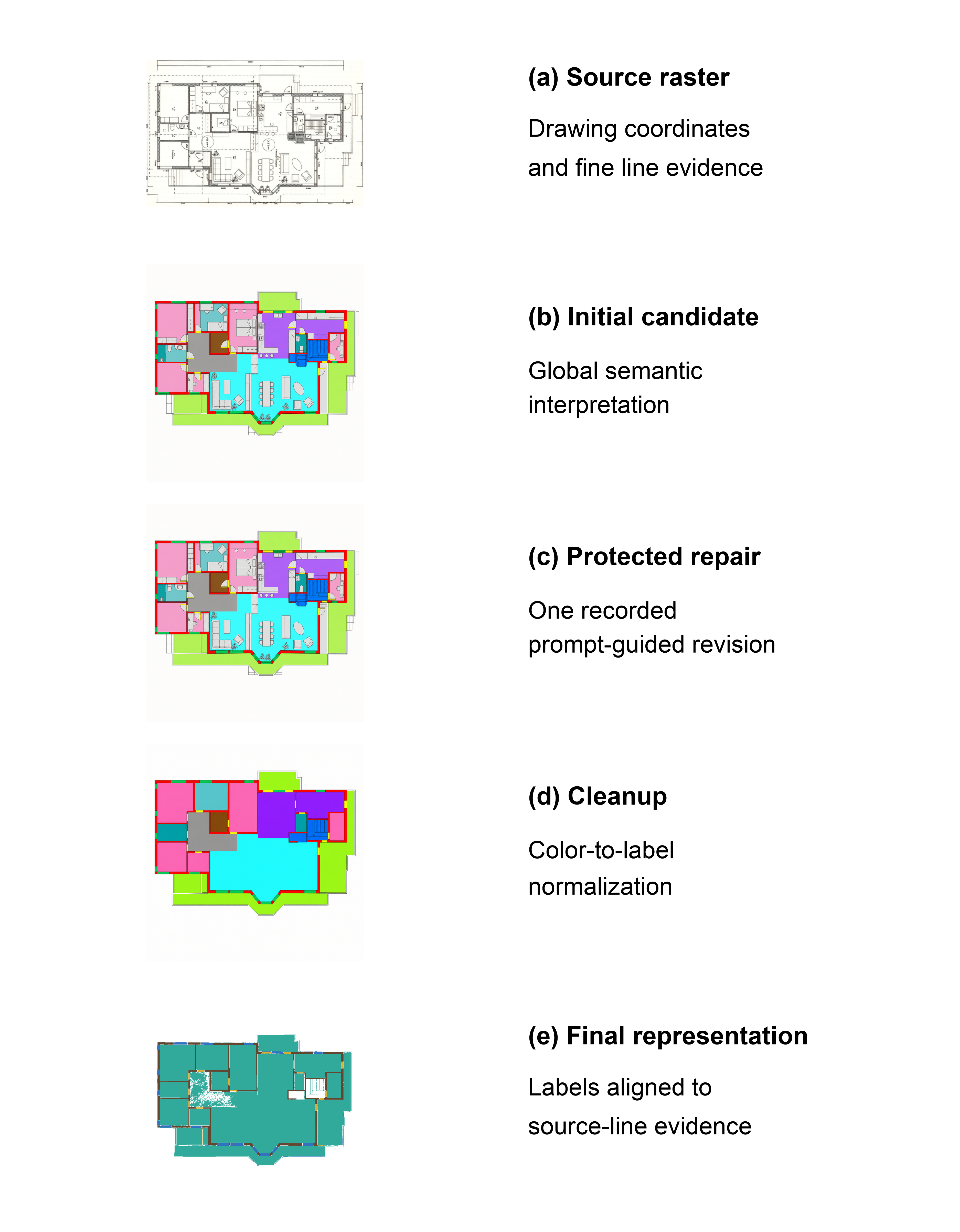}
  \caption{Recorded stages of cubicasa5k\_test\_0001: (a) source; (b) initial candidate; (c) protected repair; (d) cleanup; (e) final source-aligned labels.}
  \label{fig:3}
\end{figure}

The normal path uses two image calls and one audit call; repair adds one image call. Each request allows at most three attempts, transient repair/cleanup failures permit one stage-level re-request, and a case allows at most three end-to-end attempts. Recovery reuses intact stages. Fig.~\ref{fig:3} uses a public test example distinct from Fig.~\ref{fig:1}; its stages are measured jointly in Section 5.3.

\FloatBarrier

\subsection{Geometric output and Rhino interface}\label{geometric-output-and-rhino-interface}

Representation generation uses two complementary geometric layers. Native sparse objects store simplified exterior and hole rings; auxiliary rendering groups use class-specific smoothing and expansion for raster replay. Source-only enhanced correlation coefficient (ECC) registration \cite{ref32} aligns accepted coordinates to drawing line evidence, and linework repair \cite{ref33} admits invalid-object corrections under raster-change and positive-area guards. Appendix B specifies both layers. The retained model outputs precede this subsequently evaluated local extension.

Eq.~\eqref{eq:2} defines internal render-consistency rate (RCR) on the non-background union. We report RCR separately for delivered sparse geometry and auxiliary rendering rings, thereby separating representation agreement from the annotation-aligned recognition and geometry measures reported in Appendix C.

\begin{equation}
U=\{p:S(p)\ne0\lor\widehat S(p)\ne0\},\qquad\mathrm{RCR}=\frac{\sum_{p\in U}\mathbf1[S(p)=\widehat S(p)]}{|U|}.
\label{eq:2}
\end{equation}

\begin{figure}[!htbp]
  \centering
  \includegraphics[width=\textwidth,height=0.82\textheight,keepaspectratio]{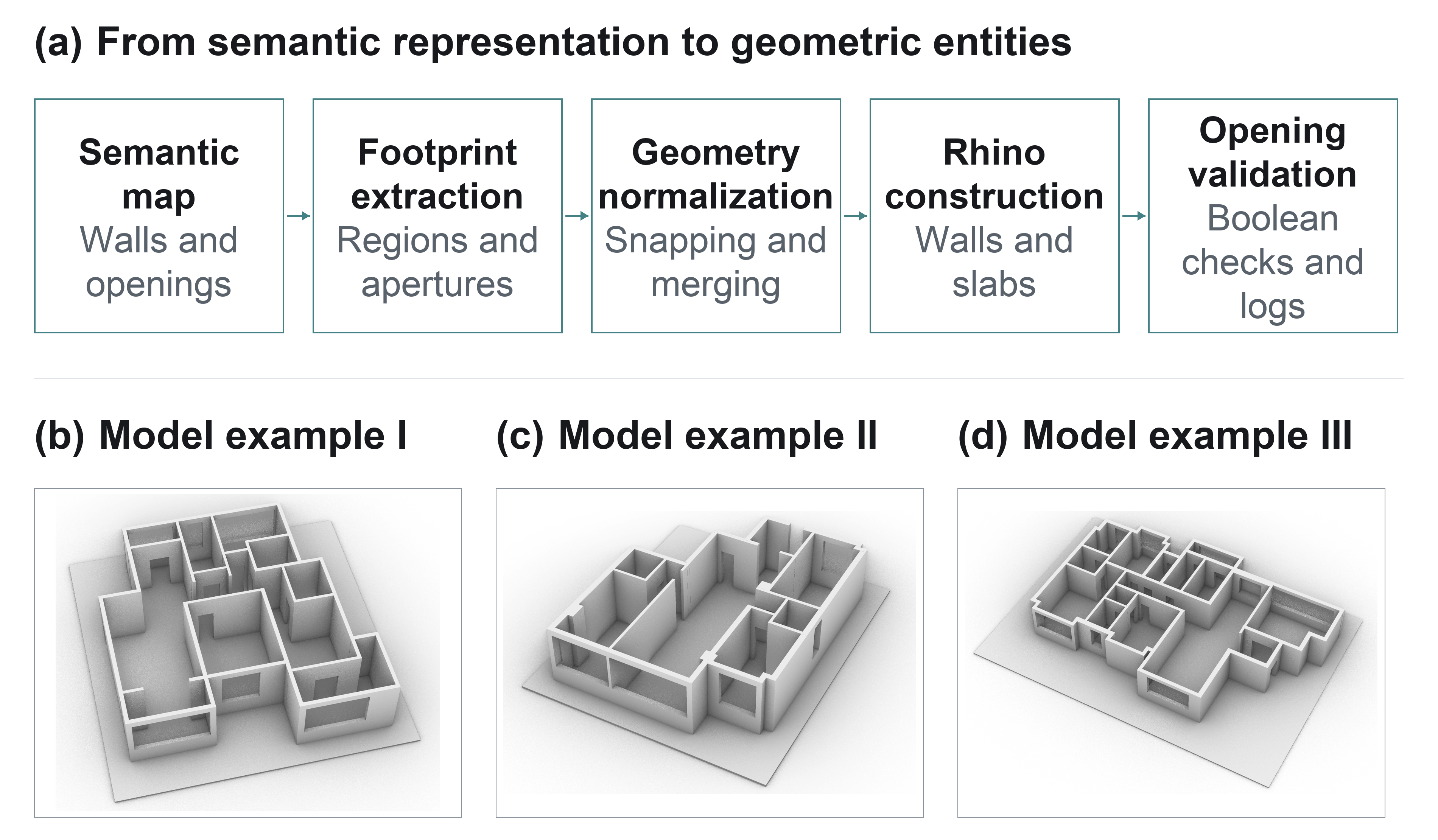}
  \caption{Semantic-to-Rhino interface and retained model illustrations.}
  \label{fig:4}
\end{figure}

Labeled footprints and openings provide inputs for Rhino wall, slab, and opening operations (Fig.~\ref{fig:4}). The illustrated interface establishes the handoff from semantic-vector output to model construction; project workflows then apply their source-coordinate, scale, and extrusion specifications.

\FloatBarrier

\section{Corpus and experimental design}\label{corpus-and-experimental-design}

\subsection{ArchP10k collection and governance}\label{archp10k-collection-and-governance}

We organize web-sourced architectural plans by source project and recorded building category. Exact SHA-256 duplicates and 64-bit perceptual-hash candidates with Hamming distance at most 6 undergo relationship review. Removing 174 duplicates from 11,708 inputs yields 11,534 working images standardized to 1024 × 1024 pixels. Source links, categories, and governance records are retained separately from predictions.

The corpus contains 7,210 residential and 4,324 other plans. Frozen project-level partitions contain 9,225 development, 1,157 validation, and 1,152 test plans; development denotes a study partition, not supervised model training. The hard subset contains the 400 highest-complexity test plans. Table~\ref{tab:3} summarizes the evidence; Appendix A retains source-group counts and predicted semantic composition.

\Needspace{9\baselineskip}
\noindent\begin{minipage}{\linewidth}\captionof{table}{ArchP10k research corpus and evidence.}\label{tab:3}\end{minipage}\addtocounter{table}{-1}

\begingroup
\small
\begin{longtable}[]{@{}
  >{\raggedright\arraybackslash}p{(\linewidth - 2\tabcolsep) * \real{0.5000}}
  >{\raggedright\arraybackslash}p{(\linewidth - 2\tabcolsep) * \real{0.5000}}@{}}
\toprule\noalign{}
\begin{minipage}[b]{\linewidth}\raggedright
Item
\end{minipage} & \begin{minipage}[b]{\linewidth}\raggedright
Record
\end{minipage} \\
\midrule\noalign{}
\endhead
\bottomrule\noalign{}
\endlastfoot
Working plans / format & 11,534 / 1024 × 1024 PNG \\
Development / validation / test & 9,225 / 1,157 / 1,152 \\
Hard subset & 400 within test \\
Independent-reference annotation package & 200 stratified tasks \\
Release governance & Source-specific access, privacy, and redistribution review \\
\end{longtable}
\endgroup

\subsection{Public GT and baseline protocols}\label{public-gt-and-baseline-protocols}

CubiCasa\allowbreak 5K provides 4,200 training, 400 validation, and 400 test plans \cite{ref02}. All 400 test plans enter pixel evaluation on the same aspect-preserving 1024 × 1024 canvas. The shared evaluation targets structural boundary, door/opening, window, and room: walls and railings form the first class, door/window labels override spaces, and SALI-FP stairs and elevators map to background within this common protocol.

We rerun the trained CubiCasa\allowbreak 5K checkpoint with its official four-rotation handling and retain all 44 prediction channels. The native argmax and official polygon postprocessor are evaluated separately. Raster2Seq uses its released CubiCasa\allowbreak 5K checkpoint and native room decoding, MiT-UNet retains its released wall-only configuration, and Table~\ref{tab:4} identifies the output scope of each method.

\Needspace{9\baselineskip}
\noindent\begin{minipage}{\linewidth}\captionof{table}{Official-weight baseline configurations.}\label{tab:4}\end{minipage}\addtocounter{table}{-1}

\begingroup
\small
\begin{longtable}[]{@{}
  >{\raggedright\arraybackslash}p{(\linewidth - 4\tabcolsep) * \real{0.3333}}
  >{\raggedright\arraybackslash}p{(\linewidth - 4\tabcolsep) * \real{0.3333}}
  >{\raggedright\arraybackslash}p{(\linewidth - 4\tabcolsep) * \real{0.3333}}@{}}
\toprule\noalign{}
\begin{minipage}[b]{\linewidth}\raggedright
Method
\end{minipage} & \begin{minipage}[b]{\linewidth}\raggedright
Preprocessing and decoder
\end{minipage} & \begin{minipage}[b]{\linewidth}\raggedright
Evaluation scope
\end{minipage} \\
\midrule\noalign{}
\endhead
\bottomrule\noalign{}
\endlastfoot
CubiCasa\allowbreak 5K \cite{ref02} & Four right-angle rotations; full 44 channels & Four-class argmax; official polygon decoder \\
Raster2Seq \cite{ref27} & 256-pixel bicubic input; released EMA checkpoint & Native room polygons; room, corner, angle matching \\
MiT-UNet \cite{ref25} & 512-pixel native normalization; wall threshold 0.5 & Visible walls only \\
SALI-FP & Retained image/audit outputs; frozen representation stage & Four common classes; stored sparse room geometry \\
\end{longtable}
\endgroup

Raster2Seq inference uses the repository's pure PyTorch deformable-attention reference on CPU, with released weights loaded without training or architecture changes. CubiCasa\allowbreak 5K retains its junction channels for polygon decoding. Code, checkpoints, adapters, and per-case outputs are hashed. Appendix C reports the resulting run conditions separately from published timing settings.

The original 60/40 validation allocation yields 44 calibration and 32 holdout outputs. Registration is selected on calibration, checked on holdout, and then frozen. The resulting comparisons use fixed inputs, transformations, and evaluation records; Appendix C gives the complete calibration and stage analyses.

\subsection{Multi-layer evaluation and statistical analysis}\label{multi-layer-evaluation-and-statistical-analysis}

We separate GT recognition, GT-referenced room geometry, representation cost, and corpus execution. Pixel metrics are pooled mean intersection over union (mIoU), mean class pixel accuracy (PA), overall accuracy (OA), and per-class IoU. Boundary IoU (BIoU) \cite{ref26} averages ten diagonal-relative bands from 0.1\% to 1.0\%; boundary F1 and visible-wall clDice \cite{ref34} provide supplementary contour and centerline measures. Appendix B defines denominators and empty-set handling.

Native room, corner, and angle evaluation uses the Raster2Seq CubiCasa entry Evaluator\_RPlan with room IoU above 0.5, corner tolerance 10 pixels on a 256-pixel canvas, and angle tolerance 5 degrees. GT follows official source-coordinate rounding, closing-token removal, upper-bound clipping, and the 100-pixel source-area cutoff before the frozen common-canvas transform. Every method uses the same disabled-overlap primary setting and enabled-overlap sensitivity setting. Geometry evaluation contains 398 eligible room references, while the complete 400-case record is retained; the 5 failed CubiCasa polygon predictions contribute zero where an eligible reference exists. The common-canvas reruns provide a controlled comparison of the released methods, and angle is reported as its own geometric measure.

For accuracy-cost curves, every common-class raster passes through the same pixel-cell decoder with topology-preserving simplification at 0, 0.25, 0.5, 1, 2, 4, and 8 pixels. Each globally fixed tolerance is evaluated against GT. Appendix B and Tables C.9-C.10 additionally report the own-prediction fidelity measurement for representation analysis.

Public-test intervals use 10,000 paired bias-corrected and accelerated (BCa) bootstrap resamples. Pixel metrics pool confusion counts; native geometry first averages plan precision and recall and then computes their harmonic F1, with mean per-plan F1 reported separately. Every replicate repeats that aggregation. Corpus inference resamples source-project clusters, factor tests use Holm correction, and stage analyses report original coordinates alongside the same frozen final source-only transform applied to every stage. Interquartile ranges (IQRs) describe dispersion.

\subsection{Implementation and reproducibility}\label{implementation-and-reproducibility}

Production records identify gpt-image-2 for image generation, repair, and cleanup, and gpt-5.5 for multimodal audit through an OpenAI-compatible gateway. Image editing takes images and text and returns 1024 × 1024 images; auditing takes two annotated images and text and returns structured suggestions. Production used 16 client workers. Deployment aliases, response identifiers, and frozen local materials establish the operational record; Appendix E documents the reproducibility inventory.

Existing model outputs are retained unchanged. Local representation export uses OpenCV 5.0.0, Shapely 2.1.2, and 8 CPU workers; shared pixel-cell extraction uses Rasterio 1.4.4. The new stage and official-weight comparisons make no image-model or audit-model API calls. Appendix E records exact code and weight hashes, prompt availability, compatible-runtime changes, and unresolved historical configuration gaps.

\section{Results and analysis}\label{results-and-analysis}

\subsection{Benchmark calibration and engineering-oriented evidence}\label{benchmark-calibration-and-engineering-oriented-evidence}

RQ1 is answered through a deliberately linked engineering-evidence path. The CubiCasa\allowbreak 5K protocol provides necessary annotation-aligned calibration on shared coordinates. Fixed-taxonomy pixel and room metrics, however, cannot alone show whether heterogeneous drawings retain coherent spatial organization, editable geometry, and a traceable review path. The corpus audit establishes delivery of semantic, object, vector, relation, and state records at production scale. Appendix F makes the corresponding room-scale organization, opening retention, oblique boundaries, circulation continuity, and residual errors directly inspectable on matched complex plans. Appendix C retains the complete public-test scores, intervals, native-room measures, stage trajectories, and pixel diagnostics.

\begin{figure}[!htbp]
  \centering
  \includegraphics[width=\textwidth,height=0.82\textheight,keepaspectratio]{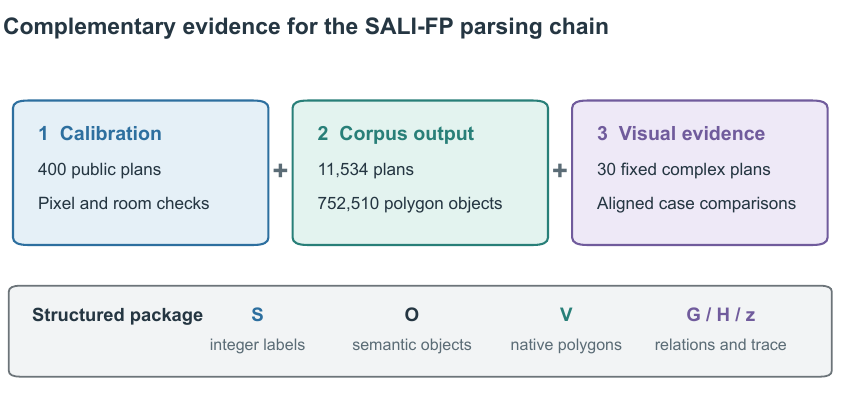}
  \caption{Evidence route for SALI-FP. Strict public-test calibration, full-corpus structured output, and fixed visual comparisons answer complementary parts of the evaluation. Detailed baseline scores and diagnostics are retained in Appendix C; the 30 matched complex-plan comparisons are in Appendix F.}
  \label{fig:5}
\end{figure}

The principal contribution of SALI-FP is a complete, traceable interpretation-to-geometry chain. Table~\ref{tab:5} records its delivered structure across the full ArchP10k corpus; Fig.~\ref{fig:5} shows how controlled calibration, corpus-scale output, and visual structural reading form one engineering-oriented evidence argument.

\Needspace{9\baselineskip}
\noindent\begin{minipage}{\linewidth}\captionof{table}{Corpus-scale structured-output evidence.}\label{tab:5}\end{minipage}\addtocounter{table}{-1}

\begingroup
\small
\begin{longtable}[]{@{}
  >{\raggedright\arraybackslash}p{(\linewidth - 6\tabcolsep) * \real{0.2500}}
  >{\raggedright\arraybackslash}p{(\linewidth - 6\tabcolsep) * \real{0.2500}}
  >{\raggedright\arraybackslash}p{(\linewidth - 6\tabcolsep) * \real{0.2500}}
  >{\raggedright\arraybackslash}p{(\linewidth - 6\tabcolsep) * \real{0.2500}}@{}}
\toprule\noalign{}
\begin{minipage}[b]{\linewidth}\raggedright
Output layer
\end{minipage} & \begin{minipage}[b]{\linewidth}\raggedright
Delivered evidence
\end{minipage} & \begin{minipage}[b]{\linewidth}\raggedright
Scale
\end{minipage} & \begin{minipage}[b]{\linewidth}\raggedright
Engineering interpretation
\end{minipage} \\
\midrule\noalign{}
\endhead
\bottomrule\noalign{}
\endlastfoot
Semantic labels \(S\) & Nonempty normalized semantic map & 11,534 / 11,534 plans & Common input to object and geometry extraction \\
Objects \(O\) & Typed shape records & 783,299 shapes & Explicit semantic units for review \\
Native vectors \(V\) & Polygon-bearing objects passing the delivery audit & 752,510 / 783,299 (96.07\%) & Editable geometry with retained status \\
Relations \(G\) & Generated relation pairs & 1,591,739 pairs & Preserved structural context for downstream inspection \\
Trace \(H,z\) & State and recovery records & 11,534 plans & Recoverable execution and review path \\
\end{longtable}
\endgroup

\FloatBarrier

\begin{figure}[!htbp]
  \centering
  \includegraphics[width=\textwidth,height=0.82\textheight,keepaspectratio]{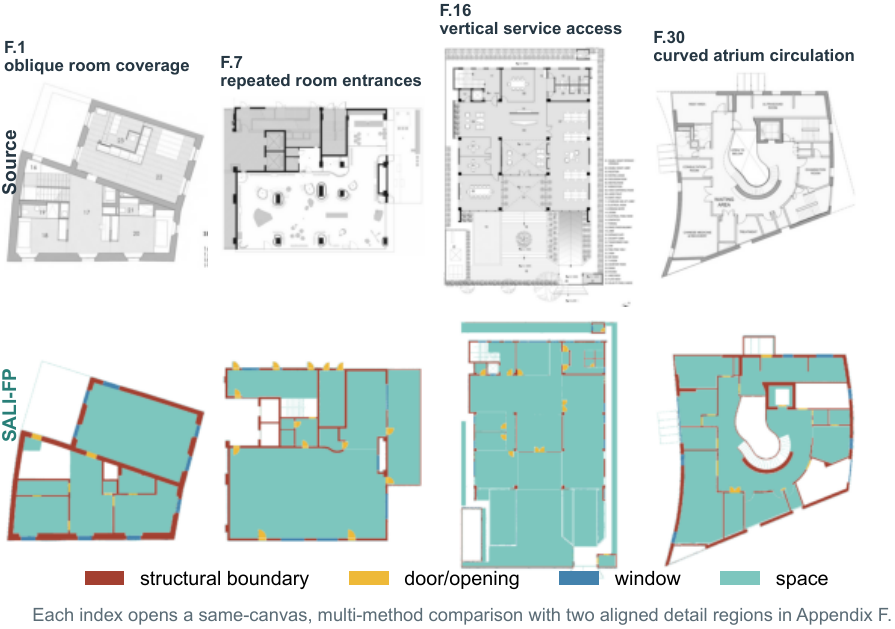}
  \caption{Entry points to the fixed visual-evidence set. Each source/SALI-FP pair corresponds to a full, matched baseline comparison with two aligned detail regions in Appendix F.1, F.7, F.16, and F.30; Appendix F contains all 30 cases.}
  \label{fig:6}
\end{figure}
\FloatBarrier

Appendix F is an integral RQ1 result rather than supplementary illustration and should be inspected alongside the calibration tables. Its 30 fixed complex-plan pages use common canvases, paired local regions, and case-specific observations to expose room-scale coverage, opening retention, nonorthogonal boundaries, circulation continuity, and residual errors. The complete score tables in Appendix C provide the complementary population-level calibration.

\subsection{Full-corpus output and geometric validity}\label{full-corpus-output-and-geometric-validity}

\begin{figure}[!htbp]
  \centering
  \includegraphics[width=\textwidth,height=0.82\textheight,keepaspectratio]{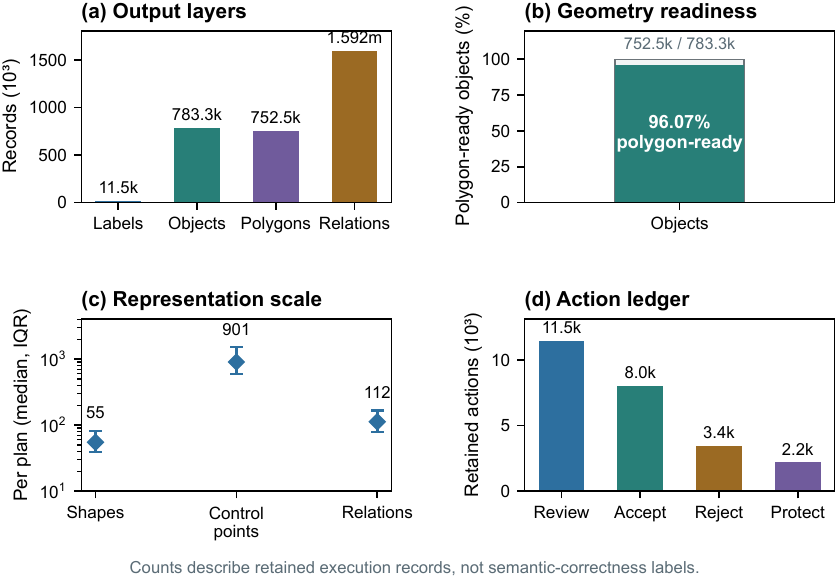}
  \caption{ArchP10k structured-output audit: (a) delivered output layers; (b) polygon-bearing geometry passing the delivery audit; (c) per-plan representation scale; and (d) retained audit-action records. Quantities have different denominators and are presented separately.}
  \label{fig:7}
\end{figure}

All 11,534 ArchP10k plans produce nonempty structured outputs after recovery, answering the coverage component of RQ3. The corpus yields 783,299 shapes, 15,002,422 control points, and 1,591,739 generated relation pairs. Of these objects, 752,510 (96.07\%) carry polygon geometry that passes the delivery audit. Fig.~\ref{fig:7} connects output coverage, geometric readiness, representation scale, and retained execution states across the complete SALI-FP chain.

Residual object identities and geometric causes remain attached to the review record, making the delivered representation directly inspectable. Tables C.18--C.21 provide the full cause taxonomy and category cross-tabs. This traceability is central to SALI-FP: it delivers a structured starting point for review, revision, and later modeling work.

\subsection{Recorded stage changes and gating}\label{recorded-stage-changes-and-gating}

The recorded stages make every intervention path inspectable. Across ArchP10k, 5,273 plans trigger one revision and 6,261 retain the initial candidate; 11,481 audit proposals yield 8,033 accepted actions and 3,448 rejected actions, while 2,181 plans record a protective state or fallback. Appendix Tables C.11, C.16, and Fig. C.7 provide the complete coordinate-conditioned stage analysis, paired intervals, and skipped-output record.

\FloatBarrier

Complexity is associated with a larger representation burden: median control points rise from 651 to 1,288 across quartiles. Project-cluster analysis separates the five factors (Table~\ref{tab:D.1}); hard versus non-hard patterns are retained in Table~\ref{tab:D.2}. The fallback class occupies 16.85\% of predicted foreground and 28.0\% in office/education/research sources, locating functionally unresolved regions for later review.

\subsection{Visual structural evidence and interface observations}\label{visual-structural-evidence-and-interface-observations}

Public error matrices and spatial diagnostics in Appendix Tables C.17 and C.22 and Fig. C.8 locate the calibration targets. Appendix F supplies their case-level engineering reading: every page aligns the source plan, official baselines, SALI-FP output, and two matched local regions, then identifies the visible structural evidence in that case. The resulting comparisons make room-scale completeness, aperture retention, oblique envelopes, and continuous circulation available for direct review.

\section{Discussion and practical implications}\label{discussion-and-practical-implications}

\subsection{Evidence-led method positioning}\label{evidence-led-method-positioning}

SALI-FP is designed for an architectural work unit that combines global reading, admissible local action, preserved state, and structured geometry in one inspectable process. The public benchmark anchors annotation-aligned localization, while the corpus audit and Appendix F demonstrate how the same process retains coherent rooms, openings, boundaries, and usable geometric records across heterogeneous drawings.

Fixed-taxonomy and pixel-alignment measures provide the annotation-aligned calibration reported in Appendix C and remain essential for comparable evaluation. They capture a different evidence layer from spatial organization, geometric editability, and review traceability. SALI-FP unifies these operational attributes with explicit gates, recoverable actions, corpus-scale output, and object-level geometry, thereby supplying the reviewed semantic-vector material required by downstream design workflows.

\subsection{Deployment and reproducibility}\label{deployment-and-reproducibility}

The production audit records 39,875 logical calls and 41,931 HTTP attempts across the full corpus. Retained protocols, prompts where recoverable, input identifiers, output states, and local geometry records support inspection and local replay. The same semantic-object-vector output form has supported initial drawing digitization and design-model preparation in practical design work associated with the Architectural Design and Research Institute of Tsinghua University. This use connects visual interpretation to a reviewable geometry handoff. The resource profile is appropriate for asynchronous archive conversion and design-model preparation, where a reviewer can inspect the resulting semantic and geometric evidence before downstream use; detailed latency and reference-rate records are in Tables C.3 and C.19.

\subsection{Evidence scope and geometric use}\label{evidence-scope-and-geometric-use}

The study combines public annotation-aligned calibration, full-corpus structured-output coverage, and purpose-selected visual comparison. Together, these evidence scopes connect measured recognition, production-scale delivery, and design-relevant interpretation. The frozen 200-plan annotation package and the independent proposal-review package provide the next validation instruments for ArchP10k reference accuracy and decision quality.

For downstream solid construction, the exported status record identifies objects for review, while project-specific source-coordinate, opening, scale, and height specifications complete the model-construction workflow. Appendix E records the retained materials and the planned open-weight, repeatability, and annotation extensions.

\section{Conclusion}\label{conclusion}

We propose SALI-FP as an evidence-gated multimodal method that turns architectural floor-plan interpretation into a reviewable structured-geometry handoff. Its integrated chain combines global multimodal reading, explicit authorization of local revision, preserved execution state, and inspectable semantic-object-vector-relation output.

Across 11,534 heterogeneous ArchP10k plans, SALI-FP produces nonempty structured outputs for every plan and delivers 752,510 valid polygon-bearing objects. Appendix C supplies the complete public benchmark calibration; Appendix F supplies the matched visual evidence needed to inspect the method on complex rooms, openings, oblique envelopes, and circulation structure. Together, these results establish SALI-FP as an engineering-oriented workflow for reviewed architectural archive digitization, existing-building information recovery, and initial design-model preparation.

\section*{Data Availability}
\addcontentsline{toc}{section}{Data Availability}

CubiCasa\allowbreak 5K data and the released baseline code and weights are available from their authors \cite{ref02,ref25,ref27}. The local reproduction package contains protocols, code and weight hashes, prediction identifiers, metric records, recoverable prompts, and annotation templates. Original predictions, derived geometry, and reviewer keys are maintained as distinct materials. ArchP10k follows source-specific access, external-processing, privacy, and redistribution governance; the public package stages reproducible materials while keeping reviewer keys and private source mappings protected.

\section*{Declaration of generative AI and AI-assisted technologies in the writing process}
\addcontentsline{toc}{section}{Declaration of generative AI and AI-assisted technologies in the writing process}

OpenAI Codex image generation (backend version not exposed) was used for architecture design studies on 2026-09-13 and 2026-09-15, including three alternative layouts for Fig.~\ref{fig:2}. The final figure was reconstructed with AI-assisted vector plotting code, independently typeset labels and data paths, and frozen experimental images and proposal records. Generated floor plans, pictograms, and unverified model-proposed connections were not used as scientific evidence.

AI tools assisted language editing, code preparation, and record-based analysis. The authors verified the analyses, approved the manuscript, and take responsibility for the publication disclosures.

\renewcommand{\refname}{References}

\appendix
\counterwithin{table}{section}
\counterwithin{figure}{section}
\counterwithin{equation}{section}
\renewcommand{\thetable}{\thesection.\arabic{table}}
\renewcommand{\thefigure}{\thesection.\arabic{figure}}
\renewcommand{\theequation}{\thesection.\arabic{equation}}
\section{Semantic mappings and corpus composition}\label{app:A}

\Needspace{9\baselineskip}
\noindent\begin{minipage}{\linewidth}\captionof{table}{Semantic palette and common-class display mapping.}\label{tab:A.1}\end{minipage}\addtocounter{table}{-1}

\begingroup
\small
\begin{longtable}[]{@{}lll@{}}
\toprule\noalign{}
Class & RGB & Common target \\
\midrule\noalign{}
\endhead
\bottomrule\noalign{}
\endlastfoot
wall & 255, 0, 0 & Structural boundary \\
door & 255, 255, 0 & Door/opening \\
window & 0, 200, 83 & Window \\
stair & 0, 102, 255 & Background (unmapped) \\
elevator & 255, 128, 0 & Background (unmapped) \\
living\_room & 0, 255, 255 & Room \\
bedroom & 255, 77, 184 & Room \\
kitchen & 128, 0, 255 & Room \\
bathroom & 0, 153, 153 & Room \\
balcony & 153, 255, 0 & Room \\
entrance & 255, 0, 255 & Room \\
storage & 128, 64, 0 & Room \\
corridor & 128, 128, 128 & Room \\
study & 32, 32, 160 & Room \\
room\_default & 184, 184, 184 & Room \\
\end{longtable}
\endgroup

SALI-FP stair and elevator labels have no separate target in the frozen common-class raster mapping and are mapped to background, not silently excluded from the denominator. All other SALI-FP space labels, including the fallback class, map to room. CubiCasa\allowbreak 5K native room types map to room except walls/railings; native door/window icons override the room map. The common target is therefore a disclosed operational mapping rather than a claim that every source ontology is identical.

\Needspace{9\baselineskip}
\noindent\begin{minipage}{\linewidth}\captionof{table}{Source-group composition and predicted fallback-class share.}\label{tab:A.2}\end{minipage}\addtocounter{table}{-1}

\begingroup
\small
\begin{longtable}[]{@{}
  >{\raggedright\arraybackslash}p{(\linewidth - 4\tabcolsep) * \real{0.2727}}
  >{\raggedleft\arraybackslash}p{(\linewidth - 4\tabcolsep) * \real{0.3636}}
  >{\raggedleft\arraybackslash}p{(\linewidth - 4\tabcolsep) * \real{0.3636}}@{}}
\toprule\noalign{}
\begin{minipage}[b]{\linewidth}\raggedright
Source group
\end{minipage} & \begin{minipage}[b]{\linewidth}\raggedleft
Plans
\end{minipage} & \begin{minipage}[b]{\linewidth}\raggedleft
\texttt{room\_default} / predicted non-background pixels
\end{minipage} \\
\midrule\noalign{}
\endhead
\bottomrule\noalign{}
\endlastfoot
Residential & 7,210 & 13.6\% \\
Commercial and hospitality & 942 & 15.1\% \\
Office, education, and research & 1,169 & 28.0\% \\
Mixed use & 305 & 20.4\% \\
Civic, cultural, and health & 941 & 27.2\% \\
General interior & 678 & 14.8\% \\
Industrial and infrastructure & 82 & 21.8\% \\
Landscape and urban & 53 & 17.9\% \\
Unspecified source group & 154 & 18.8\% \\
\end{longtable}
\endgroup

\Needspace{9\baselineskip}
\noindent\begin{minipage}{\linewidth}\captionof{table}{Predicted semantic pixel composition of the full ArchP10k run.}\label{tab:A.3}\end{minipage}\addtocounter{table}{-1}

\begingroup
\small
\begin{longtable}[]{@{}lrr@{}}
\toprule\noalign{}
Predicted class & Pixels & Share of predicted non-background pixels \\
\midrule\noalign{}
\endhead
\bottomrule\noalign{}
\endlastfoot
\texttt{living\_room} & 1,419,369,397 & 24.21\% \\
\texttt{room\_default} & 987,723,922 & 16.85\% \\
\texttt{bedroom} & 764,413,067 & 13.04\% \\
\texttt{wall} & 699,703,665 & 11.94\% \\
\texttt{balcony} & 487,712,685 & 8.32\% \\
\texttt{kitchen} & 349,764,932 & 5.97\% \\
\texttt{bathroom} & 284,839,808 & 4.86\% \\
\texttt{stair} & 281,147,325 & 4.80\% \\
\texttt{corridor} & 196,332,949 & 3.35\% \\
\texttt{window} & 130,412,363 & 2.22\% \\
\texttt{entrance} & 76,736,868 & 1.31\% \\
\texttt{door} & 69,874,801 & 1.19\% \\
\texttt{storage} & 39,998,446 & 0.68\% \\
\texttt{elevator} & 37,906,193 & 0.65\% \\
\texttt{study} & 36,222,820 & 0.62\% \\
\end{longtable}
\endgroup

Pixel shares in Tables A.2-A.3 describe predictions. They cannot establish class prevalence in independently annotated drawings.

\section{Metric definitions and geometric audit}\label{app:B}

Let \(n_{ij}\) denote the number of GT-class \(i\) pixels predicted as \(j\). Class IoU, class accuracy, mean PA, and mIoU are given in Eqs. (B.1)-(B.2). Background contributes to confusion counts and OA but not to the four-class means. Empty-class unions are undefined rather than perfect; available class values enter each plan-level boundary macro.

\begin{equation}
\mathrm{IoU}_i=\frac{n_{ii}}{\sum_jn_{ij}+\sum_jn_{ji}-n_{ii}},\qquad \mathrm{PA}_i=\frac{n_{ii}}{\sum_jn_{ij}}.
\label{eq:B1}
\end{equation}

\begin{equation}
\mathrm{mIoU}=\frac1{|K|}\sum_{i\in K}\mathrm{IoU}_i,\quad\mathrm{mean\ PA}=\frac1{|K|}\sum_{i\in K}\mathrm{PA}_i,\quad\mathrm{OA}=\frac{\sum_i n_{ii}}{\sum_{ij}n_{ij}}.
\label{eq:B2}
\end{equation}

Boundary masks follow the official zero-padded erosion implementation \cite{ref26}. The band width is the rounded image-diagonal fraction, with a minimum of 1 pixel. Ten fractions from 0.1\% to 1.0\% are evaluated; their mean, the four-class mean, and the plan mean define the reported BIoU. Eq.~\eqref{eq:B3} applies to each class and band. The image-border padding is essential for a mask that touches the canvas.

\begin{equation}
\mathrm{BIoU}_d(P,T)=\frac{|B_d(P)\cap B_d(T)|}{|B_d(P)\cup B_d(T)|}.
\label{eq:B3}
\end{equation}

Original validity is measured from stored exterior and hole coordinates without repair, using Shapely 2.1.2, positive area, nonempty geometry, and its validity predicate. Malformed inputs count as invalid. The separately versioned repair is evaluated on the same original object IDs. A valid geometry collection with polygon area counts as polygon-containing, but a collapsed line does not. Sparse objects and auxiliary rendering groups have different denominators; Table~\ref{tab:B.1} retains both.

\Needspace{20\baselineskip}
\noindent\begin{minipage}{\linewidth}\captionof{table}{Original and guarded-repair geometry denominators.}\label{tab:B.1}\end{minipage}\addtocounter{table}{-1}

\begingroup
\small
\begin{longtable}[]{@{}ll@{}}
\toprule\noalign{}
Item & Count \\
\midrule\noalign{}
\endhead
\bottomrule\noalign{}
\endlastfoot
Original sparse objects & 783299 \\
Originally valid & 673880 \\
Valid polygon-containing objects after repair & 752510 \\
Accepted repairs & 78630 \\
Raster-guard rejection & 22963 \\
Collapsed without polygon & 7820 \\
Malformed inputs & 6 \\
All-object-valid plans after repair & 2932 \\
Original auxiliary render groups & 825215 \\
Invalid auxiliary render groups & 21534 \\
\end{longtable}
\endgroup

For boundary F1, let \(B_P,B_T\) be one-pixel contours and \(N_t\) the Euclidean neighborhood with radius \(t\) times the diagonal. Eq.~\eqref{eq:B4} counts each predicted or reference boundary pixel once. Tolerances are 0.25\%, 0.50\%, and 1.00\%; one absent boundary yields zero, while two absent boundaries are excluded with their count reported. Scores average over available classes and then plans.

Visible-wall clDice \cite{ref34} uses hard skeletons in the shared wall-evaluation domain, with precision and recall in Eq.~\eqref{eq:B5}. It measures centerline coverage, not door-room adjacency. Eq.~\eqref{eq:B6} gives class Dice from pooled confusion counts.

\begin{equation}
p_t=\frac{|B_P\cap N_t(B_T)|}{|B_P|},\quad r_t=\frac{|B_T\cap N_t(B_P)|}{|B_T|},\quad \mathrm{BF1}_t=\frac{2p_tr_t}{p_t+r_t}.
\label{eq:B4}
\end{equation}

\begin{equation}
p_c=\frac{|\operatorname{skel}(P)\cap T|}{|\operatorname{skel}(P)|},\quad r_c=\frac{|\operatorname{skel}(T)\cap P|}{|\operatorname{skel}(T)|},\quad \mathrm{clDice}=\frac{2p_cr_c}{p_c+r_c}.
\label{eq:B5}
\end{equation}

\begin{equation}
\mathrm{Dice}_i=\frac{2n_{ii}}{\sum_jn_{ij}+\sum_jn_{ji}}.
\label{eq:B6}
\end{equation}

\Needspace{9\baselineskip}
\noindent\begin{minipage}{\linewidth}\captionof{table}{Frozen registration and numerical guards.}\label{tab:B.2}\end{minipage}\addtocounter{table}{-1}

\begingroup
\small
\begin{longtable}[]{@{}
  >{\raggedright\arraybackslash}p{(\linewidth - 2\tabcolsep) * \real{0.5000}}
  >{\raggedright\arraybackslash}p{(\linewidth - 2\tabcolsep) * \real{0.5000}}@{}}
\toprule\noalign{}
\begin{minipage}[b]{\linewidth}\raggedright
Setting
\end{minipage} & \begin{minipage}[b]{\linewidth}\raggedright
Value
\end{minipage} \\
\midrule\noalign{}
\endhead
\bottomrule\noalign{}
\endlastfoot
Working resolution / ECC iterations / termination & 512 pixels / 150 / 0.00001 \\
Affine singular-value range / maximum ratio & 0.65-1.35 / 1.25 \\
Maximum rotation / corner displacement & 5 degrees / 25\% of diagonal \\
Foreground retained / source-score gain & At least 98\% / strictly positive \\
Calibration source-gain candidates & 0 / 0.01 / 0.03 / 0.06 \\
Calibration selection & Maximum pooled mIoU; BIoU decline at most 0.005 \\
Holdout acceptance & Pooled mIoU gain at least 0.02; BIoU decline at most 0.005; paired CI above zero \\
Polygon repair raster-difference cap & 2\% of before/after union \\
Affine numerical-repair area limit & max(1e-8, 1e-9 x absolute area) \\
On inadmissible registration or numerical repair & Identity transform; retain original frame \\
\end{longtable}
\endgroup

\subsection{Representation-stage implementation and audit}\label{representation-stage-implementation-and-audit}

The local extension operates on retained outputs and makes no model calls. It estimates an affine map from predicted structural boundaries to line evidence in the source drawing, using enhanced-correlation registration \cite{ref32}. At 512-pixel working resolution, initialization combines identity and similarity alignment. A candidate must improve the source-only structural overlap score, retain at least 98\% of foreground, and satisfy scale, rotation, anisotropy, and displacement guards. The same map is applied to integer labels by nearest-neighbor sampling and to vector coordinates; classes are not reassigned. Failed or inadmissible registration retains identity. This source-driven criterion uses no GT at inference and is not itself an accuracy metric.

Invalid sparse geometry is processed separately with linework-based validity repair \cite{ref33}. Originally valid coordinates are preserved. An invalid object's repair is accepted only when the raster symmetric difference is at most 2\% of the before/after union and the result is valid with positive polygon area. Geometry collections can retain lower-dimensional remnants; a collapsed line is not counted as a usable polygon. Original objects, rejected changes, malformed inputs, and per-object dispositions remain auditable. Affine roundoff is repaired only within numerical-area and raster guards; otherwise the whole plan retains its original frame. Relation records are referenced but not recomputed, so repaired geometry does not establish corrected topology.

Postprocessing was selected on existing validation outputs, separately from the fixed test evaluation. Of a previously fixed 60/40 calibration/holdout allocation, 44 calibration and 32 holdout outputs were complete; the missing 16/8 cases were not replaced. Identity, similarity, and affine registration were compared on calibration only. The selected configuration improved holdout pooled mIoU from 0.1808 to 0.3794, with a paired plan-mean change of 0.2107 {[}0.1733, 0.2476{]}. The configuration was then frozen before applying it to all 400 test plans. Earlier raw test scores had already been inspected, so this is an exploratory follow-up, not a claim of an untouched-test development history.

The supervised reruns retain their released checkpoints and preprocessing. The frozen local extension uses OpenCV 5.0.0, Shapely 2.1.2, and CPU processing with 8 workers for the corpus export. A lookup over all 16,777,216 RGB colors accelerates the existing pixelwise classifier without changing its labels; equality was checked on 20 real outputs, randomized layouts, gradients, and legacy color aliases. No remote requests, retraining, new image generation, or test-GT registration are involved. The same source-only registration is also applied to each baseline as a sensitivity check, while unmodified baseline outputs remain the primary comparison. Code, input and output hashes, transforms, and fallback reasons accompany the derived release.

Of 783,299 original sparse objects, 673,880 are valid (86.03\%). Guarded repair accepts 78,630 changes, raising valid polygon-containing objects to 752,510 (96.07\%) and all-object-valid plans from 384 to 2,932. Rejected or collapsed objects are retained with explicit status, not silently deleted. Among accepted repairs, raster differences total 1,090,285 pixels over a 1,241,315,793-pixel union (0.0878\%). Auxiliary rendering groups remain a distinct layer; their original validity rate is 97.39\%. Appendix F retains the original 30-case illustration set and is not used to evaluate the new repairs.

\Needspace{9\baselineskip}
\noindent\begin{minipage}{\linewidth}\captionof{table}{Frozen shared-decoder representation protocol.}\label{tab:B.3}\end{minipage}\addtocounter{table}{-1}

\begingroup
\small
\begin{longtable}[]{@{}
  >{\raggedright\arraybackslash}p{(\linewidth - 2\tabcolsep) * \real{0.5000}}
  >{\raggedright\arraybackslash}p{(\linewidth - 2\tabcolsep) * \real{0.5000}}@{}}
\toprule\noalign{}
\begin{minipage}[b]{\linewidth}\raggedright
Setting
\end{minipage} & \begin{minipage}[b]{\linewidth}\raggedright
Value
\end{minipage} \\
\midrule\noalign{}
\endhead
\bottomrule\noalign{}
\endlastfoot
Input & Same 400 test plans; final SALI-FP and native baselines \\
Extraction & Pixel-cell polygons; four-connectivity; retain holes and components \\
Simplification grid (pixels) & 0, 0.25, 0.5, 1, 2, 4, 8; preserve topology \\
Fidelity thresholds & 0.95, 0.98 (primary), 0.995; minimum nonempty-class IoU \\
Selection & Fewest vertices; then bytes; then tolerance \\
Serialization & Compact UTF-8 JSON; six decimal places; identical fields \\
Pixel replay & Pixel centers; room, boundary, door, window priority \\
Missing or empty input & Retain case in denominator; no compactness score \\
\end{longtable}
\endgroup

The experiment freezes source and code hashes before evaluating vector costs and makes no model calls. Zero-tolerance pixel-cell reconstruction is required to reproduce every source prediction exactly. Linework validity splitting resolves self-touching cell rings without deleting area. Polygon and hole counts are checked across every simplification candidate. Coordinates are serialized before raster replay, so quantization is included in the fidelity check. All selected candidates must be valid; a 100\% valid fraction here is an admission condition of the shared decoder, not evidence that native model geometry is universally valid. BCa intervals that are undefined for degenerate statistics are recorded as unavailable rather than replaced with another interval method.

The shared-decoder quantity \(V_\beta\) in Eq.~\eqref{eq:B7} is the smallest tested vertex budget whose classwise raster replay reaches fidelity \(\beta\) against its own prediction P. It is not a GT accuracy metric. \(N\) counts exterior and hole vertices without repeated closing points; \(\mathcal E\) is the fixed tolerance grid.

\begin{equation}
\begin{aligned}
F(P,V)&=\min_{c:\,|P_c|>0}\frac{|P_c\cap\widehat P_c(V)|}{|P_c\cup\widehat P_c(V)|},\\
V_\beta(P)&=\min_{\epsilon\in\mathcal E:\,F(P,V_\epsilon)\geq\beta,\;V_\epsilon\;\mathrm{valid}}N(V_\epsilon).
\end{aligned}
\label{eq:B7}
\end{equation}

\Needspace{9\baselineskip}
\noindent\begin{minipage}{\linewidth}\captionof{table}{Gate predicates and parameter provenance.}\label{tab:B.4}\end{minipage}\addtocounter{table}{-1}

\begingroup
\small
\begin{longtable}[]{@{}
  >{\raggedright\arraybackslash}p{(\linewidth - 2\tabcolsep) * \real{0.5000}}
  >{\raggedright\arraybackslash}p{(\linewidth - 2\tabcolsep) * \real{0.5000}}@{}}
\toprule\noalign{}
\begin{minipage}[b]{\linewidth}\raggedright
Step
\end{minipage} & \begin{minipage}[b]{\linewidth}\raggedright
Evidence and decision
\end{minipage} \\
\midrule\noalign{}
\endhead
\bottomrule\noalign{}
\endlastfoot
Schema and location & Require source-coordinate bbox, valid palette target, confidence and evidence fields \\
Confidence and area & Require confidence \textgreater= 0.80 and target mask \textgreater= 16 pixels \\
Current target protection & Reject existing target coverage \textgreater= 0.70; construction threshold 0.15 \\
Named-room protection & Reject other named-room fraction \textgreater= 0.30 \\
Room support & Require blank or fallback support \textgreater= 0.35 \\
Construction support & Require source intensity \textless{} 170 over \textgreater= 0.02 of the region; door exception \\
Image-change protection & Repair 0.55, cleanup 0.35, non-white fraction change 0.15 \\
Parameter origin & Inspected heuristic configuration; full historical source equivalence not established \\
\end{longtable}
\endgroup

\Needspace{9\baselineskip}
\noindent\begin{minipage}{\linewidth}\captionof{table}{Audit fields and coordinate roles.}\label{tab:B.5}\end{minipage}\addtocounter{table}{-1}

\begingroup
\small
\begin{longtable}[]{@{}
  >{\raggedright\arraybackslash}p{(\linewidth - 2\tabcolsep) * \real{0.5000}}
  >{\raggedright\arraybackslash}p{(\linewidth - 2\tabcolsep) * \real{0.5000}}@{}}
\toprule\noalign{}
\begin{minipage}[b]{\linewidth}\raggedright
Field
\end{minipage} & \begin{minipage}[b]{\linewidth}\raggedright
Meaning
\end{minipage} \\
\midrule\noalign{}
\endhead
\bottomrule\noalign{}
\endlastfoot
bbox; source\_coordinate\_space & Rough source-image region; image1\_source\_plan \\
semantic\_key; expected\_semantic\_key & Requested class from the frozen palette \\
current\_semantic\_key & Candidate class or white\_or\_unmapped / unknown \\
first\_pass\_problem; problem\_evidence & Visible candidate error and supporting observation \\
source\_evidence; fix\_rationale & Drawing cue and rationale for the proposed correction \\
confidence; source\_cues; reason & Model assessment and concise textual support \\
Source / candidate grids & 32 / 64 pixels; source is the coordinate authority \\
\end{longtable}
\endgroup

The reproduction materials distinguish archived generation/repair/cleanup prompts from the audit prompt recovered from currently inspected source. Full audit text cannot be certified against every historical run. Retained records use deployment aliases rather than immutable snapshots; cache reuse in the inspected workflow checks output availability, successful status, prompt hash, and image size, not a complete historical model fingerprint. A content-addressed evaluation cache is frozen separately for this revision.

Integer-label assignment is pixelwise and order-dependent. Direct matches require every RGB channel to lie within 20 of a current or legacy palette entry; the minimum squared RGB distance wins, with palette order breaking ties. Only unassigned chromatic pixels enter white-mixture matching: the median channelwise mixture coefficient must lie in {[}0.06, 1.05{]}, off-ray deviation must not exceed 24, and channel span must be at least 4. The first qualifying palette entry wins. Remaining pixels enter the ordered rules in Table~\ref{tab:B.7} only if value is at least 40, channel span at least 35, and they are not near-white background (minimum RGB channel at least 240 and span at most 25). Each rule claims only unassigned pixels; the remainder is background. Gray classes can be assigned by direct matching before the chromatic stages. The frozen RGB lookup performs assignment only, not object-size filtering or morphological cleanup. Its source hash matches the inspected classifier; historical object parameters are read from each saved pointset, not inferred from current defaults. Tables B.6--B.8 and the full configuration inventory preserve these distinctions.

\Needspace{9\baselineskip}
\noindent\begin{minipage}{\linewidth}\captionof{table}{Executed color-assignment stages.}\label{tab:B.6}\end{minipage}\addtocounter{table}{-1}

\begingroup
\small
\begin{longtable}[]{@{}
  >{\raggedright\arraybackslash}p{(\linewidth - 2\tabcolsep) * \real{0.5000}}
  >{\raggedright\arraybackslash}p{(\linewidth - 2\tabcolsep) * \real{0.5000}}@{}}
\toprule\noalign{}
\begin{minipage}[b]{\linewidth}\raggedright
Stage
\end{minipage} & \begin{minipage}[b]{\linewidth}\raggedright
Rule
\end{minipage} \\
\midrule\noalign{}
\endhead
\bottomrule\noalign{}
\endlastfoot
1. Palette thresholds & Per-channel tolerance 20; minimum squared RGB distance; current and legacy aliases \\
2. White mixtures & Unassigned chromatic pixels; t in {[}0.06, 1.05{]}; off-ray \textless= 24; span \textgreater= 4 \\
3. Ordered hue predicates & Unassigned, non-background pixels; value \textgreater= 40; span \textgreater= 35; Table~\ref{tab:B.7} \\
4. Background & All remaining pixels receive integer ID 0 \\
Frozen provenance & All-RGB LUT and classifier source hashes; stored pointset configuration per plan \\
\end{longtable}
\endgroup

\Needspace{9\baselineskip}
\noindent\begin{minipage}{\linewidth}\captionof{table}{Ordered hue predicates after direct and mixture matching; H is in degrees.}\label{tab:B.7}\end{minipage}\addtocounter{table}{-1}

\begingroup
\small
\begin{longtable}[]{@{}
  >{\raggedright\arraybackslash}p{(\linewidth - 4\tabcolsep) * \real{0.0900}}
  >{\raggedright\arraybackslash}p{(\linewidth - 4\tabcolsep) * \real{0.2000}}
  >{\raggedright\arraybackslash}p{(\linewidth - 4\tabcolsep) * \real{0.7100}}@{}}
\toprule\noalign{}
\begin{minipage}[b]{\linewidth}\raggedright
Order
\end{minipage} & \begin{minipage}[b]{\linewidth}\raggedright
Target
\end{minipage} & \begin{minipage}[b]{\linewidth}\raggedright
Predicate
\end{minipage} \\
\midrule\noalign{}
\endhead
\bottomrule\noalign{}
\endlastfoot
1 & wall & ((H \textless= 15) or (H \textgreater= 350)) and (R \textgreater{} 180) and (G \textless{} 100) and (B \textless{} 100) \\
2 & door & (H \textgreater= 45) and (H \textless= 68) and (R \textgreater{} 180) and (G \textgreater{} 180) and (B \textless{} 100) \\
3 & window & (H \textgreater= 85) and (H \textless= 150) and (G \textgreater{} 100) and (R \textless{} 110) and (B \textless{} 130) \\
4 & stair & (H \textgreater= 200) and (H \textless= 245) and (B \textgreater{} 120) and (R \textless{} 100) and (G \textless{} 160) and (maxRGB \textgreater= 170) \\
5 & elevator & (H \textgreater= 20) and (H \textless= 44) and (R \textgreater{} 180) and (G \textgreater{} 70) and (G \textless{} 180) and (B \textless{} 120) \\
6 & living room & (H \textgreater= 170) and (H \textless= 192) and (R \textless{} 90) and (G \textgreater{} 190) and (B \textgreater{} 190) and (maxRGB \textgreater= 220) \\
7 & bathroom & (H \textgreater= 170) and (H \textless= 192) and (R \textless{} 90) and (G \textgreater= 90) and (B \textgreater= 90) and (maxRGB \textless{} 220) \\
8 & living room & (H \textgreater= 176) and (H \textless= 196) and (R \textless{} 130) and (G \textgreater= 180) and (B \textgreater= 190) and (maxRGB \textgreater= 200) \\
9 & bedroom & (H \textgreater= 315) and (H \textless= 350) and (R \textgreater{} 180) and (G \textgreater= 50) and (G \textless= 170) and (B \textgreater= 120) and (B \textless= 235) \\
10 & bedroom & (H \textgreater= 316) and (H \textless= 340) and (R \textgreater{} 180) and (G \textless{} 60) and (B \textgreater= 100) and (B \textless= 210) and (B * 100 \textless{} R * 90) \\
11 & kitchen & (H \textgreater= 255) and (H \textless= 290) and (B \textgreater{} 150) and (R \textgreater= 80) and (R \textless= 215) and (G \textless{} 130) \\
12 & balcony & (H \textgreater= 68) and (H \textless= 100) and (R \textgreater= 120) and (G \textgreater= 150) and (B \textless{} 120) \\
13 & entrance & (H \textgreater= 290) and (H \textless= 315) and (R \textgreater{} 180) and (B \textgreater{} 180) and (G \textless{} 90) \\
14 & storage & (H \textgreater= 10) and (H \textless= 35) and (R \textgreater= 80) and (R \textless= 180) and (G \textgreater= 30) and (G \textless= 130) and (B \textless{} 120) and (maxRGB \textless{} 220) \\
15 & storage & (H \textgreater= 25) and (H \textless= 52) and (R \textgreater= 180) and (G \textgreater= 135) and (G \textless= 225) and (B \textgreater= 100) and (B \textless= 195) and (span \textgreater= 45) and (maxRGB \textgreater= 170) \\
16 & study & (H \textgreater= 220) and (H \textless= 250) and (B \textgreater= 80) and (R \textless{} 100) and (G \textless{} 100) and (maxRGB \textless{} 190) \\
17 & study & (H \textgreater= 205) and (H \textless= 250) and (B \textgreater= 80) and (R \textless{} 120) and (G \textless{} 130) and (maxRGB \textless{} 190) \\
18 & room default & (maxRGB \textgreater= 185) and (minRGB \textgreater= 145) and (span \textgreater= 30) and (span \textless= 95) \\
\end{longtable}
\endgroup

\Needspace{9\baselineskip}
\noindent\begin{minipage}{\linewidth}\captionof{table}{Serialized native and auxiliary representation settings (all 11,534 records).}\label{tab:B.8}\end{minipage}\addtocounter{table}{-1}

\begingroup
\small
\begin{longtable}[]{@{}
  >{\raggedright\arraybackslash}p{(\linewidth - 2\tabcolsep) * \real{0.5000}}
  >{\raggedright\arraybackslash}p{(\linewidth - 2\tabcolsep) * \real{0.5000}}@{}}
\toprule\noalign{}
\begin{minipage}[b]{\linewidth}\raggedright
Operation
\end{minipage} & \begin{minipage}[b]{\linewidth}\raggedright
Saved setting
\end{minipage} \\
\midrule\noalign{}
\endhead
\bottomrule\noalign{}
\endlastfoot
Native contour simplification / minimum area & 1.2 px / 24 px squared \\
Rectangle eligibility & Area ratio \textgreater= 0.9; at most 6 approximated corners \\
Coordinate offset / short-edge length & 0.5 / 2.0 px \\
Native orthogonal snap & 1.5 px \\
Auxiliary room / construction simplification & 0.6 / 4.0 px \\
Auxiliary expansion: room / construction / opening & 1 / 1 / 1 px \\
Construction axis snap / gap close & 5.0 / 1 px \\
Construction rectangle fill / outlier limit & 0.72 / 0.08 \\
Surface gap kernel / room-fragment area & 9 px / 2048 px squared \\
Maximum linkage points per pair & 4 \\
\end{longtable}
\endgroup

Native objects retain exterior and hole coordinates; qualifying hole-free contours may use rectangle fitting, otherwise they use sparse polygon approximation and orthogonal/short-edge cleanup. Auxiliary rings follow class-specific masks and smoothing, so their group count can differ from native objects. Existing holes are not silently counted as filled area in native storage. Rendering order and masks are frozen with the saved artifacts. Some inspected code defaults, including an additional construction-polygon tolerance, were not serialized in historical records; their equivalence across all historical runs cannot be certified. This inventory therefore distinguishes recorded values from reconstruction assumptions.

\Needspace{24\baselineskip}
\section{Detailed measured results}\label{app:C}

\Needspace{18\baselineskip}
\noindent\begin{minipage}{\linewidth}\captionof{table}{CubiCasa\allowbreak 5K test: original and identically registered outputs.}\label{tab:C.1}\end{minipage}\addtocounter{table}{-1}

\begingroup
\small
\begin{longtable}[]{@{}lll@{}}
\toprule\noalign{}
Method / stage & Metric & Estimate {[}95\% BCa CI{]} \\
\midrule\noalign{}
\endhead
\bottomrule\noalign{}
\endlastfoot
SALI-FP / raw & oa & 0.7830 {[}0.7757, 0.7899{]} \\
SALI-FP / raw & mean\_pa & 0.3001 {[}0.2903, 0.3123{]} \\
SALI-FP / raw & miou & 0.2074 {[}0.2000, 0.2162{]} \\
SALI-FP / raw & biou & 0.0570 {[}0.0530, 0.0623{]} \\
SALI-FP / raw & Visible-wall IoU & 0.1116 {[}0.1013, 0.1245{]} \\
SALI-FP / raw & Wall clDice & 0.1728 {[}0.1589, 0.1892{]} \\
SALI-FP / registered & oa & 0.8657 {[}0.8574, 0.8733{]} \\
SALI-FP / registered & mean\_pa & 0.4797 {[}0.4639, 0.4951{]} \\
SALI-FP / registered & miou & 0.3596 {[}0.3449, 0.3740{]} \\
SALI-FP / registered & biou & 0.1858 {[}0.1756, 0.1961{]} \\
SALI-FP / registered & Visible-wall IoU & 0.3325 {[}0.3105, 0.3549{]} \\
SALI-FP / registered & Wall clDice & 0.5113 {[}0.4857, 0.5362{]} \\
CubiCasa\allowbreak 5K / raw & oa & 0.9517 {[}0.9456, 0.9566{]} \\
CubiCasa\allowbreak 5K / raw & mean\_pa & 0.8230 {[}0.8122, 0.8313{]} \\
CubiCasa\allowbreak 5K / raw & miou & 0.7390 {[}0.7270, 0.7496{]} \\
CubiCasa\allowbreak 5K / raw & biou & 0.5546 {[}0.5444, 0.5643{]} \\
CubiCasa\allowbreak 5K / raw & Visible-wall IoU & 0.7482 {[}0.7344, 0.7603{]} \\
CubiCasa\allowbreak 5K / raw & Wall clDice & 0.8575 {[}0.8482, 0.8660{]} \\
CubiCasa\allowbreak 5K / registered & oa & 0.9468 {[}0.9405, 0.9517{]} \\
CubiCasa\allowbreak 5K / registered & mean\_pa & 0.7996 {[}0.7885, 0.8085{]} \\
CubiCasa\allowbreak 5K / registered & miou & 0.7015 {[}0.6890, 0.7128{]} \\
CubiCasa\allowbreak 5K / registered & biou & 0.4947 {[}0.4842, 0.5047{]} \\
CubiCasa\allowbreak 5K / registered & Visible-wall IoU & 0.6972 {[}0.6820, 0.7107{]} \\
CubiCasa\allowbreak 5K / registered & Wall clDice & 0.8314 {[}0.8199, 0.8417{]} \\
MiT-UNet / raw & Visible-wall IoU & 0.7721 {[}0.7435, 0.7843{]} \\
MiT-UNet / raw & Wall clDice & 0.9302 {[}0.9219, 0.9372{]} \\
MiT-UNet / registered & Visible-wall IoU & 0.7129 {[}0.6892, 0.7309{]} \\
MiT-UNet / registered & Wall clDice & 0.8901 {[}0.8743, 0.9030{]} \\
\end{longtable}
\endgroup

\Needspace{9\baselineskip}
\noindent\begin{minipage}{\linewidth}\captionof{table}{ArchP10k partitions: internal consistency only.}\label{tab:C.2}\end{minipage}\addtocounter{table}{-1}

\begingroup
\small
\begin{longtable}[]{@{}llll@{}}
\toprule\noalign{}
Partition & Plans & Mean RCR {[}95\% CI{]} & Micro RCR \\
\midrule\noalign{}
\endhead
\bottomrule\noalign{}
\endlastfoot
full & 11534 & 0.9569 {[}0.9565, 0.9573{]} & 0.9595 \\
development & 9225 & 0.9571 {[}0.9566, 0.9575{]} & 0.9597 \\
validation & 1157 & 0.9566 {[}0.9552, 0.9578{]} & 0.9592 \\
test & 1152 & 0.9556 {[}0.9541, 0.9569{]} & 0.9582 \\
hard & 400 & 0.9507 {[}0.9483, 0.9530{]} & 0.9531 \\
nonhard test & 752 & 0.9582 {[}0.9565, 0.9596{]} & 0.9612 \\
\end{longtable}
\endgroup

\Needspace{15\baselineskip}
\noindent\begin{minipage}{\linewidth}\captionof{table}{Retained production call and stage records.}\label{tab:C.3}\end{minipage}\addtocounter{table}{-1}

\begingroup
\small
\begin{longtable}[]{@{}
  >{\raggedright\arraybackslash}p{(\linewidth - 4\tabcolsep) * \real{0.3333}}
  >{\raggedright\arraybackslash}p{(\linewidth - 4\tabcolsep) * \real{0.3333}}
  >{\raggedright\arraybackslash}p{(\linewidth - 4\tabcolsep) * \real{0.3333}}@{}}
\toprule\noalign{}
\begin{minipage}[b]{\linewidth}\raggedright
Quantity
\end{minipage} & \begin{minipage}[b]{\linewidth}\raggedright
Value
\end{minipage} & \begin{minipage}[b]{\linewidth}\raggedright
Interpretation
\end{minipage} \\
\midrule\noalign{}
\endhead
\bottomrule\noalign{}
\endlastfoot
Logical calls / HTTP attempts & 39,875 / 41,931 & 2,056 excess request attempts \\
Audit median {[}IQR{]} & 39.3 {[}24.0, 62.5{]} s & Recorded remote stage \\
Conditional repair median {[}IQR{]} & 164.8 {[}100.9, 235.0{]} s & Only triggered repairs \\
Cleanup median {[}IQR{]} & 138.4 {[}93.8, 211.5{]} s & Recorded remote stage \\
Recorded subtotal median & 260.8 s & Not full end-to-end time \\
Reference-rate total / mean & \$9,430.83 / \$0.818 & Rate-card calculation, not billing \\
Reference-rate median {[}IQR{]} & \$0.82 {[}\$0.66, \$0.91{]} & Not a measured monetary saving \\
\end{longtable}
\endgroup

The cost calculation is retained solely as an explicitly labeled resource estimate derived from request records. It is not an invoice. Initial-generation and reused local-stage timestamps do not yield reliable full latency; aggregate wall-clock throughput is therefore omitted. No manual labor rate or editing-time saving is multiplied into a claimed return on investment.

\Needspace{9\baselineskip}
\noindent\begin{minipage}{\linewidth}\captionof{table}{Full-run complexity quartiles.}\label{tab:C.4}\end{minipage}\addtocounter{table}{-1}

\begingroup
\small
\begin{longtable}[]{@{}lllll@{}}
\toprule\noalign{}
Quartile & Plans & Mean RCR & Protected (\%) & Median points \\
\midrule\noalign{}
\endhead
\bottomrule\noalign{}
\endlastfoot
1 & 2884 & 0.9607 & 15.46 & 651.0 \\
2 & 2883 & 0.9578 & 18.38 & 827.0 \\
3 & 2883 & 0.9561 & 19.11 & 999.0 \\
4 & 2884 & 0.9529 & 22.68 & 1288.0 \\
\end{longtable}
\endgroup

\Needspace{9\baselineskip}
\noindent\begin{minipage}{\linewidth}\captionof{table}{Full-run outcome denominators.}\label{tab:C.5}\end{minipage}\addtocounter{table}{-1}

\begingroup
\small
\begin{longtable}[]{@{}ll@{}}
\toprule\noalign{}
Quantity & Count \\
\midrule\noalign{}
\endhead
\bottomrule\noalign{}
\endlastfoot
Structured outputs & 11534 \\
One repair & 5273 \\
No repair & 6261 \\
Protective status & 2181 \\
Suggested / accepted / rejected & 11,481 / 8,033 / 3,448 \\
RCR below 0.90 & 191 \\
RCR below 0.95 & 3098 \\
\end{longtable}
\endgroup

\begin{figure}[!htbp]
  \centering
  \includegraphics[width=\textwidth,height=0.82\textheight,keepaspectratio]{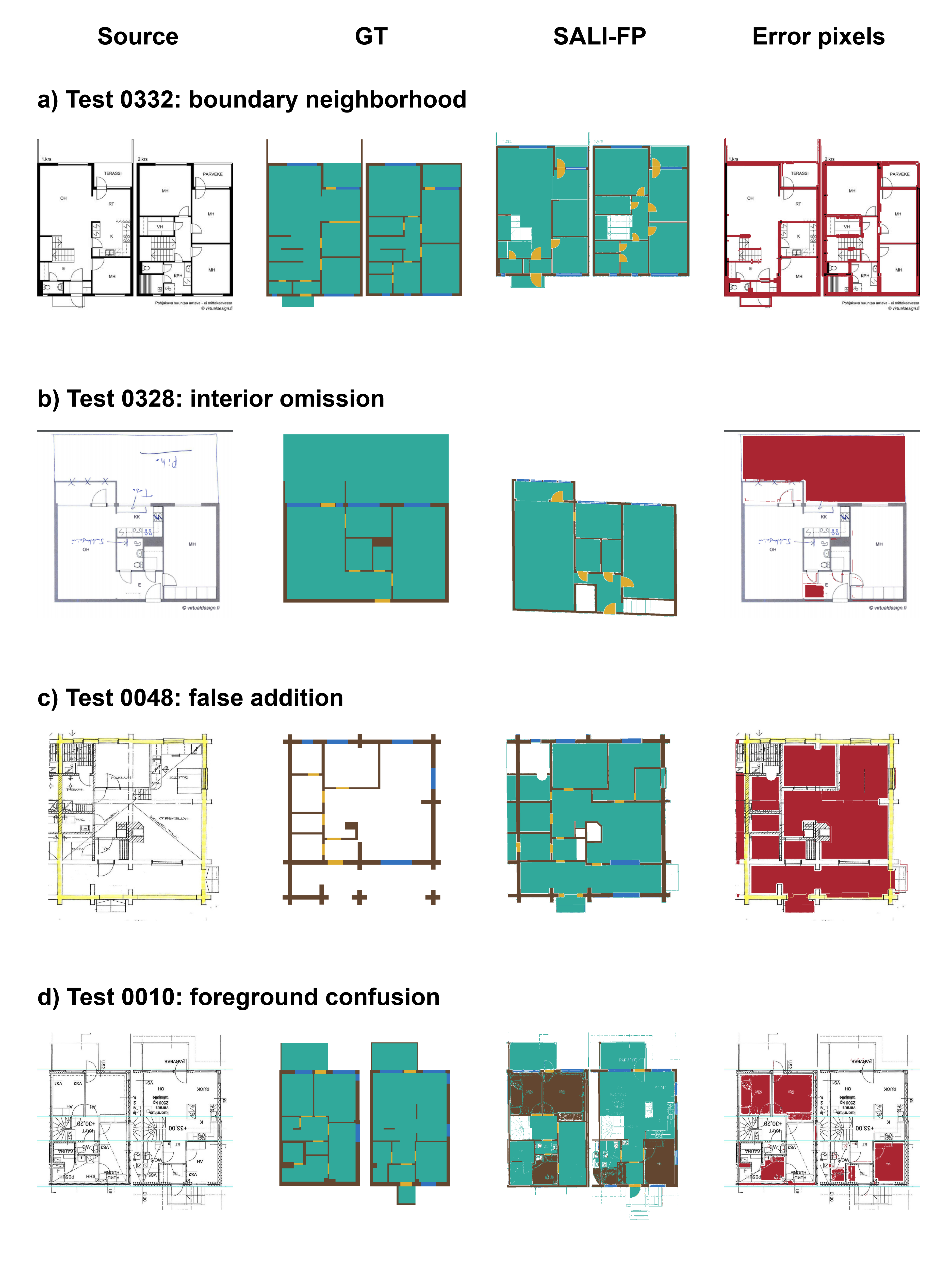}
  \caption{Systematic-error examples: source, GT, final SALI-FP, and selected error pixels. Each case maximizes one absolute error count. Test 0048 lacks Space references; its room-addition mask also reflects annotation scope (Table~\ref{tab:C.23}).}
  \label{fig:C.1}
\end{figure}
\FloatBarrier

Tables C.6-C.8 and Fig. C.2 separate export behavior, tolerance-dependent boundary agreement, and class overlap. Raw supervised outputs must not be replaced by their lower-scoring registered variants when judging SALI-FP. The original examples in Fig. C.1 and Appendix F are unchanged.

\Needspace{9\baselineskip}
\noindent\begin{minipage}{\linewidth}\captionof{table}{Full ArchP10k local export audit; no additional model calls.}\label{tab:C.6}\end{minipage}\addtocounter{table}{-1}

\begingroup
\small
\begin{longtable}[]{@{}
  >{\raggedright\arraybackslash}p{(\linewidth - 2\tabcolsep) * \real{0.5000}}
  >{\raggedright\arraybackslash}p{(\linewidth - 2\tabcolsep) * \real{0.5000}}@{}}
\toprule\noalign{}
\begin{minipage}[b]{\linewidth}\raggedright
Quantity
\end{minipage} & \begin{minipage}[b]{\linewidth}\raggedright
Measured value
\end{minipage} \\
\midrule\noalign{}
\endhead
\bottomrule\noalign{}
\endlastfoot
Exported plans & 11534 \\
Registration accepted & 10981 \\
Whole-plan geometry fallback & 14 \\
Objects receiving affine numerical repair & 72 \\
Plans with out-of-canvas coordinates & 597 \\
Mean / micro RCR, repaired geometry before registration & 0.9290 / 0.9341 \\
Mean / micro RCR, repaired geometry after registration & 0.9284 / 0.9339 \\
Local seconds per case, median {[}IQR{]} & 0.908 {[}0.729, 1.173{]} \\
Additional API calls & 0 \\
\end{longtable}
\endgroup

\Needspace{9\baselineskip}
\noindent\begin{minipage}{\linewidth}\captionof{table}{Common-class boundary F1 across diagonal tolerances (400 plans).}\label{tab:C.7}\end{minipage}\addtocounter{table}{-1}

\begingroup
\small
\begin{longtable}[]{@{}llll@{}}
\toprule\noalign{}
Method / stage & 0.25\% & 0.50\% & 1.00\% \\
\midrule\noalign{}
\endhead
\bottomrule\noalign{}
\endlastfoot
SALI-FP / raw & 0.1174 & 0.2100 & 0.3271 \\
SALI-FP / registered & 0.3264 & 0.4962 & 0.6131 \\
CubiCasa\allowbreak 5K / raw & 0.8300 & 0.8888 & 0.9127 \\
CubiCasa\allowbreak 5K / registered & 0.7820 & 0.8821 & 0.9108 \\
\end{longtable}
\endgroup

\Needspace{25\baselineskip}
\noindent\begin{minipage}{\linewidth}\captionof{table}{Pooled class overlap on the common four-class target.}\label{tab:C.8}\end{minipage}\addtocounter{table}{-1}

\begingroup
\small
\begin{longtable}[]{@{}llll@{}}
\toprule\noalign{}
Method / stage & Class & IoU & Dice \\
\midrule\noalign{}
\endhead
\bottomrule\noalign{}
\endlastfoot
SALI-FP / raw & Boundary & 0.1094 & 0.1972 \\
SALI-FP / raw & Door/opening & 0.0396 & 0.0762 \\
SALI-FP / raw & Window & 0.0520 & 0.0989 \\
SALI-FP / raw & Room & 0.6284 & 0.7718 \\
SALI-FP / registered & Boundary & 0.3180 & 0.4826 \\
SALI-FP / registered & Door/opening & 0.1454 & 0.2539 \\
SALI-FP / registered & Window & 0.2318 & 0.3763 \\
SALI-FP / registered & Room & 0.7433 & 0.8528 \\
CubiCasa\allowbreak 5K / raw & Boundary & 0.7173 & 0.8354 \\
CubiCasa\allowbreak 5K / raw & Door/opening & 0.6181 & 0.7640 \\
CubiCasa\allowbreak 5K / raw & Window & 0.7199 & 0.8372 \\
CubiCasa\allowbreak 5K / raw & Room & 0.9006 & 0.9477 \\
CubiCasa\allowbreak 5K / registered & Boundary & 0.6697 & 0.8022 \\
CubiCasa\allowbreak 5K / registered & Door/opening & 0.5720 & 0.7277 \\
CubiCasa\allowbreak 5K / registered & Window & 0.6727 & 0.8043 \\
CubiCasa\allowbreak 5K / registered & Room & 0.8917 & 0.9427 \\
\end{longtable}
\endgroup

\begin{figure}[!htbp]
  \centering
  \includegraphics[width=\textwidth,height=0.82\textheight,keepaspectratio]{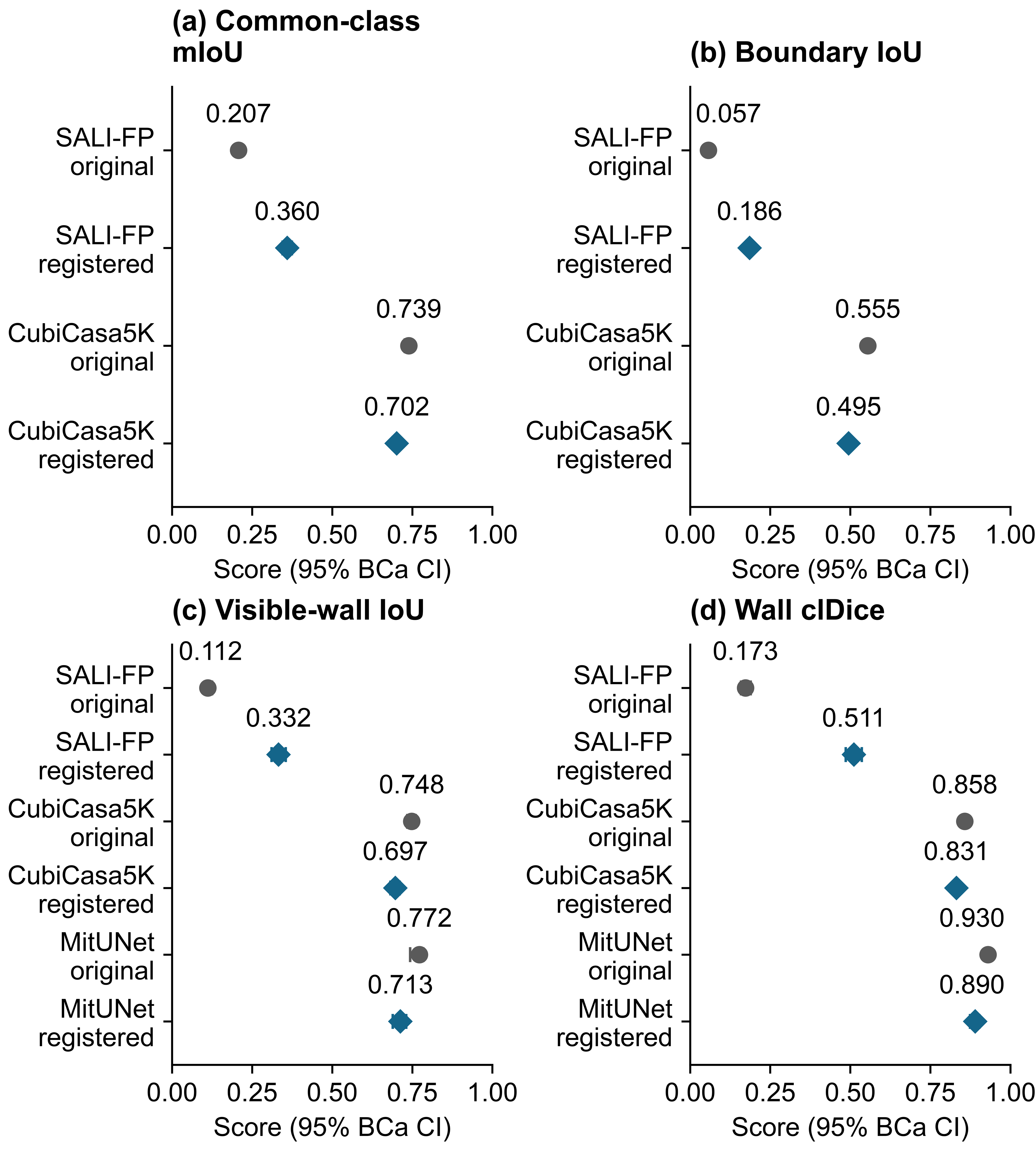}
  \caption{All-method sensitivity to identical source-only registration. Gray circles denote original outputs; blue diamonds denote registered outputs. Unmodified supervised predictions remain the primary baseline. MiT-UNet is included only for visible-wall metrics.}
  \label{fig:C.2}
\end{figure}
\FloatBarrier

Tables C.9 and C.10 retain the wall-only comparison and all primary paired cost ratios. Reconstruction fidelity refers to each predicted mask; GT wall IoU retains the visible-wall domain of the accuracy protocol. The common-four-class vertex medians {[}IQR{]} are 3,641.0 {[}2,454.0, 5,616.5{]} for SALI-FP and 5,785.0 {[}4,525.5, 6,914.5{]} for CubiCasa\allowbreak 5K; storage medians {[}IQR{]} are 79.0 {[}53.0, 122.2{]} and 91.0 {[}70.2, 109.4{]} KiB. At 0.95/0.995 fidelity the corresponding vertex medians are 3,602/3,641 and 5,660/5,785. All 400 cases satisfy each threshold; every zero-tolerance replay is exact. Per-class constraints often retain a lossless candidate to protect thin elements. This evaluation measures raster-derived representation economy, not equivalence of the methods' semantic correctness or native vector decoders.

\Needspace{9\baselineskip}
\noindent\begin{minipage}{\linewidth}\captionof{table}{Wall-target accuracy and prediction-referenced vector costs at 0.98 fidelity.}\label{tab:C.9}\end{minipage}\addtocounter{table}{-1}

\begingroup
\small
\begin{longtable}[]{@{}
  >{\raggedright\arraybackslash}p{(\linewidth - 8\tabcolsep) * \real{0.2000}}
  >{\raggedright\arraybackslash}p{(\linewidth - 8\tabcolsep) * \real{0.2000}}
  >{\raggedright\arraybackslash}p{(\linewidth - 8\tabcolsep) * \real{0.2000}}
  >{\raggedright\arraybackslash}p{(\linewidth - 8\tabcolsep) * \real{0.2000}}
  >{\raggedright\arraybackslash}p{(\linewidth - 8\tabcolsep) * \real{0.2000}}@{}}
\toprule\noalign{}
\begin{minipage}[b]{\linewidth}\raggedright
Method
\end{minipage} & \begin{minipage}[b]{\linewidth}\raggedright
GT wall IoU
\end{minipage} & \begin{minipage}[b]{\linewidth}\raggedright
Vertices {[}IQR{]}
\end{minipage} & \begin{minipage}[b]{\linewidth}\raggedright
KiB {[}IQR{]}
\end{minipage} & \begin{minipage}[b]{\linewidth}\raggedright
Achieved
\end{minipage} \\
\midrule\noalign{}
\endhead
\bottomrule\noalign{}
\endlastfoot
SALI-FP & 0.3325 & 644.0 {[}422.0, 967.0{]} & 12.3 {[}7.7, 18.3{]} & 400/400 \\
CubiCasa\allowbreak 5K & 0.7482 & 2,912.0 {[}2,263.5, 3,574.5{]} & 47.5 {[}37.5, 57.7{]} & 400/400 \\
MiT-UNet & 0.7721 & 431.0 {[}309.5, 570.5{]} & 6.5 {[}4.6, 8.8{]} & 400/400 \\
\end{longtable}
\endgroup

\Needspace{9\baselineskip}
\noindent\begin{minipage}{\linewidth}\captionof{table}{Matched median cost ratios: SALI-FP divided by each baseline.}\label{tab:C.10}\end{minipage}\addtocounter{table}{-1}

\begingroup
\small
\begin{longtable}[]{@{}lllll@{}}
\toprule\noalign{}
Target & Baseline & Cost & Ratio {[}95\% BCa CI{]} & Paired \\
\midrule\noalign{}
\endhead
\bottomrule\noalign{}
\endlastfoot
common4 & CubiCasa\allowbreak 5K & vertices & 0.639 {[}0.610, 0.678{]} & 400/400 \\
common4 & CubiCasa\allowbreak 5K & bytes & 0.886 {[}0.824, 0.941{]} & 400/400 \\
wall & CubiCasa\allowbreak 5K & vertices & 0.217 {[}0.202, 0.232{]} & 400/400 \\
wall & CubiCasa\allowbreak 5K & bytes & 0.245 {[}0.233, 0.269{]} & 400/400 \\
wall & MiT-UNet & vertices & 1.438 {[}1.353, 1.534{]} & 400/400 \\
wall & MiT-UNet & bytes & 1.692 {[}1.599, 1.797{]} & 400/400 \\
\end{longtable}
\endgroup

\Needspace{9\baselineskip}
\noindent\begin{minipage}{\linewidth}\captionof{table}{Recorded paired stage changes; per-plan mean differences.}\label{tab:C.11}\end{minipage}\addtocounter{table}{-1}

\begingroup
\small
\begin{longtable}[]{@{}
  >{\raggedright\arraybackslash}p{(\linewidth - 8\tabcolsep) * \real{0.2000}}
  >{\raggedright\arraybackslash}p{(\linewidth - 8\tabcolsep) * \real{0.2000}}
  >{\raggedright\arraybackslash}p{(\linewidth - 8\tabcolsep) * \real{0.2000}}
  >{\raggedright\arraybackslash}p{(\linewidth - 8\tabcolsep) * \real{0.2000}}
  >{\raggedright\arraybackslash}p{(\linewidth - 8\tabcolsep) * \real{0.2000}}@{}}
\toprule\noalign{}
\begin{minipage}[b]{\linewidth}\raggedright
Group
\end{minipage} & \begin{minipage}[b]{\linewidth}\raggedright
Transition
\end{minipage} & \begin{minipage}[b]{\linewidth}\raggedright
Metric
\end{minipage} & \begin{minipage}[b]{\linewidth}\raggedright
Change {[}95\% BCa CI{]}
\end{minipage} & \begin{minipage}[b]{\linewidth}\raggedright
Plans
\end{minipage} \\
\midrule\noalign{}
\endhead
\bottomrule\noalign{}
\endlastfoot
all & protected\_repair minus initial & plan-mean miou & -0.0088 {[}-0.0131, -0.0045{]} & 400 \\
all & protected\_repair minus initial & plan-mean biou & -0.0032 {[}-0.0060, -0.0005{]} & 400 \\
all & cleanup minus protected\_repair & plan-mean miou & 0.0055 {[}0.0032, 0.0078{]} & 400 \\
all & cleanup minus protected\_repair & plan-mean biou & -0.0040 {[}-0.0055, -0.0025{]} & 400 \\
all & final minus cleanup & plan-mean miou & 0.1699 {[}0.1579, 0.1821{]} & 400 \\
all & final minus cleanup & plan-mean biou & 0.1288 {[}0.1195, 0.1383{]} & 400 \\
repair\_triggered & protected\_repair minus initial & plan-mean miou & -0.0134 {[}-0.0201, -0.0069{]} & 263 \\
repair\_triggered & protected\_repair minus initial & plan-mean biou & -0.0049 {[}-0.0092, -0.0007{]} & 263 \\
repair\_triggered & cleanup minus protected\_repair & plan-mean miou & 0.0062 {[}0.0039, 0.0087{]} & 263 \\
repair\_triggered & cleanup minus protected\_repair & plan-mean biou & -0.0048 {[}-0.0065, -0.0032{]} & 263 \\
repair\_triggered & final minus cleanup & plan-mean miou & 0.1567 {[}0.1421, 0.1713{]} & 263 \\
repair\_triggered & final minus cleanup & plan-mean biou & 0.1187 {[}0.1075, 0.1304{]} & 263 \\
\end{longtable}
\endgroup

\Needspace{9\baselineskip}
\noindent\begin{minipage}{\linewidth}\captionof{table}{Native geometry: F1 of mean precision and recall; overlap filter disabled.}\label{tab:C.12}\end{minipage}\addtocounter{table}{-1}

\begingroup
\small
\begin{longtable}[]{@{}
  >{\raggedright\arraybackslash}p{(\linewidth - 8\tabcolsep) * \real{0.2000}}
  >{\raggedright\arraybackslash}p{(\linewidth - 8\tabcolsep) * \real{0.2000}}
  >{\raggedright\arraybackslash}p{(\linewidth - 8\tabcolsep) * \real{0.2000}}
  >{\raggedright\arraybackslash}p{(\linewidth - 8\tabcolsep) * \real{0.2000}}
  >{\raggedright\arraybackslash}p{(\linewidth - 8\tabcolsep) * \real{0.2000}}@{}}
\toprule\noalign{}
\begin{minipage}[b]{\linewidth}\raggedright
Method
\end{minipage} & \begin{minipage}[b]{\linewidth}\raggedright
Metric
\end{minipage} & \begin{minipage}[b]{\linewidth}\raggedright
F1 {[}95\% BCa CI{]}
\end{minipage} & \begin{minipage}[b]{\linewidth}\raggedright
Mean P / R
\end{minipage} & \begin{minipage}[b]{\linewidth}\raggedright
Mean plan F1
\end{minipage} \\
\midrule\noalign{}
\endhead
\bottomrule\noalign{}
\endlastfoot
SALI-FP & Room & 0.4450 {[}0.4221, 0.4683{]} & 0.3647 / 0.5705 & 0.4271 \\
SALI-FP & Corner & 0.1590 {[}0.1494, 0.1690{]} & 0.0973 / 0.4345 & 0.1559 \\
SALI-FP & Angle & 0.0932 {[}0.0863, 0.1002{]} & 0.0573 / 0.2495 & 0.0916 \\
CubiCasa\allowbreak 5K / polygons & Room & 0.6305 {[}0.6093, 0.6523{]} & 0.6446 / 0.6170 & 0.6175 \\
CubiCasa\allowbreak 5K / polygons & Corner & 0.3274 {[}0.3134, 0.3424{]} & 0.2429 / 0.5020 & 0.3167 \\
CubiCasa\allowbreak 5K / polygons & Angle & 0.2358 {[}0.2237, 0.2494{]} & 0.1763 / 0.3563 & 0.2278 \\
Raster2Seq & Room & 0.7626 {[}0.7455, 0.7787{]} & 0.7504 / 0.7752 & 0.7531 \\
Raster2Seq & Corner & 0.5458 {[}0.5297, 0.5610{]} & 0.5000 / 0.6008 & 0.5373 \\
Raster2Seq & Angle & 0.3681 {[}0.3549, 0.3809{]} & 0.3366 / 0.4062 & 0.3626 \\
\end{longtable}
\endgroup

\Needspace{9\baselineskip}
\noindent\begin{minipage}{\linewidth}\captionof{table}{Fixed-tolerance GT accuracy and shared-decoder control-point costs.}\label{tab:C.13}\end{minipage}\addtocounter{table}{-1}

\begingroup
\small
\begin{longtable}[]{@{}
  >{\raggedright\arraybackslash}p{(\linewidth - 8\tabcolsep) * \real{0.2000}}
  >{\raggedright\arraybackslash}p{(\linewidth - 8\tabcolsep) * \real{0.2000}}
  >{\raggedright\arraybackslash}p{(\linewidth - 8\tabcolsep) * \real{0.2000}}
  >{\raggedright\arraybackslash}p{(\linewidth - 8\tabcolsep) * \real{0.2000}}
  >{\raggedright\arraybackslash}p{(\linewidth - 8\tabcolsep) * \real{0.2000}}@{}}
\toprule\noalign{}
\begin{minipage}[b]{\linewidth}\raggedright
Method
\end{minipage} & \begin{minipage}[b]{\linewidth}\raggedright
Tolerance (px)
\end{minipage} & \begin{minipage}[b]{\linewidth}\raggedright
GT mIoU {[}95\% BCa CI{]}
\end{minipage} & \begin{minipage}[b]{\linewidth}\raggedright
Median vertices
\end{minipage} & \begin{minipage}[b]{\linewidth}\raggedright
Completed
\end{minipage} \\
\midrule\noalign{}
\endhead
\bottomrule\noalign{}
\endlastfoot
sali\_fp & 0 & 0.3596 {[}0.3449, 0.3740{]} & 3,869.0 & 400/400 \\
sali\_fp & 0.25 & 0.3596 {[}0.3449, 0.3740{]} & 3,869.0 & 400/400 \\
sali\_fp & 0.5 & 0.3596 {[}0.3449, 0.3740{]} & 3,641.0 & 400/400 \\
sali\_fp & 1 & 0.3613 {[}0.3464, 0.3757{]} & 1,916.5 & 400/400 \\
sali\_fp & 2 & 0.3612 {[}0.3463, 0.3756{]} & 1,734.0 & 400/400 \\
sali\_fp & 4 & 0.3609 {[}0.3461, 0.3753{]} & 1,675.0 & 400/400 \\
sali\_fp & 8 & 0.3559 {[}0.3415, 0.3702{]} & 1,618.5 & 400/400 \\
cubicasa5k & 0 & 0.7390 {[}0.7270, 0.7496{]} & 6,268.0 & 400/400 \\
cubicasa5k & 0.25 & 0.7390 {[}0.7270, 0.7496{]} & 6,268.0 & 400/400 \\
cubicasa5k & 0.5 & 0.7390 {[}0.7270, 0.7496{]} & 5,785.0 & 400/400 \\
cubicasa5k & 1 & 0.7384 {[}0.7264, 0.7489{]} & 1,725.0 & 400/400 \\
cubicasa5k & 2 & 0.7327 {[}0.7210, 0.7433{]} & 1,172.5 & 400/400 \\
cubicasa5k & 4 & 0.7105 {[}0.6990, 0.7206{]} & 884.0 & 400/400 \\
cubicasa5k & 8 & 0.6674 {[}0.6564, 0.6773{]} & 792.0 & 400/400 \\
cubicasa\_polygon & 0 & 0.6612 {[}0.6429, 0.6773{]} & 246.0 & 395/400 \\
cubicasa\_polygon & 0.25 & 0.6612 {[}0.6429, 0.6773{]} & 246.0 & 395/400 \\
cubicasa\_polygon & 0.5 & 0.6612 {[}0.6429, 0.6773{]} & 246.0 & 395/400 \\
cubicasa\_polygon & 1 & 0.6612 {[}0.6429, 0.6773{]} & 229.0 & 395/400 \\
cubicasa\_polygon & 2 & 0.6611 {[}0.6428, 0.6772{]} & 222.0 & 395/400 \\
cubicasa\_polygon & 4 & 0.6607 {[}0.6424, 0.6768{]} & 210.0 & 395/400 \\
cubicasa\_polygon & 8 & 0.6440 {[}0.6262, 0.6598{]} & 187.0 & 395/400 \\
\end{longtable}
\endgroup

\begin{figure}[!htbp]
  \centering
  \includegraphics[width=\textwidth,height=0.82\textheight,keepaspectratio]{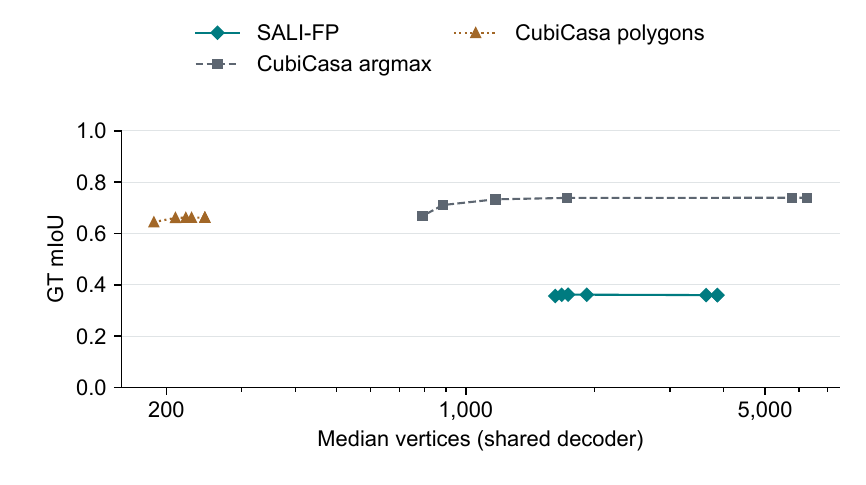}
  \caption{GT accuracy versus shared-decoder control-point cost at the seven fixed simplification tolerances in Table~\ref{tab:C.13}. Every setting is applied to all test plans; no per-plan selection uses GT. Markers denote the frozen tolerance settings.}
  \label{fig:C.3}
\end{figure}
\FloatBarrier

\Needspace{9\baselineskip}
\noindent\begin{minipage}{\linewidth}\captionof{table}{Additional production, stage, and decision audit.}\label{tab:C.14}\end{minipage}\addtocounter{table}{-1}

\begingroup
\small
\begin{longtable}[]{@{}
  >{\raggedright\arraybackslash}p{(\linewidth - 4\tabcolsep) * \real{0.3333}}
  >{\raggedright\arraybackslash}p{(\linewidth - 4\tabcolsep) * \real{0.3333}}
  >{\raggedright\arraybackslash}p{(\linewidth - 4\tabcolsep) * \real{0.3333}}@{}}
\toprule\noalign{}
\begin{minipage}[b]{\linewidth}\raggedright
Observation
\end{minipage} & \begin{minipage}[b]{\linewidth}\raggedright
Result
\end{minipage} & \begin{minipage}[b]{\linewidth}\raggedright
Scope
\end{minipage} \\
\midrule\noalign{}
\endhead
\bottomrule\noalign{}
\endlastfoot
Plans with extra HTTP attempts & 1,591 / 11,534 & Per-case request counters \\
Additional HTTP attempts / maximum per plan & 2,056 / 6 & Not extra logical repairs \\
Repair-triggered / skipped public plans & 263 / 137 & Same 400 test plans \\
Already-target threshold 0.56 & 350 / 3760 & One nonconstruction predicate only \\
Already-target threshold 0.7 & 200 / 3760 & One nonconstruction predicate only \\
Already-target threshold 0.84 & 64 / 3760 & One nonconstruction predicate only \\
Independent plan tasks / proposal tasks & 200 / 400 & No completed human scores \\
Accepted source-only affine transforms & 356/400 & Not GT displacement estimates \\
Absolute horizontal translation (px) & 13.061 {[}3.095, 36.444{]} & Median {[}IQR{]}, all 400 plans \\
Absolute vertical translation (px) & 57.132 {[}18.054, 106.558{]} & Median {[}IQR{]}, all 400 plans \\
Minimum affine scale & 0.934 {[}0.873, 0.976{]} & Median {[}IQR{]}, all 400 plans \\
Maximum affine scale & 0.984 {[}0.946, 1.000{]} & Median {[}IQR{]}, all 400 plans \\
\end{longtable}
\endgroup

Native room scores in Table~\ref{tab:C.12} use 398 SVGs containing room references; cubicasa5k\_test\_0048 and cubicasa5k\_test\_0219 have no Space polygons. They remain in all pixel and inference records. Method failures retain the eligible reference denominator. Angle F1 follows the official angle criterion and is not an edge score. Shared raster-component contours are not substituted for native SALI-FP sparse room objects.

The fixed-tolerance curve reports each configuration on every plan, including failed polygon decodes as empty raster predictions. Vertex medians refer to completed representations; completed counts are shown rather than assigning failures zero storage. Gate replay changes one predicate on retained statistics and does not regenerate repair outputs or measure edit correctness.

\Needspace{9\baselineskip}
\noindent\begin{minipage}{\linewidth}\captionof{table}{Native room costs and common overlap-filter sensitivity.}\label{tab:C.15}\end{minipage}\addtocounter{table}{-1}

\begingroup
\small
\begin{longtable}[]{@{}
  >{\raggedright\arraybackslash}p{(\linewidth - 10\tabcolsep) * \real{0.1667}}
  >{\raggedright\arraybackslash}p{(\linewidth - 10\tabcolsep) * \real{0.1667}}
  >{\raggedright\arraybackslash}p{(\linewidth - 10\tabcolsep) * \real{0.1667}}
  >{\raggedright\arraybackslash}p{(\linewidth - 10\tabcolsep) * \real{0.1667}}
  >{\raggedright\arraybackslash}p{(\linewidth - 10\tabcolsep) * \real{0.1667}}
  >{\raggedright\arraybackslash}p{(\linewidth - 10\tabcolsep) * \real{0.1667}}@{}}
\toprule\noalign{}
\begin{minipage}[b]{\linewidth}\raggedright
Method
\end{minipage} & \begin{minipage}[b]{\linewidth}\raggedright
Completed
\end{minipage} & \begin{minipage}[b]{\linewidth}\raggedright
Cost n
\end{minipage} & \begin{minipage}[b]{\linewidth}\raggedright
Vertices {[}IQR{]}
\end{minipage} & \begin{minipage}[b]{\linewidth}\raggedright
KiB {[}IQR{]}
\end{minipage} & \begin{minipage}[b]{\linewidth}\raggedright
Filtered room F1
\end{minipage} \\
\midrule\noalign{}
\endhead
\bottomrule\noalign{}
\endlastfoot
SALI-FP & 400/400 & 398 & 294 {[}199, 431{]} & 5.219 {[}3.596, 7.622{]} & 0.4443 \\
CubiCasa\allowbreak 5K / polygons & 395/400 & 393 & 132 {[}92, 188{]} & 2.266 {[}1.619, 3.136{]} & 0.6273 \\
Raster2Seq & 400/400 & 398 & 70 {[}52, 103.75{]} & 1.457 {[}1.075, 2.086{]} & 0.7018 \\
\end{longtable}
\endgroup

Native costs retain exterior and hole vertices without a repeated closing point. Compact UTF-8 JSON uses common fields, room class, 1024-canvas coordinates, and two decimal places. Cost summaries exclude failed inference and absent room GT, with their counts explicit; accuracy keeps failed predictions as zero. All three methods use the same primary and sensitivity overlap switches.

\Needspace{9\baselineskip}
\noindent\begin{minipage}{\linewidth}\captionof{table}{Stages under one frozen final source-only transform.}\label{tab:C.16}\end{minipage}\addtocounter{table}{-1}

\begingroup
\small
\begin{longtable}[]{@{}
  >{\raggedright\arraybackslash}p{(\linewidth - 8\tabcolsep) * \real{0.2200}}
  >{\raggedright\arraybackslash}p{(\linewidth - 8\tabcolsep) * \real{0.0600}}
  >{\raggedright\arraybackslash}p{(\linewidth - 8\tabcolsep) * \real{0.2200}}
  >{\raggedright\arraybackslash}p{(\linewidth - 8\tabcolsep) * \real{0.3400}}
  >{\raggedright\arraybackslash}p{(\linewidth - 8\tabcolsep) * \real{0.1600}}@{}}
\toprule\noalign{}
\begin{minipage}[b]{\linewidth}\raggedright
Subset
\end{minipage} & \begin{minipage}[b]{\linewidth}\raggedright
n
\end{minipage} & \begin{minipage}[b]{\linewidth}\raggedright
Stage
\end{minipage} & \begin{minipage}[b]{\linewidth}\raggedright
Pooled mIoU {[}95\% BCa CI{]}
\end{minipage} & \begin{minipage}[b]{\linewidth}\raggedright
BIoU
\end{minipage} \\
\midrule\noalign{}
\endhead
\bottomrule\noalign{}
\endlastfoot
all & 400 & initial & 0.2763 {[}0.2653, 0.2885{]} & 0.1152 \\
all & 400 & protected repair & 0.3159 {[}0.3028, 0.3295{]} & 0.1515 \\
all & 400 & cleanup & 0.3596 {[}0.3449, 0.3740{]} & 0.1858 \\
repair triggered & 263 & initial & 0.2410 {[}0.2315, 0.2519{]} & 0.0849 \\
repair triggered & 263 & protected repair & 0.2984 {[}0.2833, 0.3144{]} & 0.1402 \\
repair triggered & 263 & cleanup & 0.3384 {[}0.3213, 0.3553{]} & 0.1708 \\
repair skipped & 137 & initial & 0.3521 {[}0.3270, 0.3774{]} & 0.1734 \\
repair skipped & 137 & protected repair & 0.3521 {[}0.3270, 0.3774{]} & 0.1734 \\
repair skipped & 137 & cleanup & 0.4040 {[}0.3769, 0.4297{]} & 0.2146 \\
\end{longtable}
\endgroup

The repair-triggered fixed-frame mean per-plan mIoU change is 0.0702 {[}0.0574, 0.0842{]}. These plan-mean contrasts and pooled mIoU differences use different aggregation; all per-metric paired intervals are retained in the evidence table. The original-coordinate contrasts remain in Table~\ref{tab:C.11}. No stage-specific refitting or GT-derived selection is used.

\begin{figure}[!htbp]
  \centering
  \includegraphics[width=\textwidth,height=0.82\textheight,keepaspectratio]{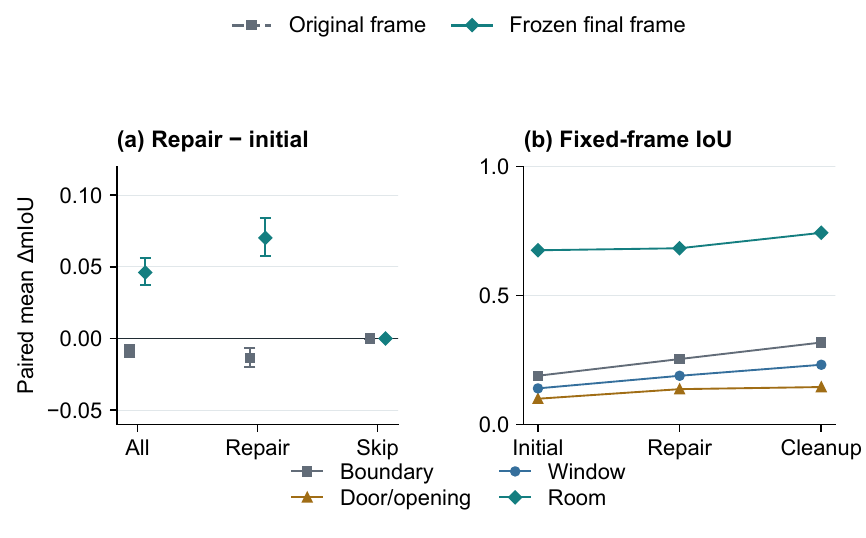}
  \caption{Stage-detail audit: (a) paired plan-mean repair-minus-initial mIoU differences with 95\% BCa intervals; (b) class IoU under the frozen final transform. Repair-skipped images are identical before cleanup. Frame-conditioned results and group counts are in Table~\ref{tab:C.16}.}
  \label{fig:C.4}
\end{figure}
\FloatBarrier

\Needspace{9\baselineskip}
\noindent\begin{minipage}{\linewidth}\captionof{table}{Exclusive pixel errors; parentheses are percentages of all errors.}\label{tab:C.17}\end{minipage}\addtocounter{table}{-1}

\begingroup
\small
\begin{longtable}[]{@{}
  >{\raggedright\arraybackslash}p{(\linewidth - 12\tabcolsep) * \real{0.1800}}
  >{\raggedright\arraybackslash}p{(\linewidth - 12\tabcolsep) * \real{0.0800}}
  >{\raggedright\arraybackslash}p{(\linewidth - 12\tabcolsep) * \real{0.1200}}
  >{\raggedright\arraybackslash}p{(\linewidth - 12\tabcolsep) * \real{0.1550}}
  >{\raggedright\arraybackslash}p{(\linewidth - 12\tabcolsep) * \real{0.1550}}
  >{\raggedright\arraybackslash}p{(\linewidth - 12\tabcolsep) * \real{0.1550}}
  >{\raggedright\arraybackslash}p{(\linewidth - 12\tabcolsep) * \real{0.1550}}@{}}
\toprule\noalign{}
\begin{minipage}[b]{\linewidth}\raggedright
Method
\end{minipage} & \begin{minipage}[b]{\linewidth}\raggedright
Band
\end{minipage} & \begin{minipage}[b]{\linewidth}\raggedright
Errors
\end{minipage} & \begin{minipage}[b]{\linewidth}\raggedright
Boundary
\end{minipage} & \begin{minipage}[b]{\linewidth}\raggedright
Omitted
\end{minipage} & \begin{minipage}[b]{\linewidth}\raggedright
Added
\end{minipage} & \begin{minipage}[b]{\linewidth}\raggedright
Confused
\end{minipage} \\
\midrule\noalign{}
\endhead
\bottomrule\noalign{}
\endlastfoot
SALI-FP & 0.25\% & 56,350,103 & 19,719,198 (34.99\%) & 12,199,963 (21.65\%) & 17,045,832 (30.25\%) & 7,385,110 (13.11\%) \\
SALI-FP & 0.5\% & 56,350,103 & 25,050,419 (44.45\%) & 10,094,113 (17.91\%) & 16,115,423 (28.60\%) & 5,090,148 (9.03\%) \\
SALI-FP & 1\% & 56,350,103 & 31,796,315 (56.43\%) & 7,180,240 (12.74\%) & 14,375,243 (25.51\%) & 2,998,305 (5.32\%) \\
CubiCasa\allowbreak 5K / argmax & 0.25\% & 20,259,975 & 7,341,614 (36.24\%) & 4,223,581 (20.85\%) & 7,850,721 (38.75\%) & 844,059 (4.17\%) \\
CubiCasa\allowbreak 5K / argmax & 0.5\% & 20,259,975 & 8,266,735 (40.80\%) & 3,842,951 (18.97\%) & 7,608,690 (37.56\%) & 541,599 (2.67\%) \\
CubiCasa\allowbreak 5K / argmax & 1\% & 20,259,975 & 9,604,690 (47.41\%) & 3,222,779 (15.91\%) & 7,140,895 (35.25\%) & 291,611 (1.44\%) \\
CubiCasa\allowbreak 5K / polygons & 0.25\% & 28,911,667 & 10,312,167 (35.67\%) & 9,579,585 (33.13\%) & 7,206,143 (24.92\%) & 1,813,772 (6.27\%) \\
CubiCasa\allowbreak 5K / polygons & 0.5\% & 28,911,667 & 12,545,367 (43.39\%) & 8,440,341 (29.19\%) & 6,912,552 (23.91\%) & 1,013,407 (3.51\%) \\
CubiCasa\allowbreak 5K / polygons & 1\% & 28,911,667 & 15,533,958 (53.73\%) & 6,659,522 (23.03\%) & 6,370,200 (22.03\%) & 347,987 (1.20\%) \\
\end{longtable}
\endgroup

Every method retains 419,430,400 pixels (400 square 1024-pixel canvases). The hierarchy assigns each wrong pixel first to the GT-boundary neighborhood, otherwise to foreground-to-background omission, background-to-foreground addition, or foreground-class confusion. The band is the rounded diagonal fraction around both sides of adjacent GT-label changes. Five-class pooled and GT-row-normalized confusion matrices, per-class decomposition, and failed-case records accompany the table.

\Needspace{9\baselineskip}
\noindent\begin{minipage}{\linewidth}\captionof{table}{Exclusive geometric causes before and after delivery.}\label{tab:C.18}\end{minipage}\addtocounter{table}{-1}

\begingroup
\small
\begin{longtable}[]{@{}lll@{}}
\toprule\noalign{}
Geometric cause & Original objects & Delivered objects \\
\midrule\noalign{}
\endhead
\bottomrule\noalign{}
\endlastfoot
empty & 7765 & 7765 \\
hole or ring structure & 4 & 0 \\
malformed & 6 & 6 \\
other invalid & 90 & 11 \\
self intersection & 101478 & 22935 \\
valid & 673880 & 752510 \\
zero area or collapsed & 76 & 72 \\
\end{longtable}
\endgroup

Classification prioritizes malformed, empty, nonfinite, zero-area/collapsed, valid, self-intersection, hole/ring-structure, too-few-point, and other-invalid conditions. The raw GEOS reason remains attached to each object. Repair dispositions in Table~\ref{tab:B.1} are independent processing outcomes; neither is inferred from the other. Residual IDs are excluded from automatic solid construction pending review.

\begin{figure}[!htbp]
  \centering
  \includegraphics[width=\textwidth,height=0.82\textheight,keepaspectratio]{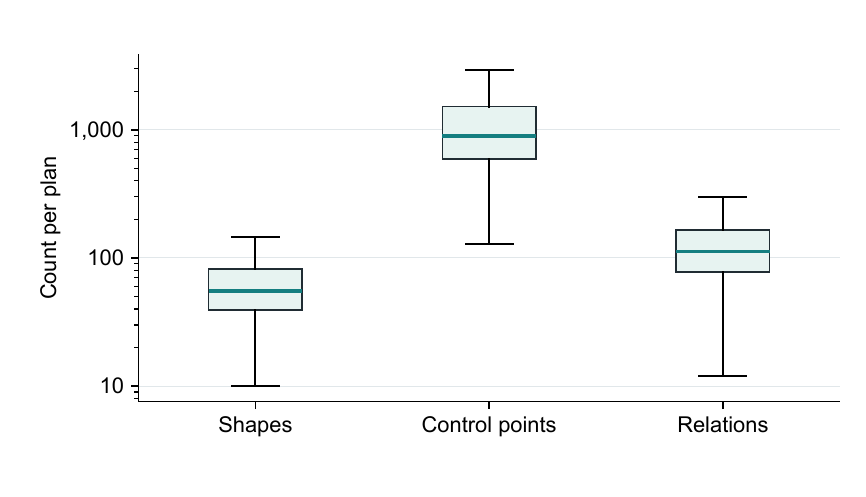}
  \caption{ArchP10k representation sizes per plan. Boxes span the interquartile range, center lines mark medians, and whiskers extend to the furthest observations within 1.5 IQR. Outliers are omitted only from this display; all plans remain in the statistics. The count axis is logarithmic; related export statistics are in Table~\ref{tab:C.6}.}
  \label{fig:C.5}
\end{figure}
\FloatBarrier

\Needspace{9\baselineskip}
\noindent\begin{minipage}{\linewidth}\captionof{table}{Retained local baseline timers; seconds per attempted plan.}\label{tab:C.19}\end{minipage}\addtocounter{table}{-1}

\begingroup
\small
\begin{longtable}[]{@{}llll@{}}
\toprule\noalign{}
Local method & n & Median (s) & IQR (s) \\
\midrule\noalign{}
\endhead
\bottomrule\noalign{}
\endlastfoot
CubiCasa\allowbreak 5K / argmax & 400 & 0.380 & 0.368--0.388 \\
CubiCasa\allowbreak 5K / polygons & 400 & 6.804 & 6.589--7.301 \\
MiT-UNet & 400 & 0.050 & 0.049--0.052 \\
Raster2Seq & 400 & 1.153 & 0.866--1.728 \\
\end{longtable}
\endgroup

Timers exclude model loading and include each retained per-case inference/export routine; polygon timers additionally include official postprocessing. Apple Silicon local CPU/MPS execution is distinct from the unobservable remote image/audit accelerator. These entries are not end-to-end deployment benchmarks. No baseline monetary bill or unmeasured SALI-FP stage is imputed; remote-stage records and reference-price estimates remain in Table~\ref{tab:C.3}.

\Needspace{9\baselineskip}
\noindent\begin{minipage}{\linewidth}\captionof{table}{Delivered geometric causes by semantic class.}\label{tab:C.20}\end{minipage}\addtocounter{table}{-1}

\begingroup
\small
\begin{longtable}[]{@{}llllll@{}}
\toprule\noalign{}
Class & Objects & Self-cross & Empty & Other invalid & Invalid share \\
\midrule\noalign{}
\endhead
\bottomrule\noalign{}
\endlastfoot
balcony & 15,524 & 296 & 160 & 1 & 2.94\% \\
bathroom & 32,109 & 261 & 11 & 0 & 0.85\% \\
bedroom & 49,056 & 2,839 & 1,582 & 12 & 9.04\% \\
corridor & 22,345 & 2,044 & 779 & 8 & 12.67\% \\
door & 160,745 & 216 & 148 & 1 & 0.23\% \\
elevator & 10,389 & 104 & 63 & 0 & 1.61\% \\
entrance & 5,528 & 83 & 10 & 0 & 1.68\% \\
kitchen & 15,098 & 130 & 100 & 0 & 1.52\% \\
living room & 18,960 & 306 & 31 & 1 & 1.78\% \\
room default & 87,979 & 8,385 & 1,659 & 23 & 11.44\% \\
stair & 34,671 & 707 & 167 & 3 & 2.53\% \\
storage & 7,053 & 264 & 30 & 5 & 4.24\% \\
study & 4,565 & 1,011 & 138 & 11 & 25.41\% \\
wall & 177,097 & 3,059 & 577 & 7 & 2.06\% \\
window & 142,180 & 3,230 & 2,310 & 17 & 3.91\% \\
\end{longtable}
\endgroup

\Needspace{9\baselineskip}
\noindent\begin{minipage}{\linewidth}\captionof{table}{Building categories with the most residual invalid objects; remainder pooled.}\label{tab:C.21}\end{minipage}\addtocounter{table}{-1}

\begingroup
\small
\begin{longtable}[]{@{}
  >{\raggedright\arraybackslash}p{(\linewidth - 10\tabcolsep) * \real{0.1667}}
  >{\raggedright\arraybackslash}p{(\linewidth - 10\tabcolsep) * \real{0.1667}}
  >{\raggedright\arraybackslash}p{(\linewidth - 10\tabcolsep) * \real{0.1667}}
  >{\raggedright\arraybackslash}p{(\linewidth - 10\tabcolsep) * \real{0.1667}}
  >{\raggedright\arraybackslash}p{(\linewidth - 10\tabcolsep) * \real{0.1667}}
  >{\raggedright\arraybackslash}p{(\linewidth - 10\tabcolsep) * \real{0.1667}}@{}}
\toprule\noalign{}
\begin{minipage}[b]{\linewidth}\raggedright
Source category
\end{minipage} & \begin{minipage}[b]{\linewidth}\raggedright
Objects
\end{minipage} & \begin{minipage}[b]{\linewidth}\raggedright
Self-cross
\end{minipage} & \begin{minipage}[b]{\linewidth}\raggedright
Empty
\end{minipage} & \begin{minipage}[b]{\linewidth}\raggedright
Other invalid
\end{minipage} & \begin{minipage}[b]{\linewidth}\raggedright
Invalid share
\end{minipage} \\
\midrule\noalign{}
\endhead
\bottomrule\noalign{}
\endlastfoot
residential other & 440,079 & 13,684 & 4,867 & 47 & 4.23\% \\
office & 51,837 & 1,417 & 446 & 7 & 3.61\% \\
interior general & 38,446 & 1,218 & 429 & 7 & 4.30\% \\
hotel & 38,203 & 1,317 & 308 & 7 & 4.27\% \\
education & 52,371 & 1,200 & 317 & 5 & 2.91\% \\
cultural other & 45,099 & 966 & 387 & 6 & 3.01\% \\
mixed use & 25,715 & 724 & 220 & 3 & 3.68\% \\
restaurant bar & 14,085 & 450 & 131 & 1 & 4.13\% \\
all remaining categories & 77,464 & 1,959 & 660 & 6 & 3.39\% \\
\end{longtable}
\endgroup

Tables C.20--C.21 retain every object. Other-invalid counts combine zero-area, nonfinite, malformed, hole/ring and other GEOS causes. The pooled remainder is not excluded; the full building-category cross-tabulation remains in the evidence files.

\Needspace{9\baselineskip}
\noindent\begin{minipage}{\linewidth}\captionof{table}{Door/window error locations; percentages of the respective GT-class pixels.}\label{tab:C.22}\end{minipage}\addtocounter{table}{-1}

\begingroup
\small
\begin{longtable}[]{@{}
  >{\raggedright\arraybackslash}p{(\linewidth - 10\tabcolsep) * \real{0.1667}}
  >{\raggedright\arraybackslash}p{(\linewidth - 10\tabcolsep) * \real{0.1667}}
  >{\raggedright\arraybackslash}p{(\linewidth - 10\tabcolsep) * \real{0.1667}}
  >{\raggedright\arraybackslash}p{(\linewidth - 10\tabcolsep) * \real{0.1667}}
  >{\raggedright\arraybackslash}p{(\linewidth - 10\tabcolsep) * \real{0.1667}}
  >{\raggedright\arraybackslash}p{(\linewidth - 10\tabcolsep) * \real{0.1667}}@{}}
\toprule\noalign{}
\begin{minipage}[b]{\linewidth}\raggedright
Method
\end{minipage} & \begin{minipage}[b]{\linewidth}\raggedright
GT class
\end{minipage} & \begin{minipage}[b]{\linewidth}\raggedright
GT pixels
\end{minipage} & \begin{minipage}[b]{\linewidth}\raggedright
Boundary error
\end{minipage} & \begin{minipage}[b]{\linewidth}\raggedright
Interior omitted
\end{minipage} & \begin{minipage}[b]{\linewidth}\raggedright
Interior confused
\end{minipage} \\
\midrule\noalign{}
\endhead
\bottomrule\noalign{}
\endlastfoot
SALI-FP & Door/opening & 2,162,846 & 65.99\% & 1.38\% & 4.23\% \\
SALI-FP & Window & 4,708,442 & 55.19\% & 4.48\% & 7.66\% \\
CubiCasa\allowbreak 5K / argmax & Door/opening & 2,162,846 & 29.63\% & 0.08\% & 1.68\% \\
CubiCasa\allowbreak 5K / argmax & Window & 4,708,442 & 18.84\% & 0.17\% & 1.37\% \\
CubiCasa\allowbreak 5K / polygons & Door/opening & 2,162,846 & 30.62\% & 0.70\% & 2.44\% \\
CubiCasa\allowbreak 5K / polygons & Window & 4,708,442 & 25.18\% & 2.70\% & 4.10\% \\
\end{longtable}
\endgroup

Correct pixels occupy the remaining GT support. Background-to-foreground additions are tabulated in Table~\ref{tab:C.17}, not assigned to an invented door/window GT instance. Spatial error categories do not identify a unique generative cause.

\Needspace{9\baselineskip}
\noindent\begin{minipage}{\linewidth}\captionof{table}{Reference-scope sensitivity; the primary evaluation retains all 400 plans.}\label{tab:C.23}\end{minipage}\addtocounter{table}{-1}

\begingroup
\small
\begin{longtable}[]{@{}
  >{\raggedright\arraybackslash}p{(\linewidth - 8\tabcolsep) * \real{0.2000}}
  >{\raggedright\arraybackslash}p{(\linewidth - 8\tabcolsep) * \real{0.2000}}
  >{\raggedright\arraybackslash}p{(\linewidth - 8\tabcolsep) * \real{0.2000}}
  >{\raggedright\arraybackslash}p{(\linewidth - 8\tabcolsep) * \real{0.2000}}
  >{\raggedright\arraybackslash}p{(\linewidth - 8\tabcolsep) * \real{0.2000}}@{}}
\toprule\noalign{}
\begin{minipage}[b]{\linewidth}\raggedright
Scope
\end{minipage} & \begin{minipage}[b]{\linewidth}\raggedright
Plans
\end{minipage} & \begin{minipage}[b]{\linewidth}\raggedright
Method
\end{minipage} & \begin{minipage}[b]{\linewidth}\raggedright
mIoU
\end{minipage} & \begin{minipage}[b]{\linewidth}\raggedright
Added pixels
\end{minipage} \\
\midrule\noalign{}
\endhead
\bottomrule\noalign{}
\endlastfoot
complete test & 400 & SALI-FP & 0.3596 & 16,115,423 \\
complete test & 400 & CubiCasa\allowbreak 5K / argmax & 0.7390 & 7,608,690 \\
complete test & 400 & CubiCasa\allowbreak 5K / polygons & 0.6612 & 6,912,552 \\
nonempty room reference & 398 & SALI-FP & 0.3603 & 15,362,301 \\
nonempty room reference & 398 & CubiCasa\allowbreak 5K / argmax & 0.7409 & 6,804,133 \\
nonempty room reference & 398 & CubiCasa\allowbreak 5K / polygons & 0.6628 & 6,081,348 \\
\end{longtable}
\endgroup

Test 0048 and Test 0219 have no SVG Space references. The apparent room additions in Test 0048 therefore expose annotation scope as well as prediction disagreement; they are not verified hallucinations. Table~\ref{tab:C.23} separately pools the 398 nonempty-room references without changing the official 400-case primary denominator.

\Needspace{9\baselineskip}
\noindent\begin{minipage}{\linewidth}\captionof{table}{Public-test benchmark calibration and native room geometry.}\label{tab:C.24}\end{minipage}\addtocounter{table}{-1}

\begingroup
\small
\begin{longtable}[]{@{}
  >{\raggedright\arraybackslash}p{(\linewidth - 8\tabcolsep) * \real{0.2000}}
  >{\raggedright\arraybackslash}p{(\linewidth - 8\tabcolsep) * \real{0.2000}}
  >{\raggedright\arraybackslash}p{(\linewidth - 8\tabcolsep) * \real{0.2000}}
  >{\raggedright\arraybackslash}p{(\linewidth - 8\tabcolsep) * \real{0.2000}}
  >{\raggedright\arraybackslash}p{(\linewidth - 8\tabcolsep) * \real{0.2000}}@{}}
\toprule\noalign{}
\begin{minipage}[b]{\linewidth}\raggedright
Method
\end{minipage} & \begin{minipage}[b]{\linewidth}\raggedright
Four-class mIoU {[}95\% BCa CI{]}
\end{minipage} & \begin{minipage}[b]{\linewidth}\raggedright
Room F1
\end{minipage} & \begin{minipage}[b]{\linewidth}\raggedright
Corner F1
\end{minipage} & \begin{minipage}[b]{\linewidth}\raggedright
Angle F1
\end{minipage} \\
\midrule\noalign{}
\endhead
\bottomrule\noalign{}
\endlastfoot
SALI-FP & 0.3596 {[}0.3449, 0.3740{]} & 0.445 & 0.159 & 0.093 \\
CubiCasa\allowbreak 5K / argmax & 0.7390 {[}0.7270, 0.7496{]} & N/A & N/A & N/A \\
CubiCasa\allowbreak 5K / polygons & 0.6612 {[}0.6422, 0.6766{]} & 0.630 & 0.327 & 0.236 \\
Raster2Seq & N/A & 0.763 & 0.546 & 0.368 \\
\end{longtable}
\endgroup

Pixel calibration uses all 400 test plans; native room geometry uses the 398 plans with SVG room references. Native polygon methods and raster argmax are distinguished explicitly. Raster2Seq has no directly comparable four-class mask in this evaluation, and MiT-UNet remains in the wall-only calibration (Table~\ref{tab:C.9}). These values establish strict annotation-aligned calibration; they do not substitute for the structured-output and matched visual evidence reported in the main text and Appendix F.

\begin{figure}[!htbp]
  \centering
  \includegraphics[width=\textwidth,height=0.82\textheight,keepaspectratio]{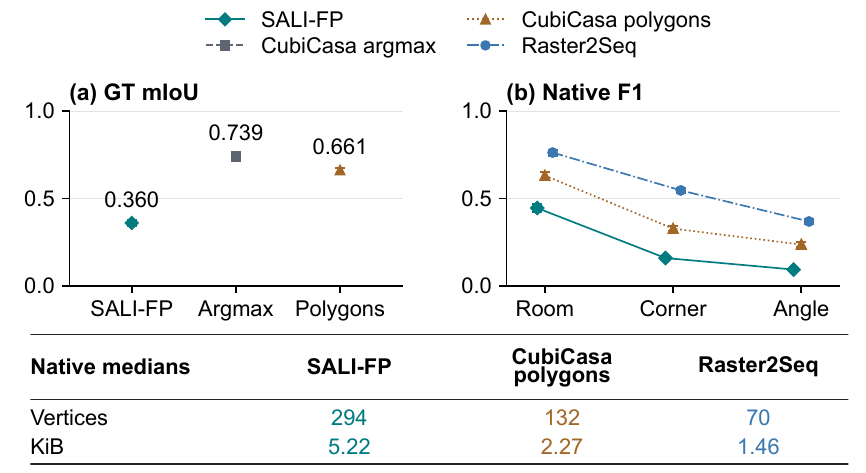}
  \caption{CubiCasa\allowbreak 5K benchmark calibration: (a) GT mIoU; (b) native room, corner, and angle F1 with 95\% BCa intervals. The lower band reports native room medians; IQRs are in Table~\ref{tab:C.15}. Fixed-tolerance curves are in Fig. C.3.}
  \label{fig:C.6}
\end{figure}
\FloatBarrier

\Needspace{9\baselineskip}
\noindent\begin{minipage}{\linewidth}\captionof{table}{Recorded SALI-FP stages on the same 400 public test plans.}\label{tab:C.25}\end{minipage}\addtocounter{table}{-1}

\begingroup
\small
\begin{longtable}[]{@{}llll@{}}
\toprule\noalign{}
Stage & mIoU {[}95\% BCa CI{]} & BIoU & Boundary F1 \\
\midrule\noalign{}
\endhead
\bottomrule\noalign{}
\endlastfoot
Initial \(S_0\) & 0.2099 {[}0.2024, 0.2193{]} & 0.0642 & 0.2358 \\
Protected \(S_1\) & 0.2015 {[}0.1942, 0.2101{]} & 0.0610 & 0.2260 \\
Cleanup labels & 0.2074 {[}0.2000, 0.2162{]} & 0.0570 & 0.2100 \\
Final representation & 0.3596 {[}0.3449, 0.3740{]} & 0.1858 & 0.4962 \\
\end{longtable}
\endgroup

\begin{figure}[!htbp]
  \centering
  \includegraphics[width=\textwidth,height=0.82\textheight,keepaspectratio]{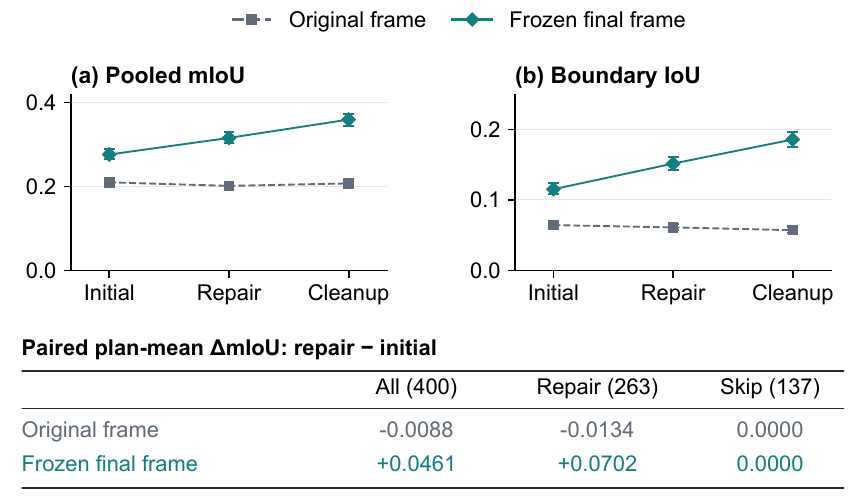}
  \caption{Recorded stages in original and frozen final coordinates: pooled mIoU and boundary IoU with 95\% BCa intervals. The band reports plan-mean repair-minus-initial differences; complete paired intervals and class trajectories are in Fig. C.4.}
  \label{fig:C.7}
\end{figure}
\FloatBarrier

\begin{figure}[!htbp]
  \centering
  \includegraphics[width=\textwidth,height=0.82\textheight,keepaspectratio]{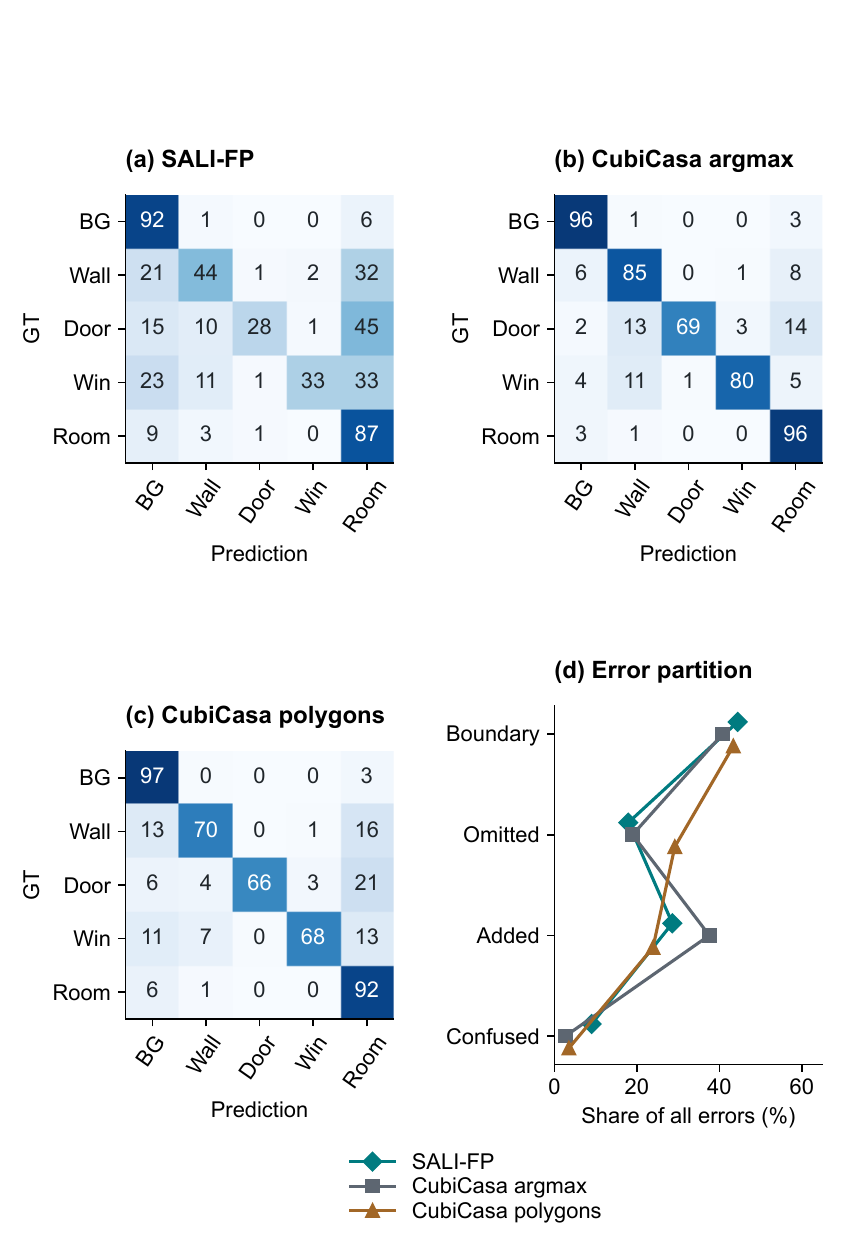}
  \caption{Systematic pixel-error diagnostics: SALI-FP, CubiCasa\allowbreak 5K argmax, and CubiCasa\allowbreak 5K polygons, with GT-row-normalized confusion and the disjoint 0.5\%-diagonal error decomposition.}
  \label{fig:C.8}
\end{figure}
\FloatBarrier

\section{Complexity definition}\label{app:D}

The frozen complexity score averages five standardized image factors: ink fraction, edge density, log connected-component count, orientation entropy, and non-orthogonal edge fraction. Eq.~\eqref{eq:D1} describes this operational score. Standardization and hard-subset selection use the frozen registry, not SALI-FP errors or baseline outputs. The test hard subset contains its 400 highest-ranked plans. Complexity is a visual proxy, not an annotation of architectural difficulty. No PCA or factor-ablation sensitivity result is reported.

\begin{equation}
C_i=\frac{1}{5}\sum_{k=1}^{5}\frac{x_{ik}-\mu_k}{\sigma_k}.
\label{eq:D1}
\end{equation}

\Needspace{9\baselineskip}
\noindent\begin{minipage}{\linewidth}\captionof{table}{Project-cluster Spearman associations; 10,000 BCa resamples.}\label{tab:D.1}\end{minipage}\addtocounter{table}{-1}

\begingroup
\small
\begin{longtable}[]{@{}
  >{\raggedright\arraybackslash}p{(\linewidth - 6\tabcolsep) * \real{0.2500}}
  >{\raggedright\arraybackslash}p{(\linewidth - 6\tabcolsep) * \real{0.2500}}
  >{\raggedright\arraybackslash}p{(\linewidth - 6\tabcolsep) * \real{0.2500}}
  >{\raggedright\arraybackslash}p{(\linewidth - 6\tabcolsep) * \real{0.2500}}@{}}
\toprule\noalign{}
\begin{minipage}[b]{\linewidth}\raggedright
Factor
\end{minipage} & \begin{minipage}[b]{\linewidth}\raggedright
Representation
\end{minipage} & \begin{minipage}[b]{\linewidth}\raggedright
rho {[}95\% CI{]}
\end{minipage} & \begin{minipage}[b]{\linewidth}\raggedright
Holm p
\end{minipage} \\
\midrule\noalign{}
\endhead
\bottomrule\noalign{}
\endlastfoot
ink ratio & auxiliary rcr & 0.072 {[}0.050, 0.094{]} & 0.0012 \\
ink ratio & delivered rcr & 0.030 {[}0.009, 0.051{]} & 0.0141 \\
edge density & auxiliary rcr & -0.093 {[}-0.114, -0.071{]} & 0.0012 \\
edge density & delivered rcr & -0.101 {[}-0.120, -0.080{]} & 0.0012 \\
component count & auxiliary rcr & 0.004 {[}-0.018, 0.027{]} & 1.0000 \\
component count & delivered rcr & 0.002 {[}-0.019, 0.023{]} & 1.0000 \\
orientation entropy & auxiliary rcr & -0.243 {[}-0.263, -0.223{]} & 0.0012 \\
orientation entropy & delivered rcr & -0.186 {[}-0.205, -0.166{]} & 0.0012 \\
nonorthogonal ratio & auxiliary rcr & -0.234 {[}-0.254, -0.212{]} & 0.0012 \\
nonorthogonal ratio & delivered rcr & -0.178 {[}-0.198, -0.158{]} & 0.0012 \\
complexity score & auxiliary rcr & -0.169 {[}-0.189, -0.147{]} & 0.0012 \\
complexity score & delivered rcr & -0.142 {[}-0.162, -0.122{]} & 0.0012 \\
\end{longtable}
\endgroup

\Needspace{9\baselineskip}
\noindent\begin{minipage}{\linewidth}\captionof{table}{Hard (400) minus non-hard test (752), clustered by 730 projects.}\label{tab:D.2}\end{minipage}\addtocounter{table}{-1}

\begingroup
\small
\begin{longtable}[]{@{}
  >{\raggedright\arraybackslash}p{(\linewidth - 8\tabcolsep) * \real{0.2000}}
  >{\raggedright\arraybackslash}p{(\linewidth - 8\tabcolsep) * \real{0.2000}}
  >{\raggedright\arraybackslash}p{(\linewidth - 8\tabcolsep) * \real{0.2000}}
  >{\raggedright\arraybackslash}p{(\linewidth - 8\tabcolsep) * \real{0.2000}}
  >{\raggedright\arraybackslash}p{(\linewidth - 8\tabcolsep) * \real{0.2000}}@{}}
\toprule\noalign{}
\begin{minipage}[b]{\linewidth}\raggedright
Outcome
\end{minipage} & \begin{minipage}[b]{\linewidth}\raggedright
Hard mean
\end{minipage} & \begin{minipage}[b]{\linewidth}\raggedright
Non-hard mean
\end{minipage} & \begin{minipage}[b]{\linewidth}\raggedright
Difference {[}95\% CI{]}
\end{minipage} & \begin{minipage}[b]{\linewidth}\raggedright
Holm p
\end{minipage} \\
\midrule\noalign{}
\endhead
\bottomrule\noalign{}
\endlastfoot
auxiliary rcr & 0.9507 & 0.9582 & -0.0075 {[}-0.0102, -0.0047{]} & 0.0003 \\
delivered rcr & 0.9133 & 0.9324 & -0.0191 {[}-0.0269, -0.0120{]} & 0.0003 \\
protected & 0.2275 & 0.1729 & 0.0546 {[}0.0043, 0.1099{]} & 0.0436 \\
\end{longtable}
\endgroup

Factor intervals resample 7379 projects with replacement and use leave-one-project-out BCa acceleration. Weighted midranks reproduce duplicated observations. Two-sided centered bootstrap tests use a plus-one correction and Holm adjustment over twelve factor-by-representation tests; hard contrasts form a separate three-test family. Component count has the same ranks as its log transform. These associations do not validate complexity against human difficulty judgments.

\section{Availability and reproducibility inventory}\label{app:E}

\Needspace{9\baselineskip}
\noindent\begin{minipage}{\linewidth}\captionof{table}{Official resource status rechecked on 2026-09-07.}\label{tab:E.1}\end{minipage}\addtocounter{table}{-1}

\begingroup
\small
\begin{longtable}[]{@{}
  >{\raggedright\arraybackslash}p{(\linewidth - 4\tabcolsep) * \real{0.3333}}
  >{\raggedright\arraybackslash}p{(\linewidth - 4\tabcolsep) * \real{0.3333}}
  >{\raggedright\arraybackslash}p{(\linewidth - 4\tabcolsep) * \real{0.3333}}@{}}
\toprule\noalign{}
\begin{minipage}[b]{\linewidth}\raggedright
Resource
\end{minipage} & \begin{minipage}[b]{\linewidth}\raggedright
Located material
\end{minipage} & \begin{minipage}[b]{\linewidth}\raggedright
Use in this study
\end{minipage} \\
\midrule\noalign{}
\endhead
\bottomrule\noalign{}
\endlastfoot
CubiCasa\allowbreak 5K \cite{ref02} & Official images, SVG labels, code, trained checkpoint & 400 GT reruns; full-channel polygon rerun; 30 retained illustrations \\
MiT-UNet \cite{ref25} & Official code, trained weights, regional-data link & 400 visible-wall tests; 30 illustrations \\
Raster2Seq \cite{ref27} & Official code and checkpoint & 400 CPU native-polygon inferences; room evaluation on 398 reference cases \\
FloorPlanFormer \cite{ref28} & Training code; validation/test links & No official inference checkpoint located \\
R2V / R3D \cite{ref03,ref05} & Code and test lists; incomplete authorized original inputs & No SALI-FP benchmark values \\
MLStructFP \cite{ref09} & Official dataset documentation and request process & Not downloaded as an unrestricted mirror \\
\end{longtable}
\endgroup

The evidence inventory identifies code commits, checkpoint SHA-256 values, input hashes, mapping rules, dependency versions, and per-case predictions for the reruns. A weight file used for human-pose initialization is not interchangeable with the trained CubiCasa\allowbreak 5K floor-plan checkpoint. Runtime adapters preserve architecture and weights; their fixed-canvas and device differences are disclosed rather than described as exact replication of published timing.

\Needspace{9\baselineskip}
\noindent\begin{minipage}{\linewidth}\captionof{table}{Unfinished validation and acceptance criteria.}\label{tab:E.2}\end{minipage}\addtocounter{table}{-1}

\begingroup
\small
\begin{longtable}[]{@{}
  >{\raggedright\arraybackslash}p{(\linewidth - 4\tabcolsep) * \real{0.3333}}
  >{\raggedright\arraybackslash}p{(\linewidth - 4\tabcolsep) * \real{0.3333}}
  >{\raggedright\arraybackslash}p{(\linewidth - 4\tabcolsep) * \real{0.3333}}@{}}
\toprule\noalign{}
\begin{minipage}[b]{\linewidth}\raggedright
Item
\end{minipage} & \begin{minipage}[b]{\linewidth}\raggedright
Prepared material
\end{minipage} & \begin{minipage}[b]{\linewidth}\raggedright
Required completion
\end{minipage} \\
\midrule\noalign{}
\endhead
\bottomrule\noalign{}
\endlastfoot
ArchP10k accuracy & 200 category/complexity-stratified plans; two blank annotation templates & Independent annotations, adjudication, frozen GT metrics \\
Gate correctness & 200 accepted and 200 rejected proposals, blinded & Independent correctness scores; weighted estimates and agreement \\
Causal stage ablation & Recorded-stage paired evaluation & Controlled reruns holding models, prompts, and inputs fixed \\
Pretraining-exposure diagnostics & Exposure is not auditable from deployment aliases & Frozen near-duplicate and memorization probes with control images; no absence-of-contamination claim \\
Open-weight substitution & Interface and label protocol & Frozen image/VL weights; matched-subset inference and GT evaluation \\
Independent repeatability & Case and prompt identifiers & Multiple uncached model runs; variance and output-identity audit \\
Rhino operator study & Interface examples & Source-linked scale/height, randomized paired operators, timing logs \\
Public release & Private source and rights registry & Permission and privacy audit; approved license and release archive \\
Author disclosures & Names, order, equal contribution, affiliation, and corresponding contact confirmed & CRediT, funding, conflicts, and final author approval remain to be confirmed \\
\end{longtable}
\endgroup

The 200-plan sample uses proportional category-by-complexity allocation with seed 20260907 and prioritizes unique projects. Predictive quality does not enter selection. Images and two empty label templates are supplied independently from predictions. The 400-proposal package hides acceptance labels; private answer keys retain the population weighting. Human annotation, adjudication, and proposal correctness remain incomplete (Table~\ref{tab:E.2}).

\section{Selected complex-plan illustrations}\label{app:F}

We examine 30 purpose-selected complex plans with successful SALI-FP outputs and high internal RCR. This subset supports detailed visual comparison rather than a representative estimate of accuracy. CubiCasa\allowbreak 5K and MiT-UNet use the official checkpoints described in Section 4.2. Each page compares the source with the three methods on a common canvas. SALI-FP is shown at the semantic-output stage, without subsequent registration or geometry repair.

The common display distinguishes structural boundaries, doors/openings, windows and spaces. MiT-UNet predicts walls only; its blank non-wall areas are not room or opening errors. The two detail regions, R1 and R2, use identical coordinates and magnification across all methods. They are selected to explain visible differences, not to estimate overall accuracy. Case-specific observations describe source-visible structure, advantages and residual errors; prediction-derived reference masks and omission counts are not used.

The compact output record reports internal RCR, shape and control-point counts, and the number of invalid polygons in the original sparse representation, before geometry repair. These are SALI-FP output properties rather than cross-method accuracy scores. Corresponding inverse-rendering and internal-disagreement images are included in the reproducibility materials.

\clearpage
\newgeometry{margin=10mm}
\begin{landscape}
\thispagestyle{empty}
\centering
\includegraphics[width=267mm,height=178mm,keepaspectratio]{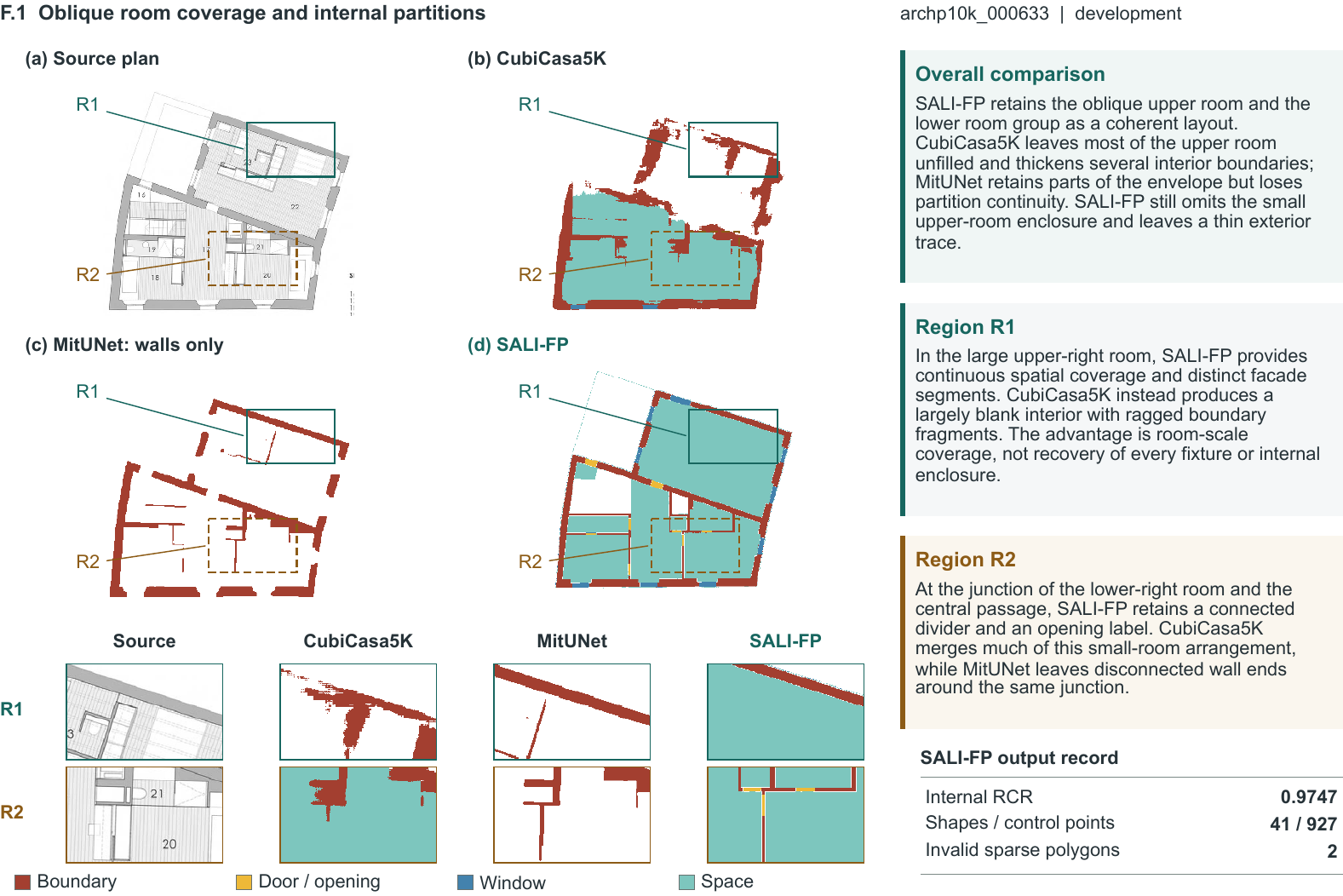}
\captionof{figure}{archp10k\_000633: oblique room coverage and internal partitions.}\label{fig:F.1}
\end{landscape}
\restoregeometry

\clearpage
\newgeometry{margin=10mm}
\begin{landscape}
\thispagestyle{empty}
\centering
\includegraphics[width=267mm,height=178mm,keepaspectratio]{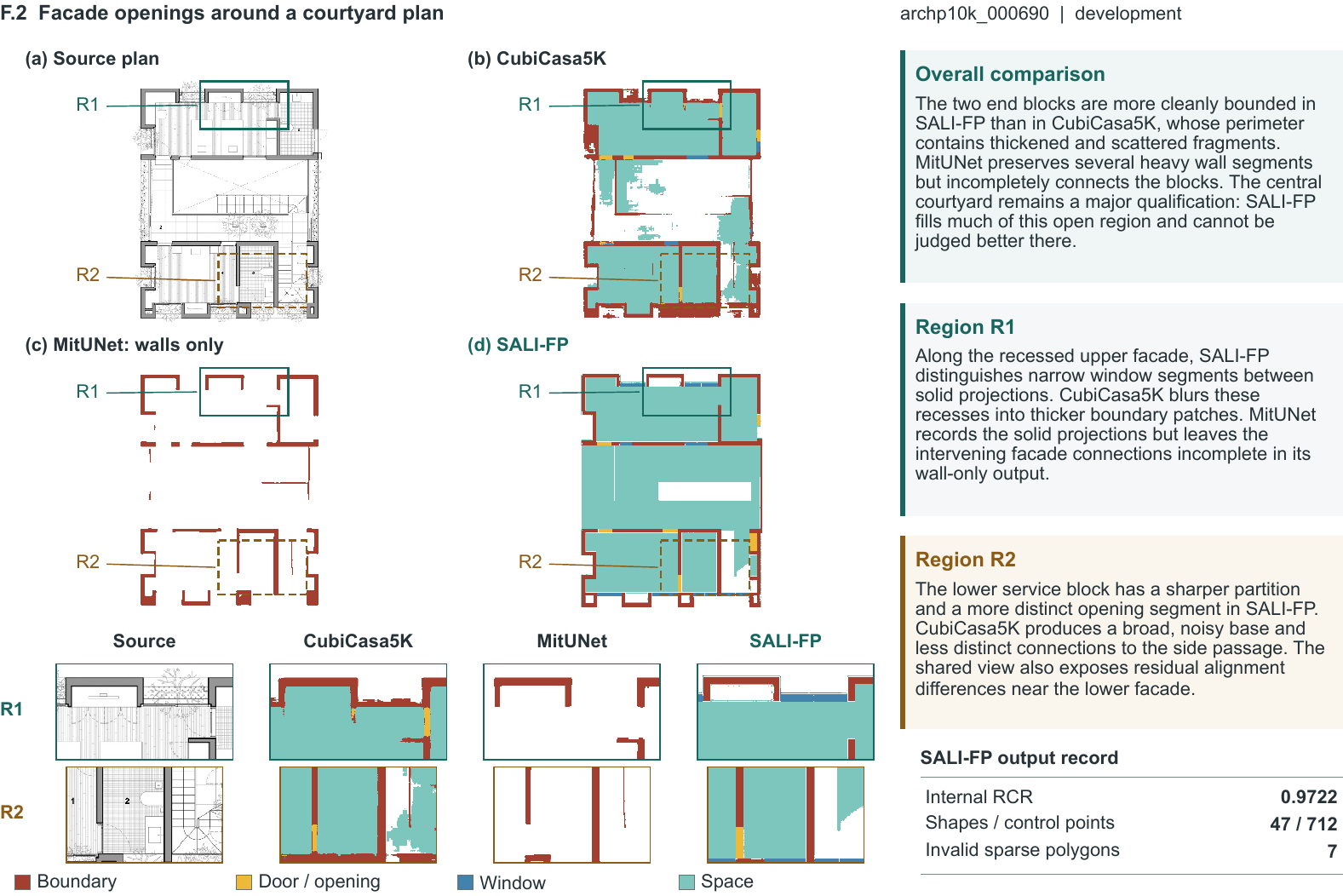}
\captionof{figure}{archp10k\_000690: facade openings around a courtyard plan.}\label{fig:F.2}
\end{landscape}
\restoregeometry

\clearpage
\newgeometry{margin=10mm}
\begin{landscape}
\thispagestyle{empty}
\centering
\includegraphics[width=267mm,height=178mm,keepaspectratio]{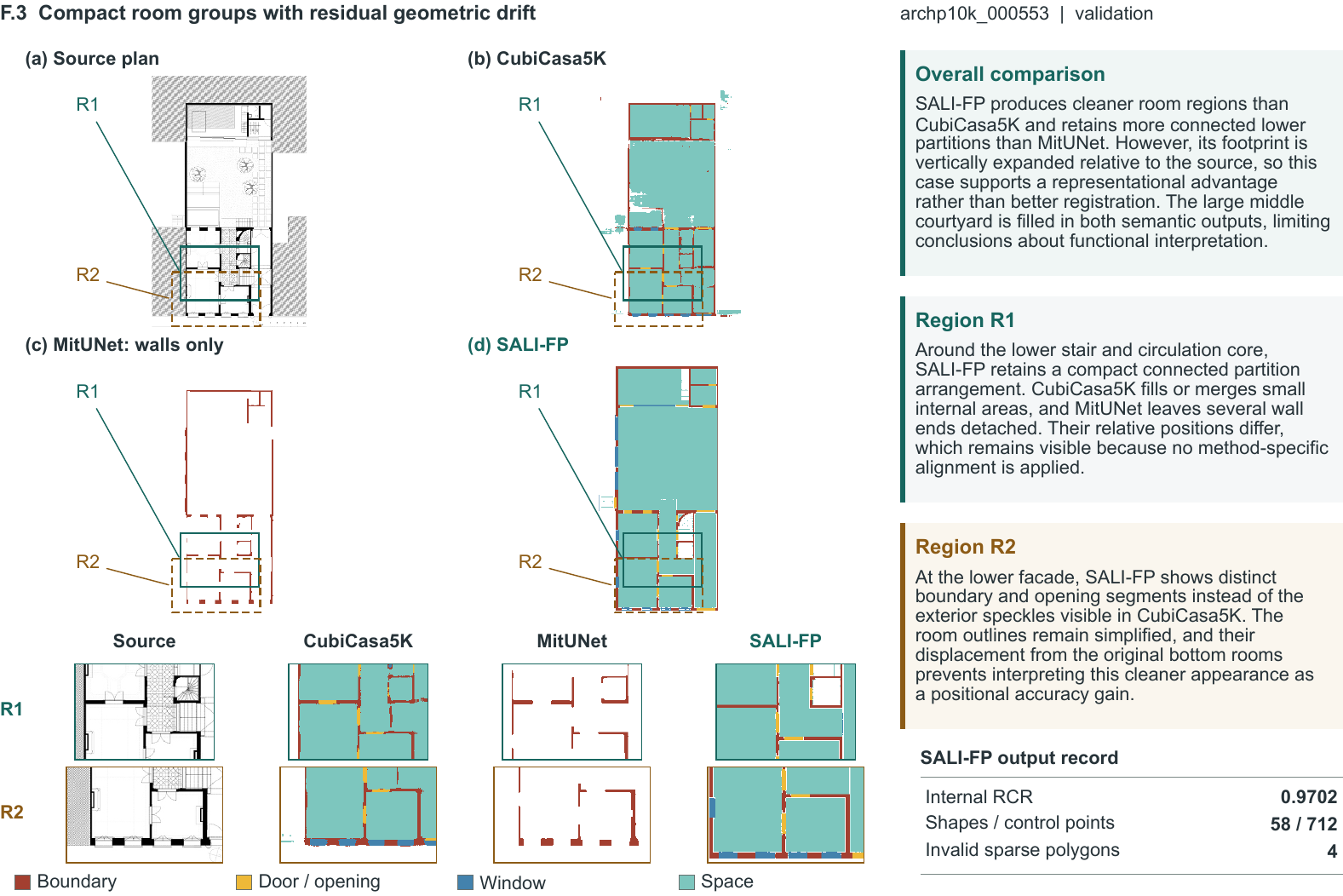}
\captionof{figure}{archp10k\_000553: compact room groups with residual geometric drift.}\label{fig:F.3}
\end{landscape}
\restoregeometry

\clearpage
\newgeometry{margin=10mm}
\begin{landscape}
\thispagestyle{empty}
\centering
\includegraphics[width=267mm,height=178mm,keepaspectratio]{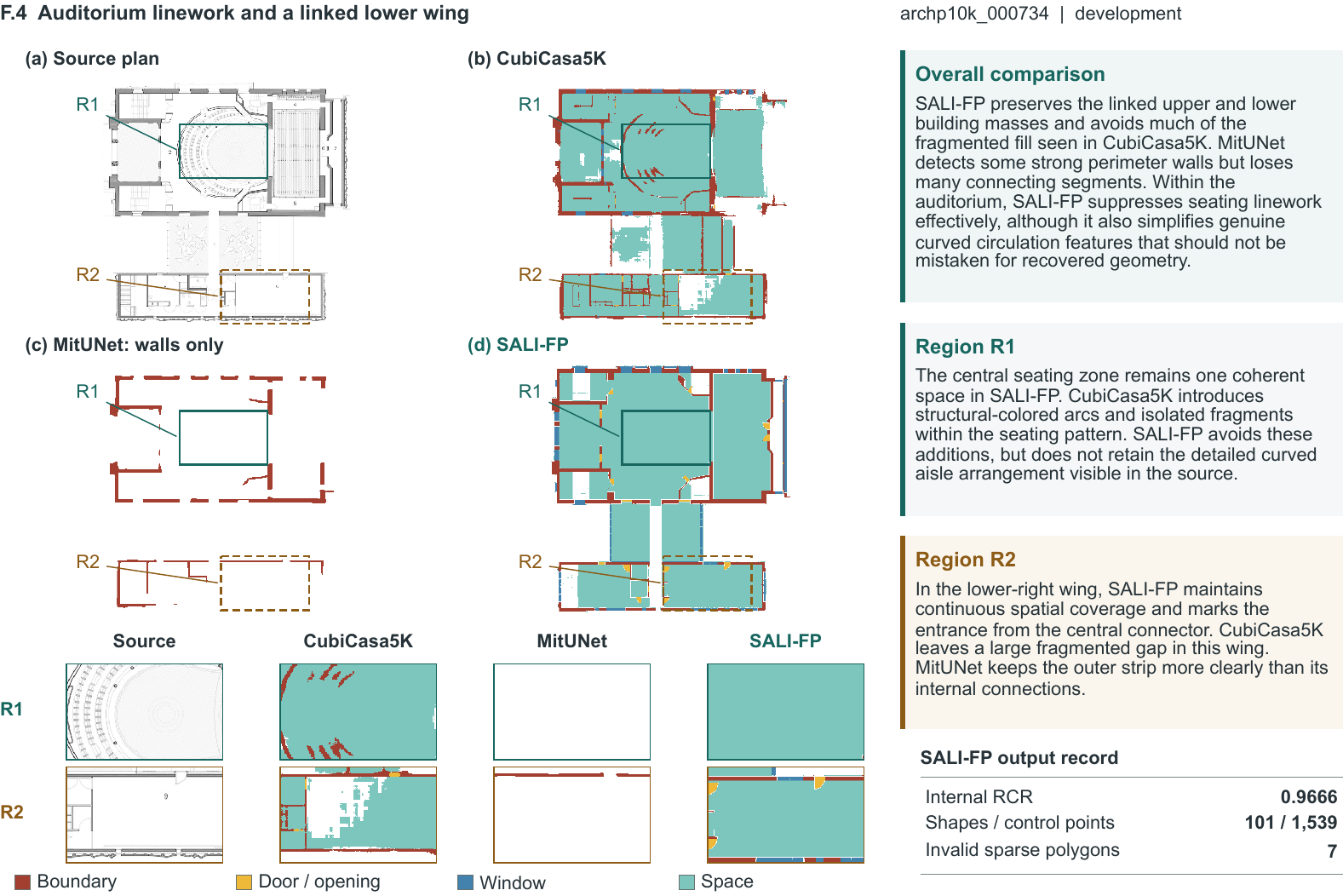}
\captionof{figure}{archp10k\_000734: auditorium linework and a linked lower wing.}\label{fig:F.4}
\end{landscape}
\restoregeometry

\clearpage
\newgeometry{margin=10mm}
\begin{landscape}
\thispagestyle{empty}
\centering
\includegraphics[width=267mm,height=178mm,keepaspectratio]{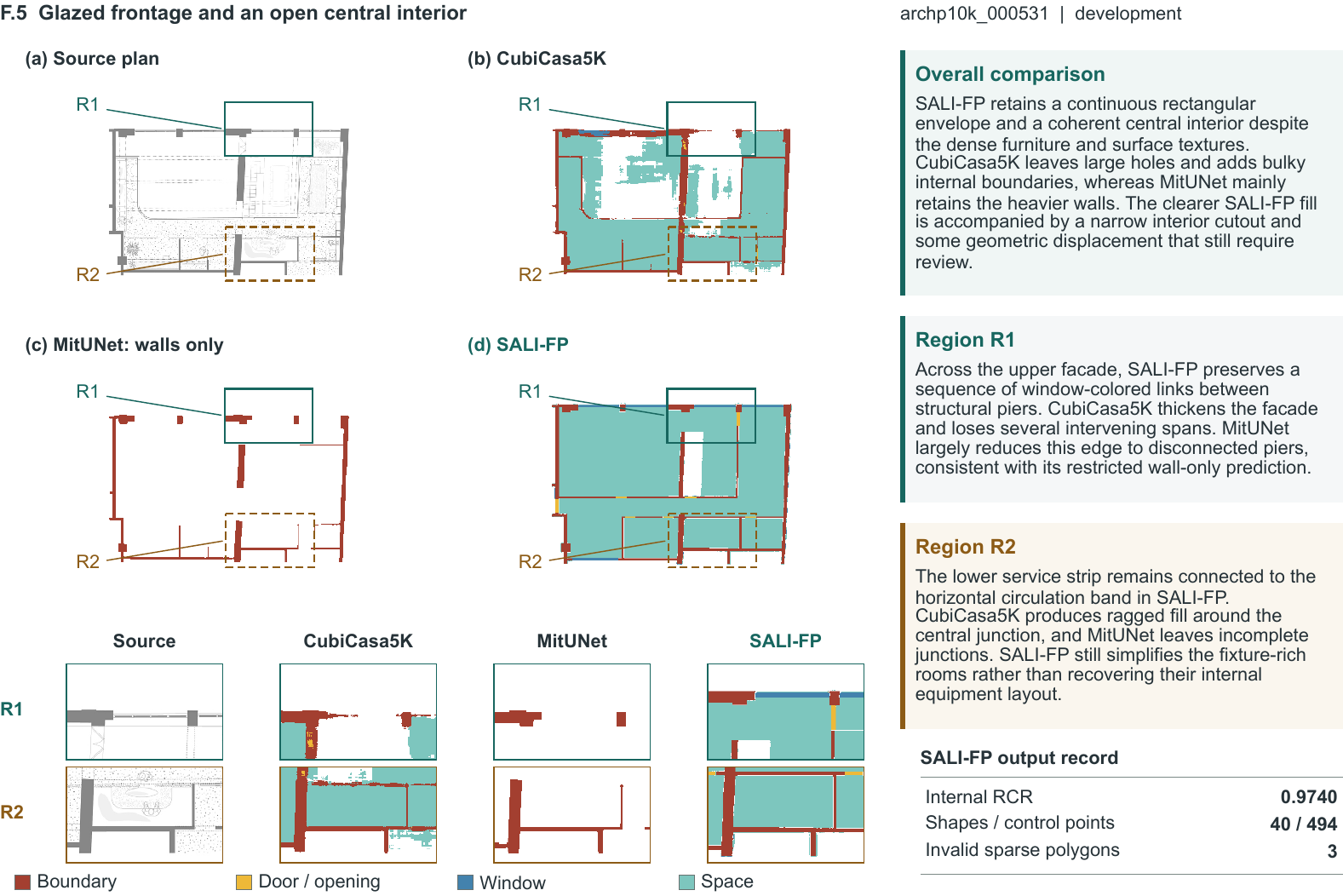}
\captionof{figure}{archp10k\_000531: glazed frontage and an open central interior.}\label{fig:F.5}
\end{landscape}
\restoregeometry

\clearpage
\newgeometry{margin=10mm}
\begin{landscape}
\thispagestyle{empty}
\centering
\includegraphics[width=267mm,height=178mm,keepaspectratio]{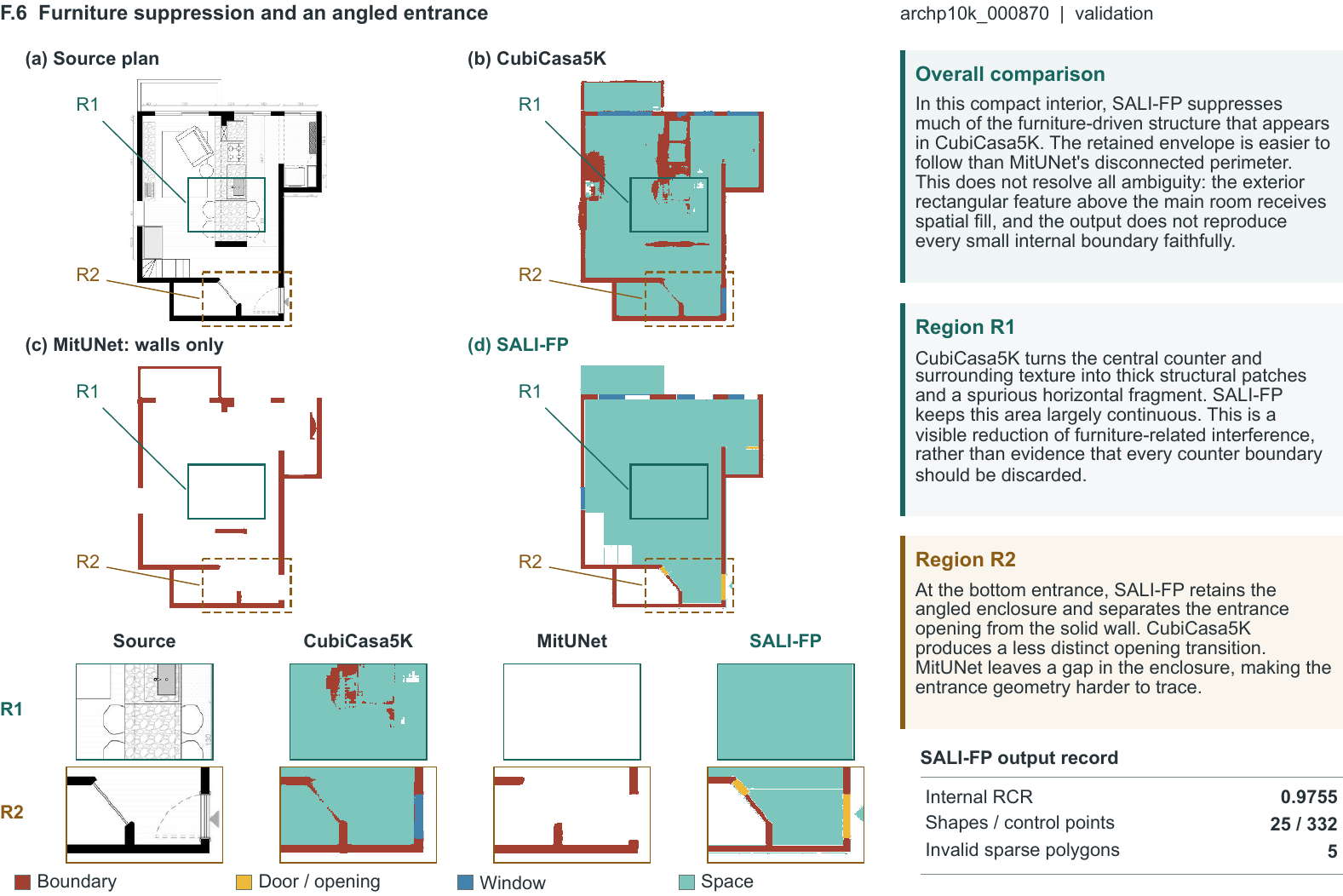}
\captionof{figure}{archp10k\_000870: furniture suppression and an angled entrance.}\label{fig:F.6}
\end{landscape}
\restoregeometry

\clearpage
\newgeometry{margin=10mm}
\begin{landscape}
\thispagestyle{empty}
\centering
\includegraphics[width=267mm,height=178mm,keepaspectratio]{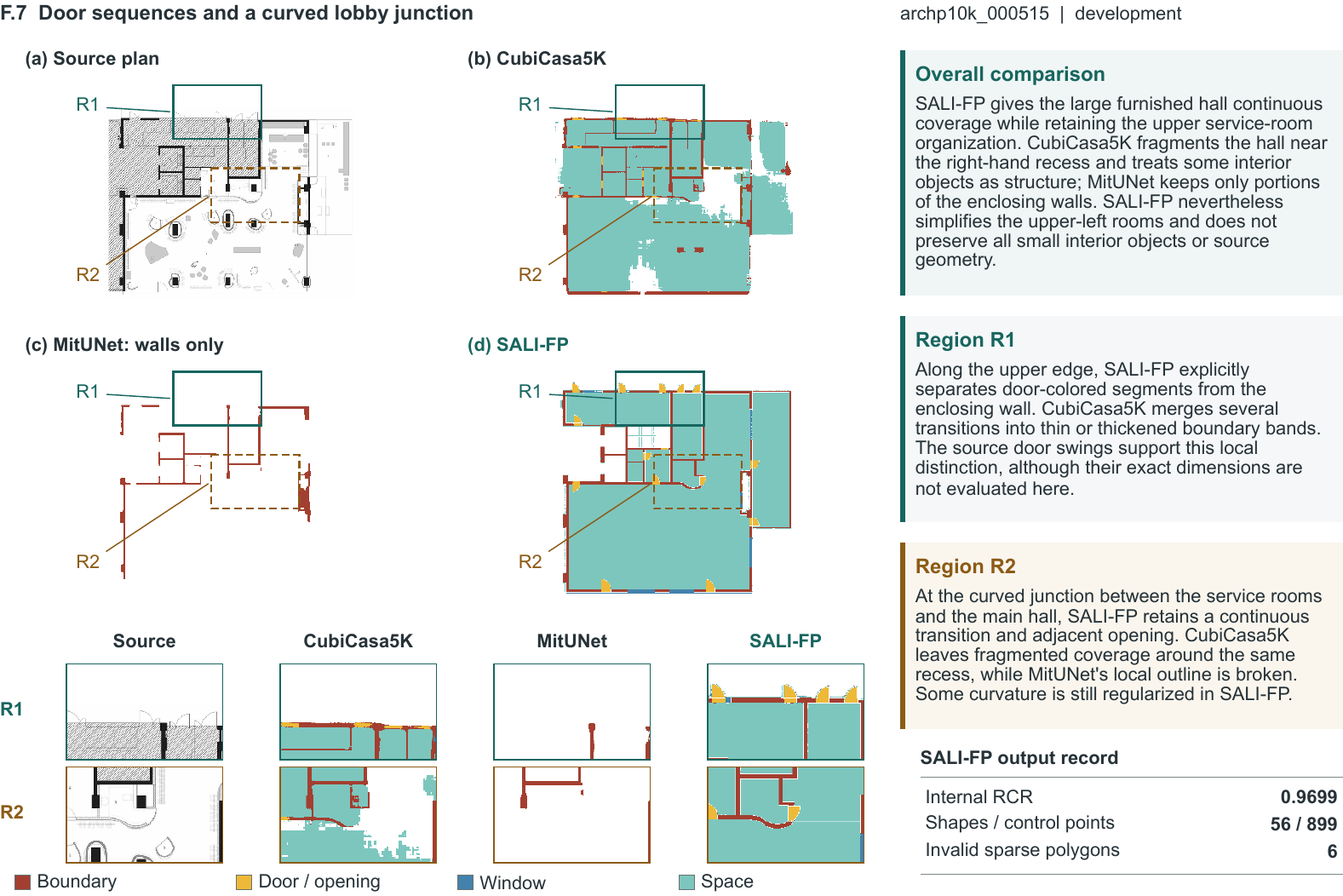}
\captionof{figure}{archp10k\_000515: door sequences and a curved lobby junction.}\label{fig:F.7}
\end{landscape}
\restoregeometry

\clearpage
\newgeometry{margin=10mm}
\begin{landscape}
\thispagestyle{empty}
\centering
\includegraphics[width=267mm,height=178mm,keepaspectratio]{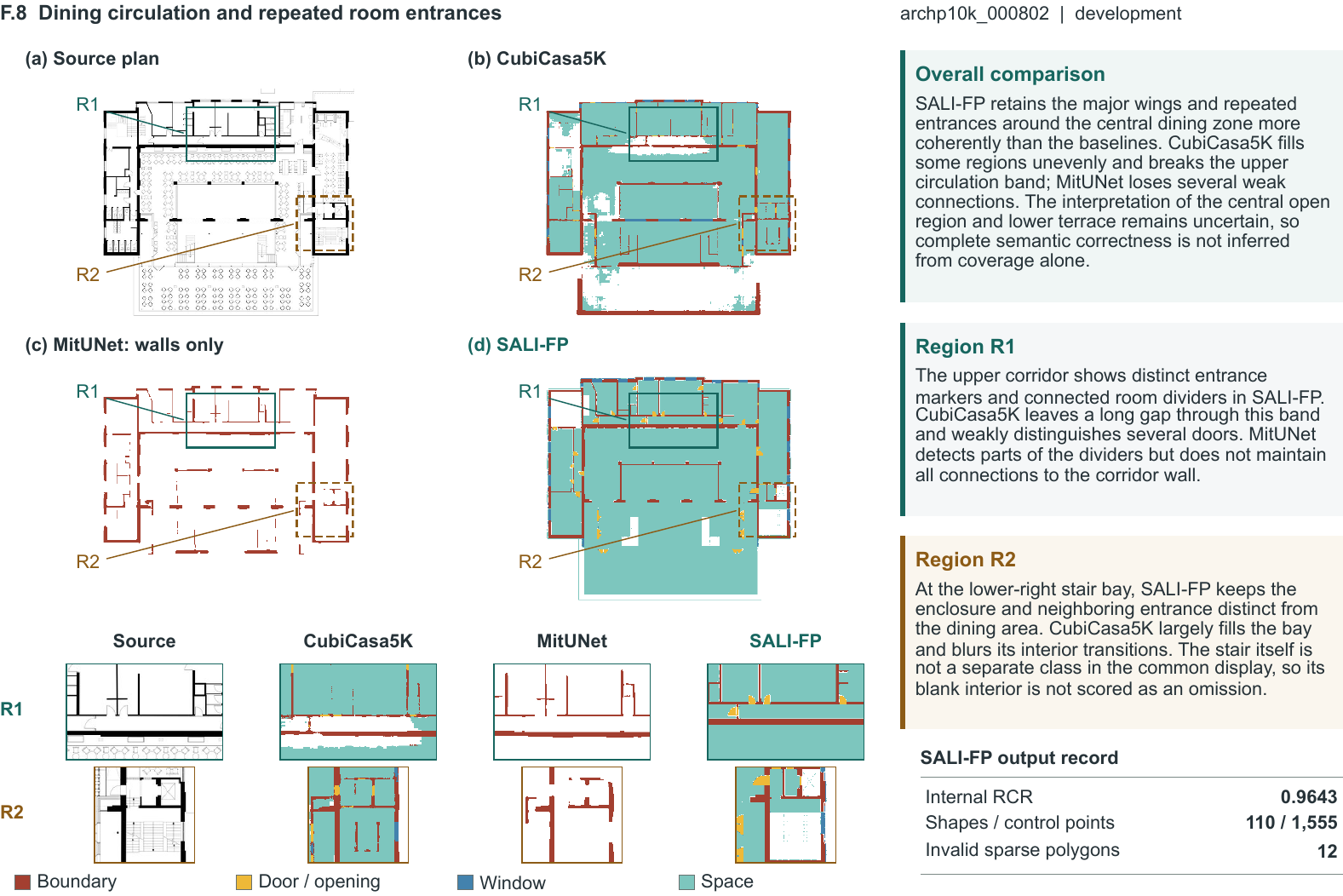}
\captionof{figure}{archp10k\_000802: dining circulation and repeated room entrances.}\label{fig:F.8}
\end{landscape}
\restoregeometry

\clearpage
\newgeometry{margin=10mm}
\begin{landscape}
\thispagestyle{empty}
\centering
\includegraphics[width=267mm,height=178mm,keepaspectratio]{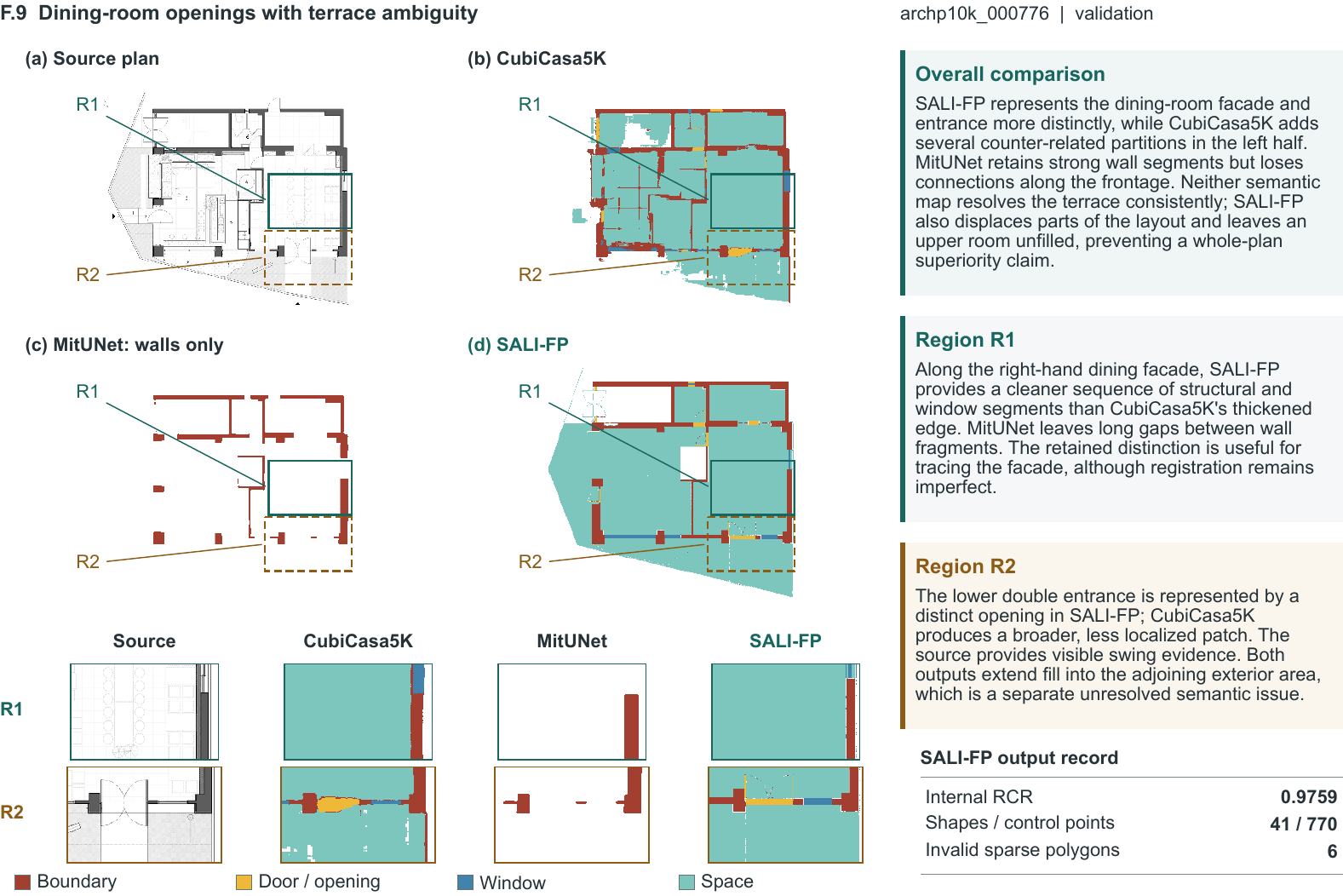}
\captionof{figure}{archp10k\_000776: dining-room openings with terrace ambiguity.}\label{fig:F.9}
\end{landscape}
\restoregeometry

\clearpage
\newgeometry{margin=10mm}
\begin{landscape}
\thispagestyle{empty}
\centering
\includegraphics[width=267mm,height=178mm,keepaspectratio]{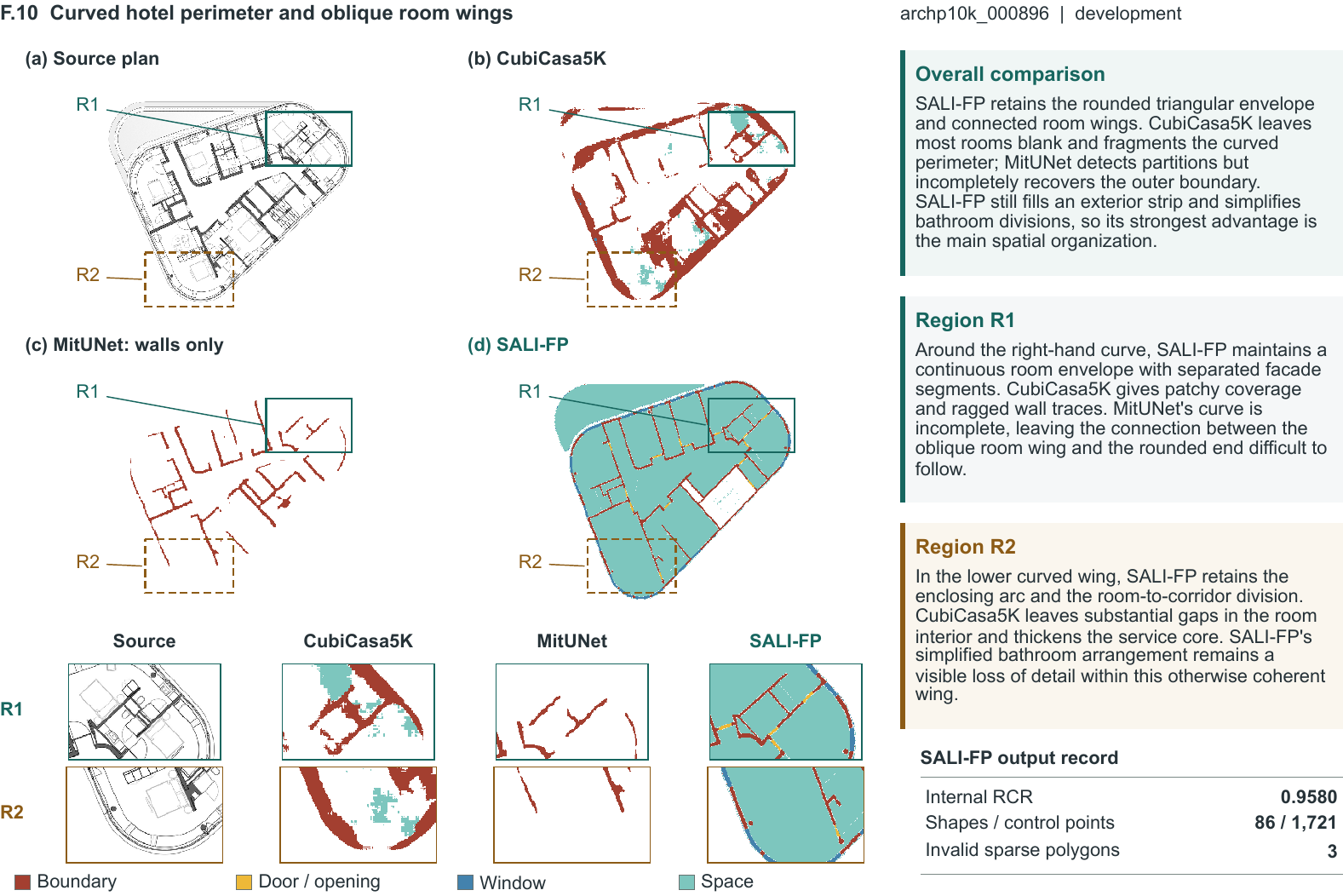}
\captionof{figure}{archp10k\_000896: curved hotel perimeter and oblique room wings.}\label{fig:F.10}
\end{landscape}
\restoregeometry

\clearpage
\newgeometry{margin=10mm}
\begin{landscape}
\thispagestyle{empty}
\centering
\includegraphics[width=267mm,height=178mm,keepaspectratio]{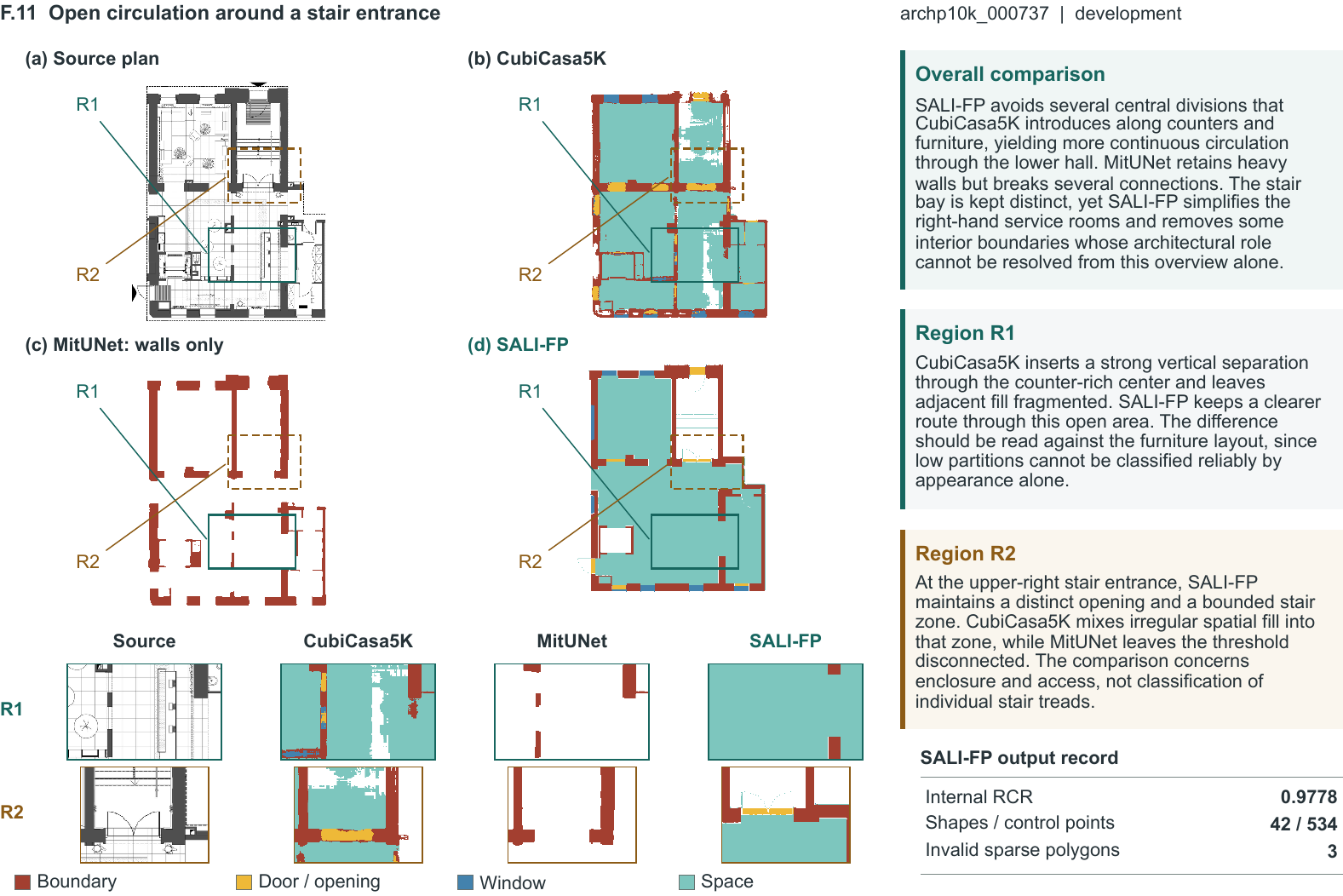}
\captionof{figure}{archp10k\_000737: open circulation around a stair entrance.}\label{fig:F.11}
\end{landscape}
\restoregeometry

\clearpage
\newgeometry{margin=10mm}
\begin{landscape}
\thispagestyle{empty}
\centering
\includegraphics[width=267mm,height=178mm,keepaspectratio]{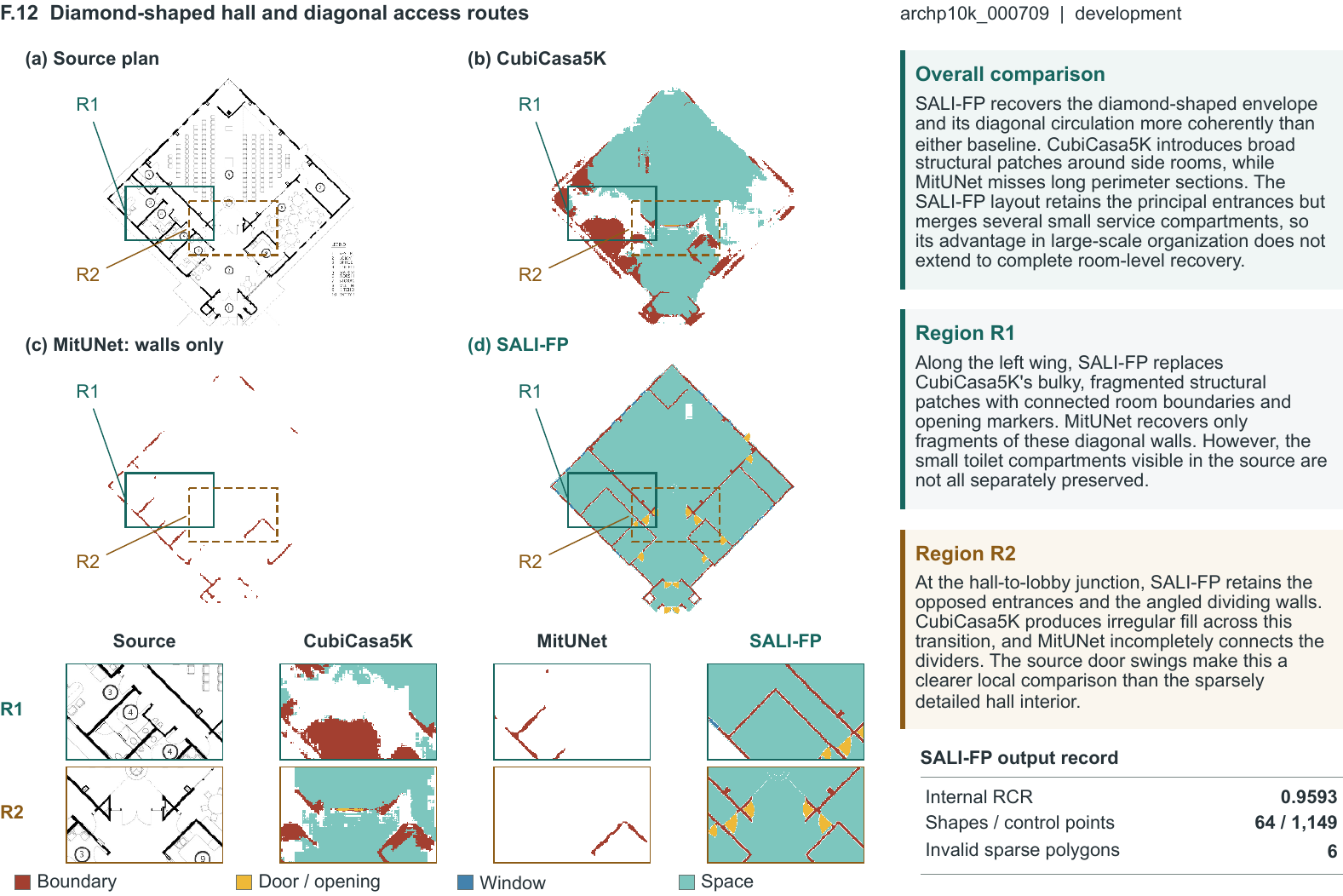}
\captionof{figure}{archp10k\_000709: diamond-shaped hall and diagonal access routes.}\label{fig:F.12}
\end{landscape}
\restoregeometry

\clearpage
\newgeometry{margin=10mm}
\begin{landscape}
\thispagestyle{empty}
\centering
\includegraphics[width=267mm,height=178mm,keepaspectratio]{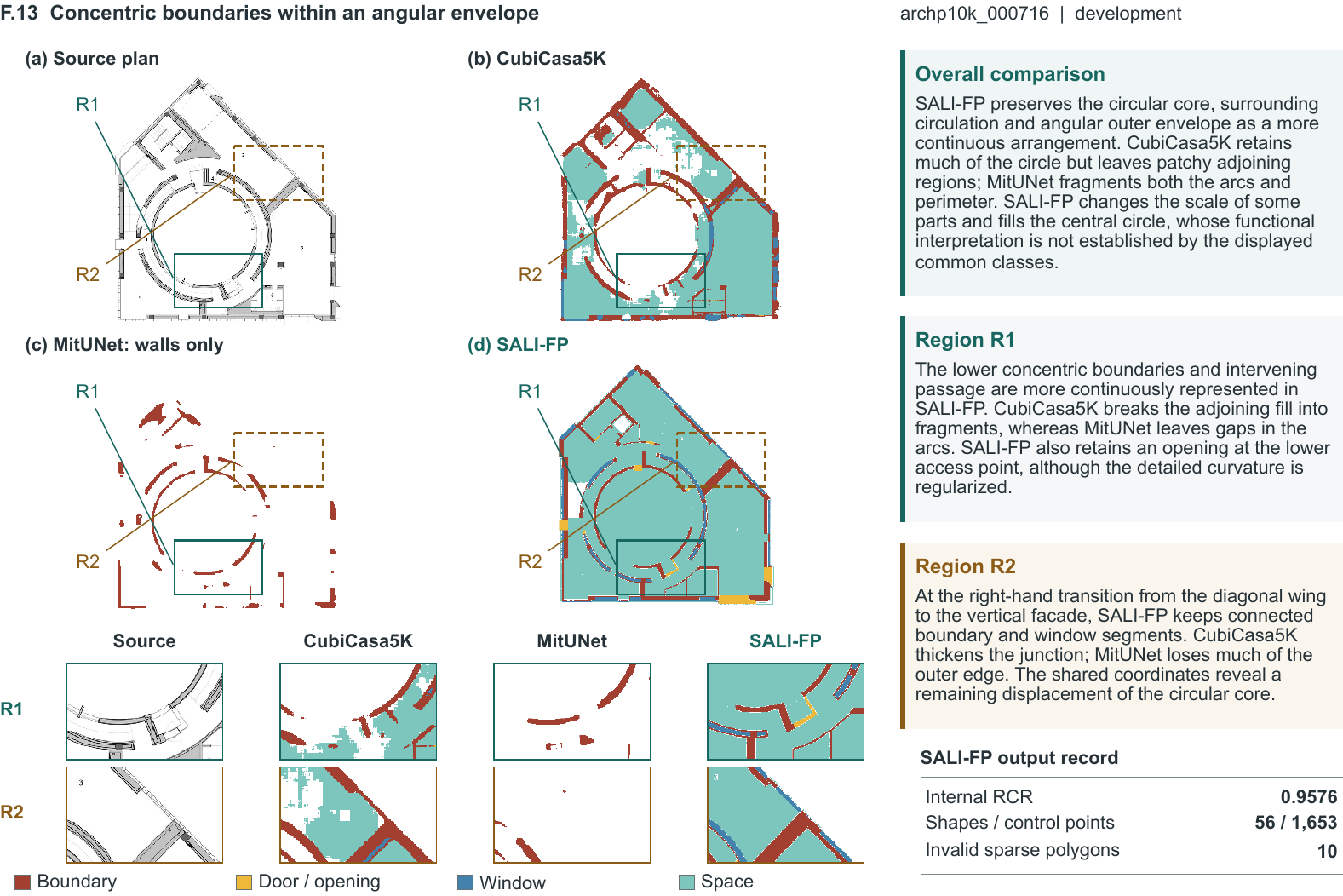}
\captionof{figure}{archp10k\_000716: concentric boundaries within an angular envelope.}\label{fig:F.13}
\end{landscape}
\restoregeometry

\clearpage
\newgeometry{margin=10mm}
\begin{landscape}
\thispagestyle{empty}
\centering
\includegraphics[width=267mm,height=178mm,keepaspectratio]{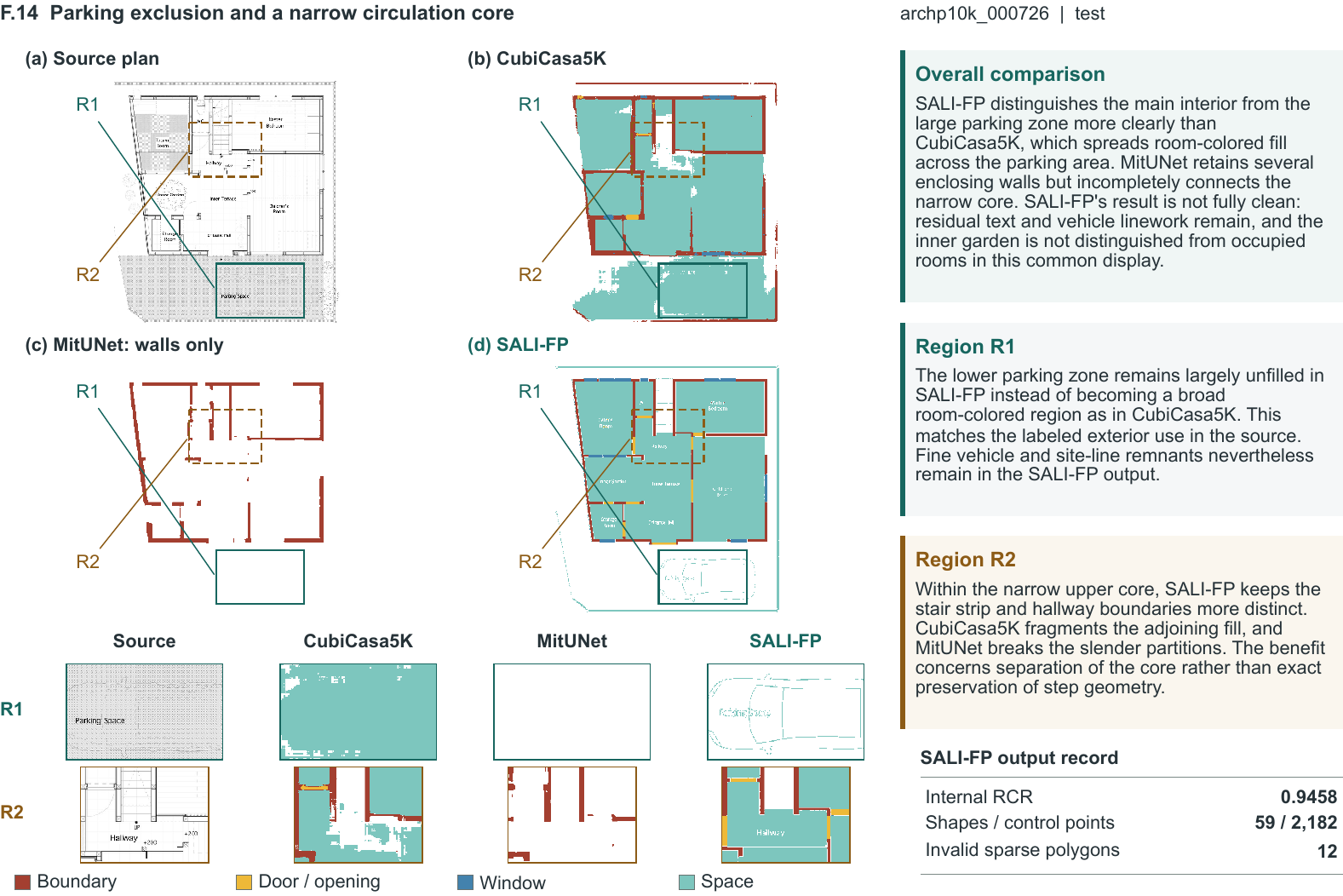}
\captionof{figure}{archp10k\_000726: parking exclusion and a narrow circulation core.}\label{fig:F.14}
\end{landscape}
\restoregeometry

\clearpage
\newgeometry{margin=10mm}
\begin{landscape}
\thispagestyle{empty}
\centering
\includegraphics[width=267mm,height=178mm,keepaspectratio]{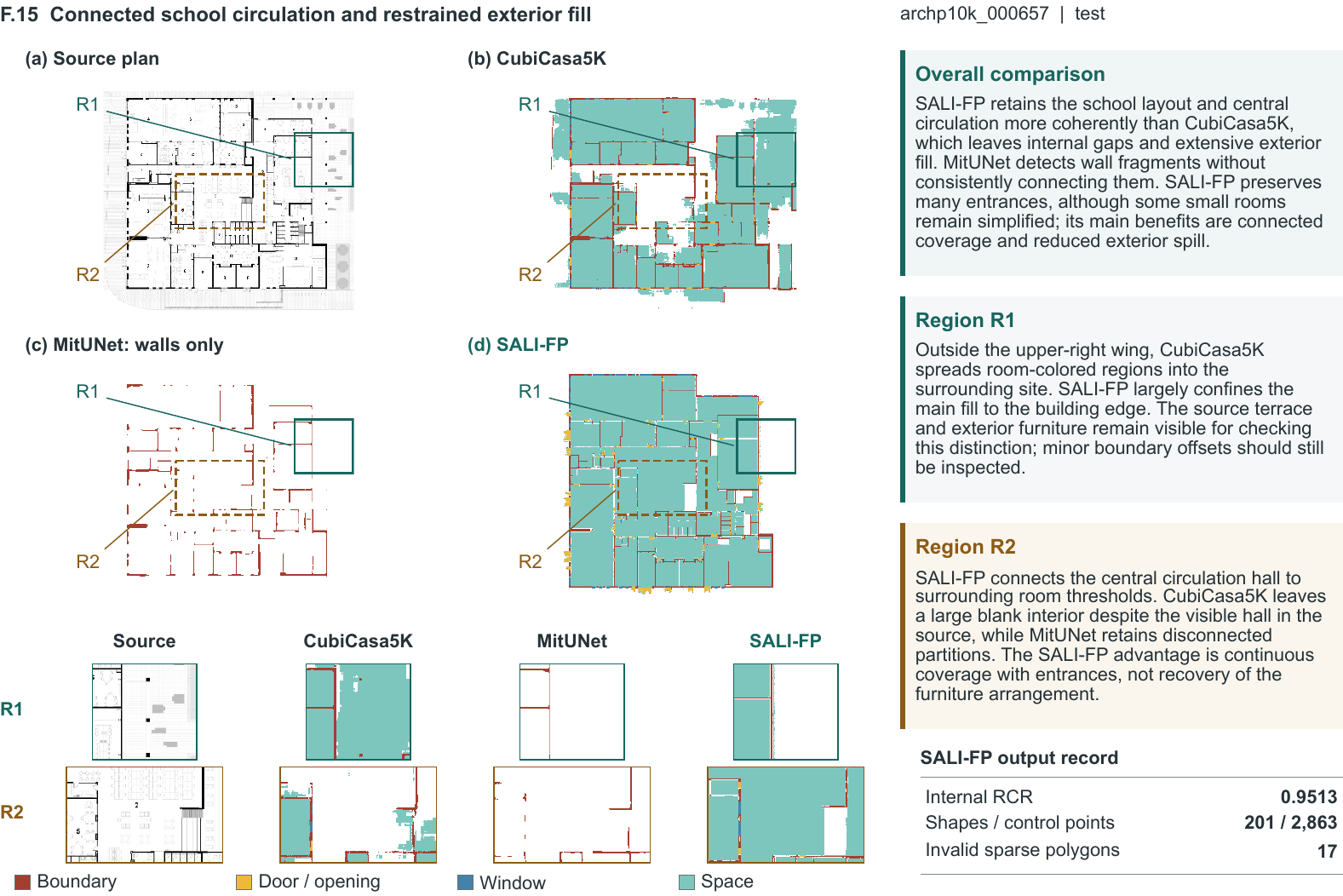}
\captionof{figure}{archp10k\_000657: connected school circulation and restrained exterior fill.}\label{fig:F.15}
\end{landscape}
\restoregeometry

\clearpage
\newgeometry{margin=10mm}
\begin{landscape}
\thispagestyle{empty}
\centering
\includegraphics[width=267mm,height=178mm,keepaspectratio]{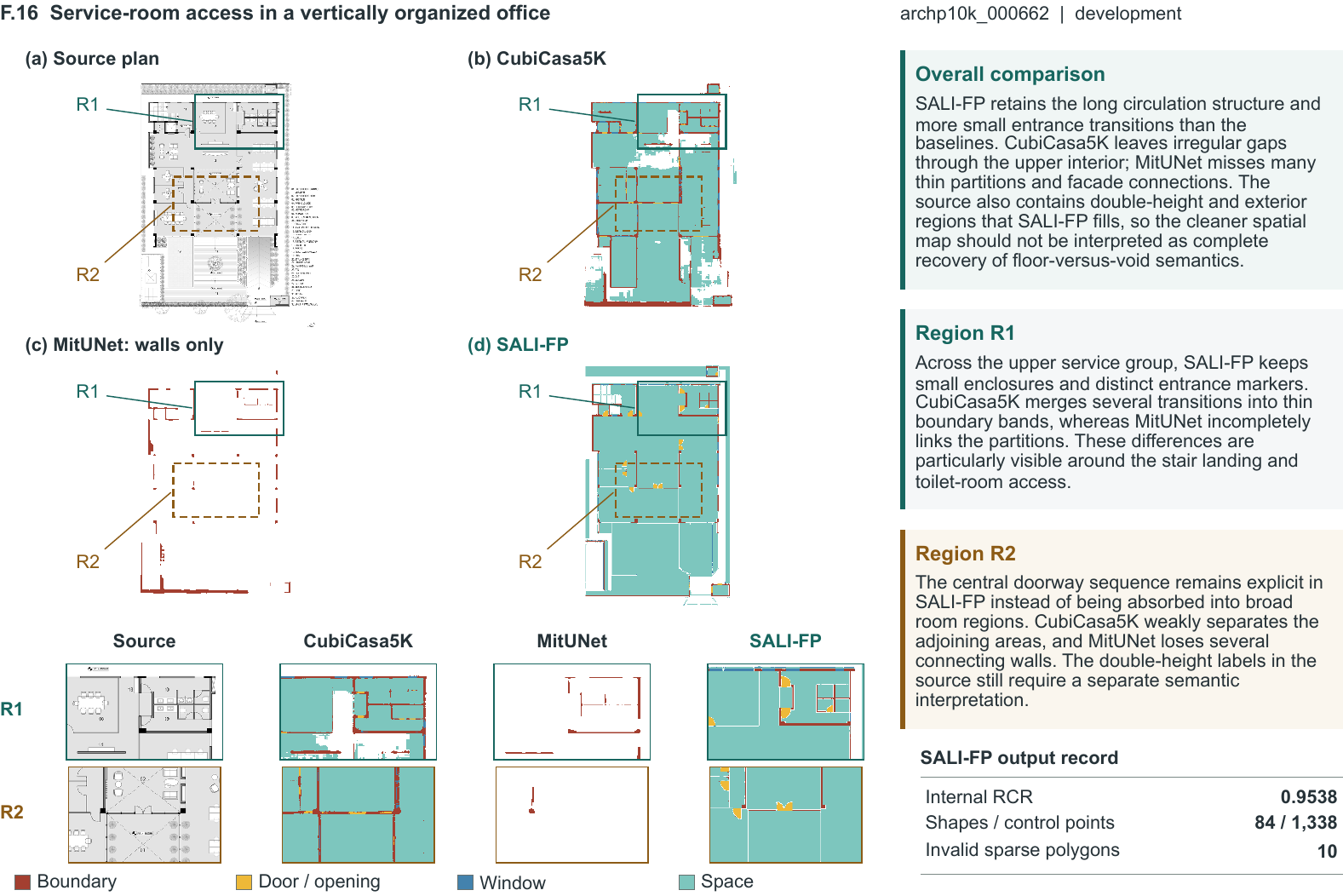}
\captionof{figure}{archp10k\_000662: service-room access in a vertically organized office.}\label{fig:F.16}
\end{landscape}
\restoregeometry

\clearpage
\newgeometry{margin=10mm}
\begin{landscape}
\thispagestyle{empty}
\centering
\includegraphics[width=267mm,height=178mm,keepaspectratio]{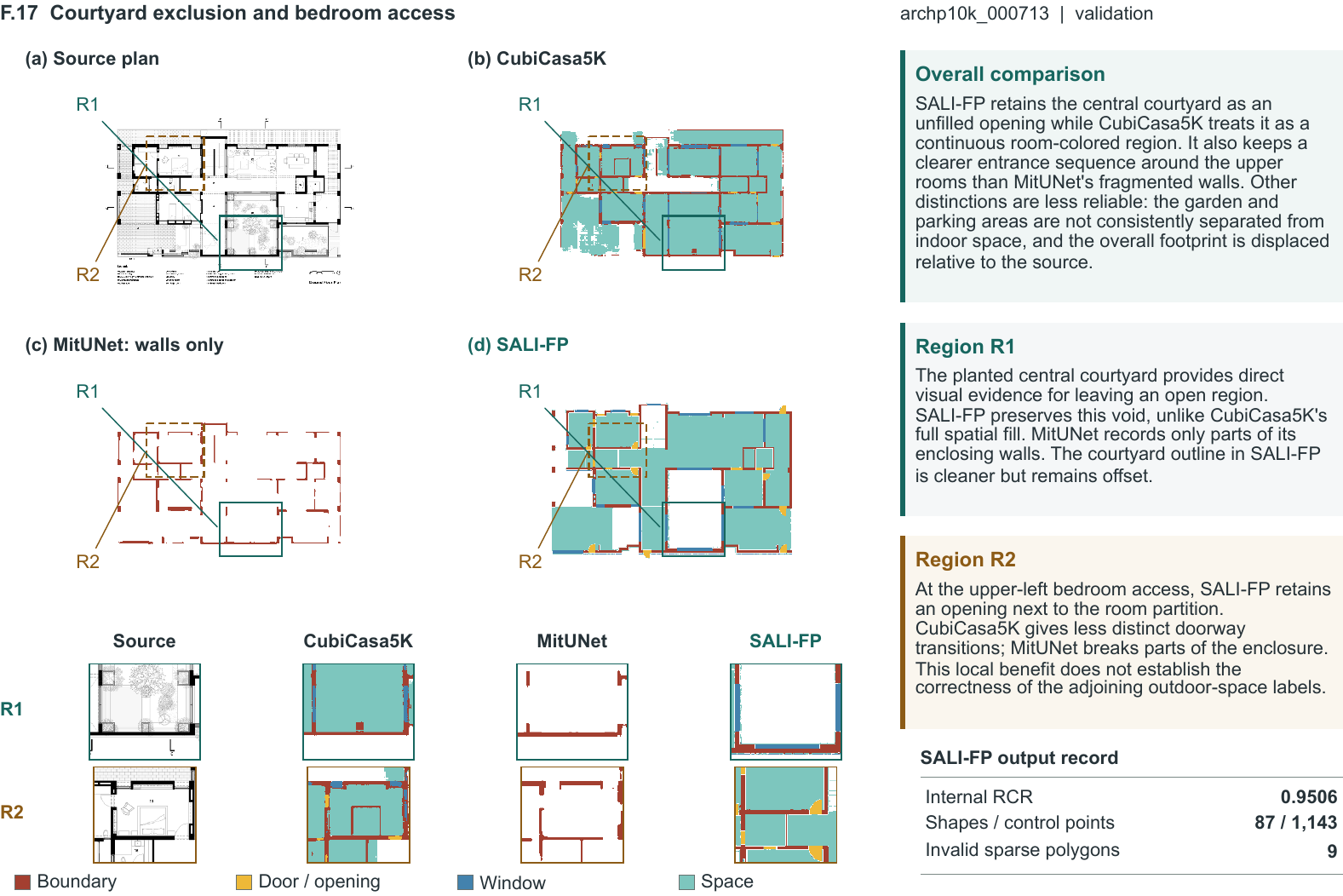}
\captionof{figure}{archp10k\_000713: courtyard exclusion and bedroom access.}\label{fig:F.17}
\end{landscape}
\restoregeometry

\clearpage
\newgeometry{margin=10mm}
\begin{landscape}
\thispagestyle{empty}
\centering
\includegraphics[width=267mm,height=178mm,keepaspectratio]{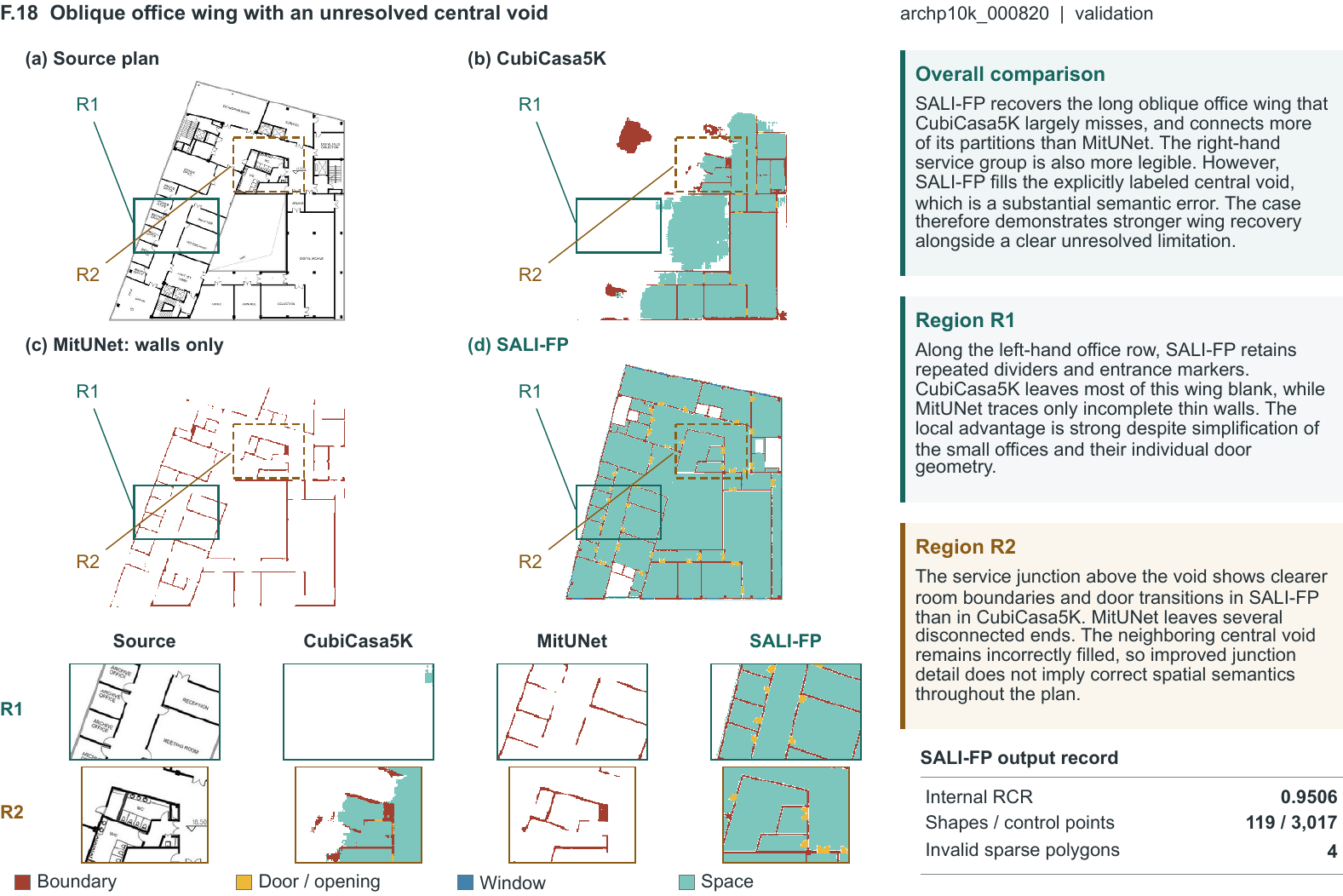}
\captionof{figure}{archp10k\_000820: oblique office wing with an unresolved central void.}\label{fig:F.18}
\end{landscape}
\restoregeometry

\clearpage
\newgeometry{margin=10mm}
\begin{landscape}
\thispagestyle{empty}
\centering
\includegraphics[width=267mm,height=178mm,keepaspectratio]{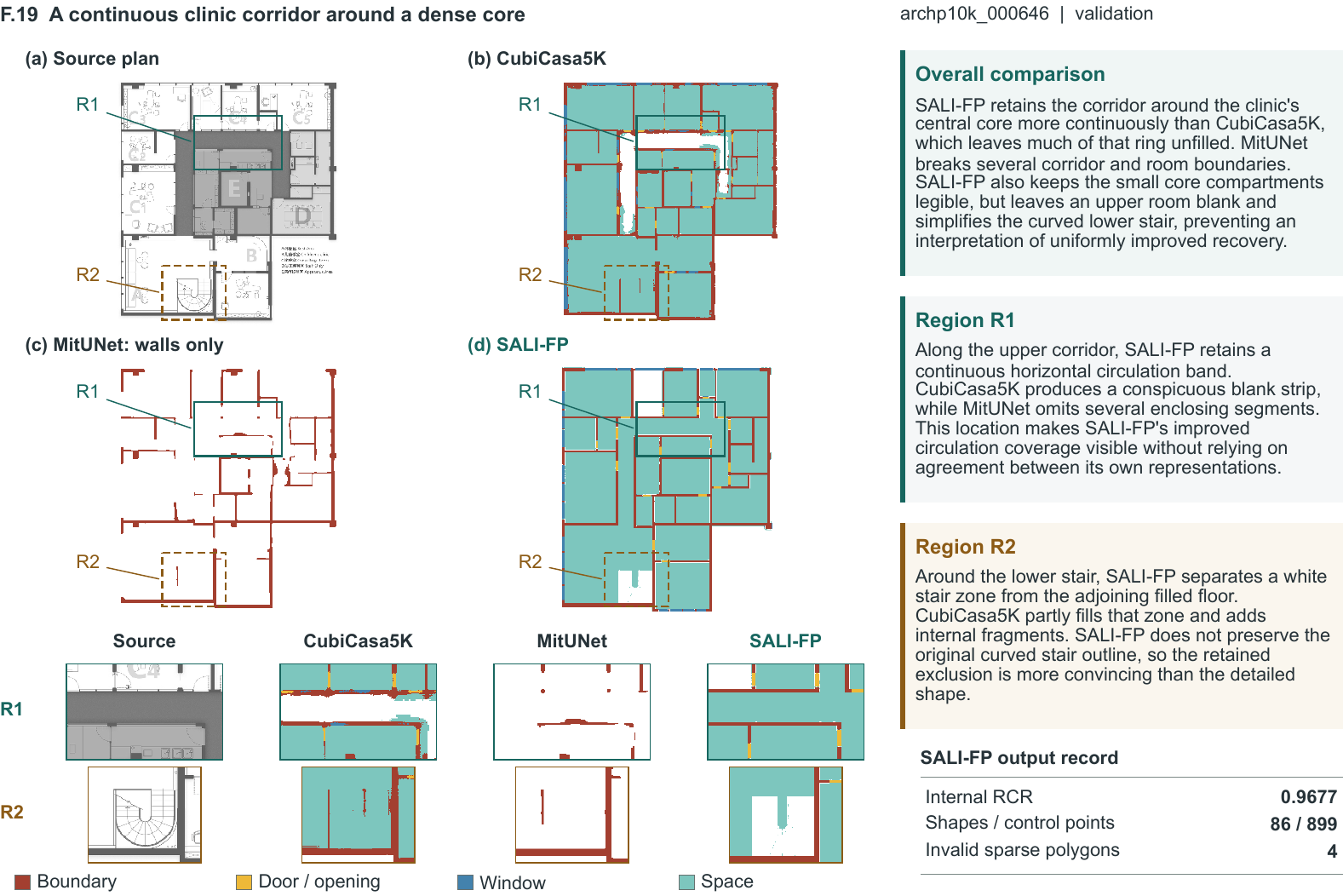}
\captionof{figure}{archp10k\_000646: a continuous clinic corridor around a dense core.}\label{fig:F.19}
\end{landscape}
\restoregeometry

\clearpage
\newgeometry{margin=10mm}
\begin{landscape}
\thispagestyle{empty}
\centering
\includegraphics[width=267mm,height=178mm,keepaspectratio]{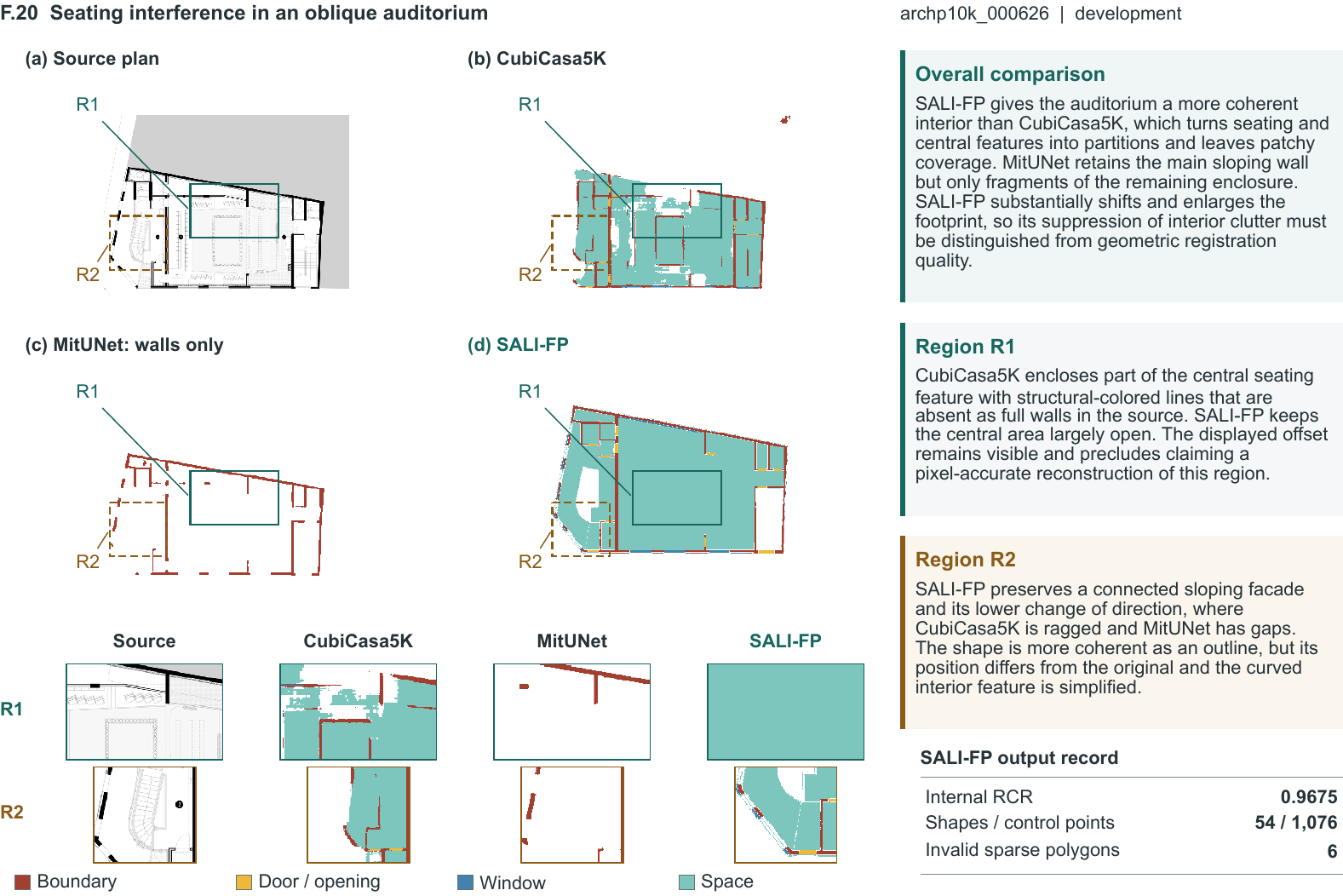}
\captionof{figure}{archp10k\_000626: seating interference in an oblique auditorium.}\label{fig:F.20}
\end{landscape}
\restoregeometry

\clearpage
\newgeometry{margin=10mm}
\begin{landscape}
\thispagestyle{empty}
\centering
\includegraphics[width=267mm,height=178mm,keepaspectratio]{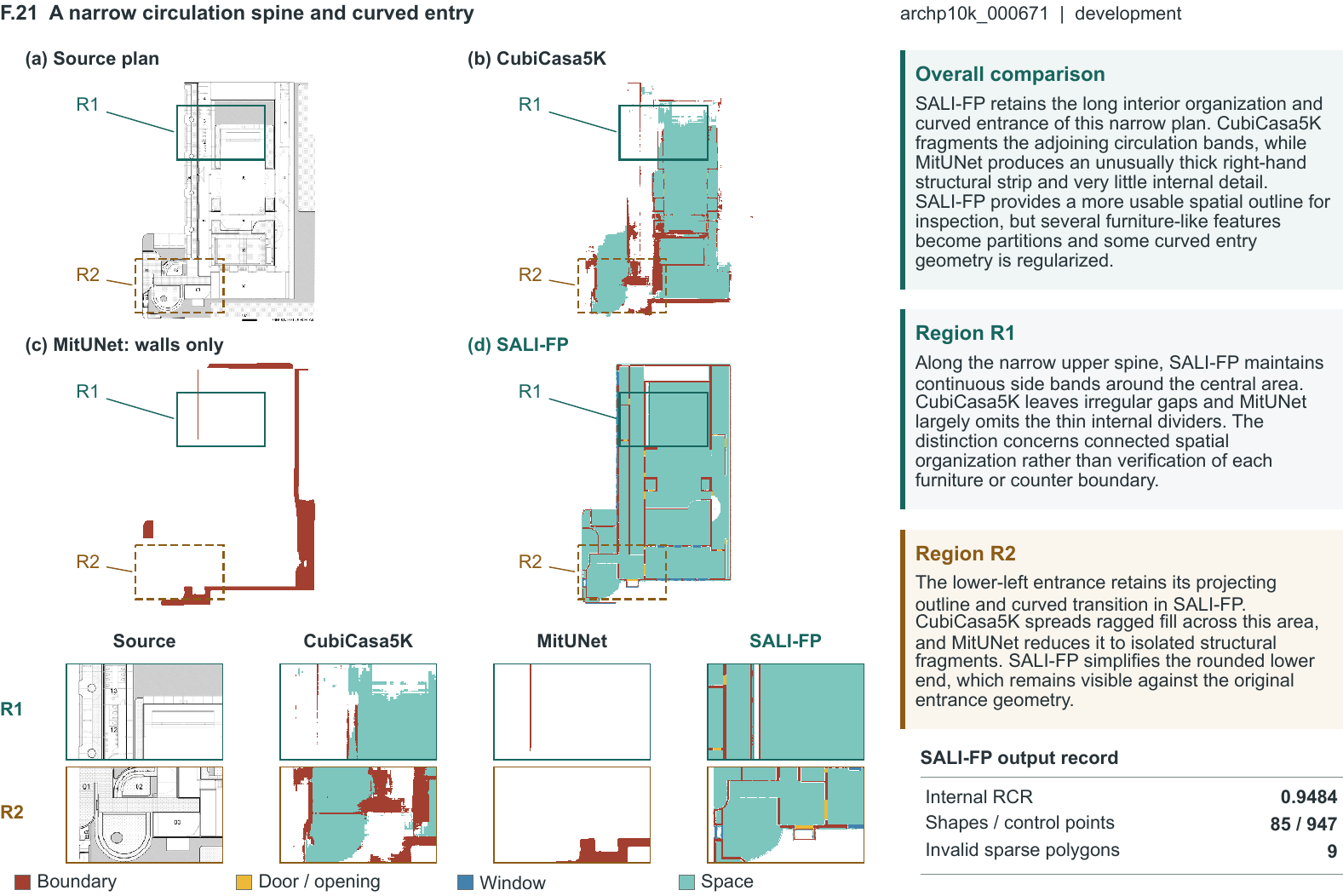}
\captionof{figure}{archp10k\_000671: a narrow circulation spine and curved entry.}\label{fig:F.21}
\end{landscape}
\restoregeometry

\clearpage
\newgeometry{margin=10mm}
\begin{landscape}
\thispagestyle{empty}
\centering
\includegraphics[width=267mm,height=178mm,keepaspectratio]{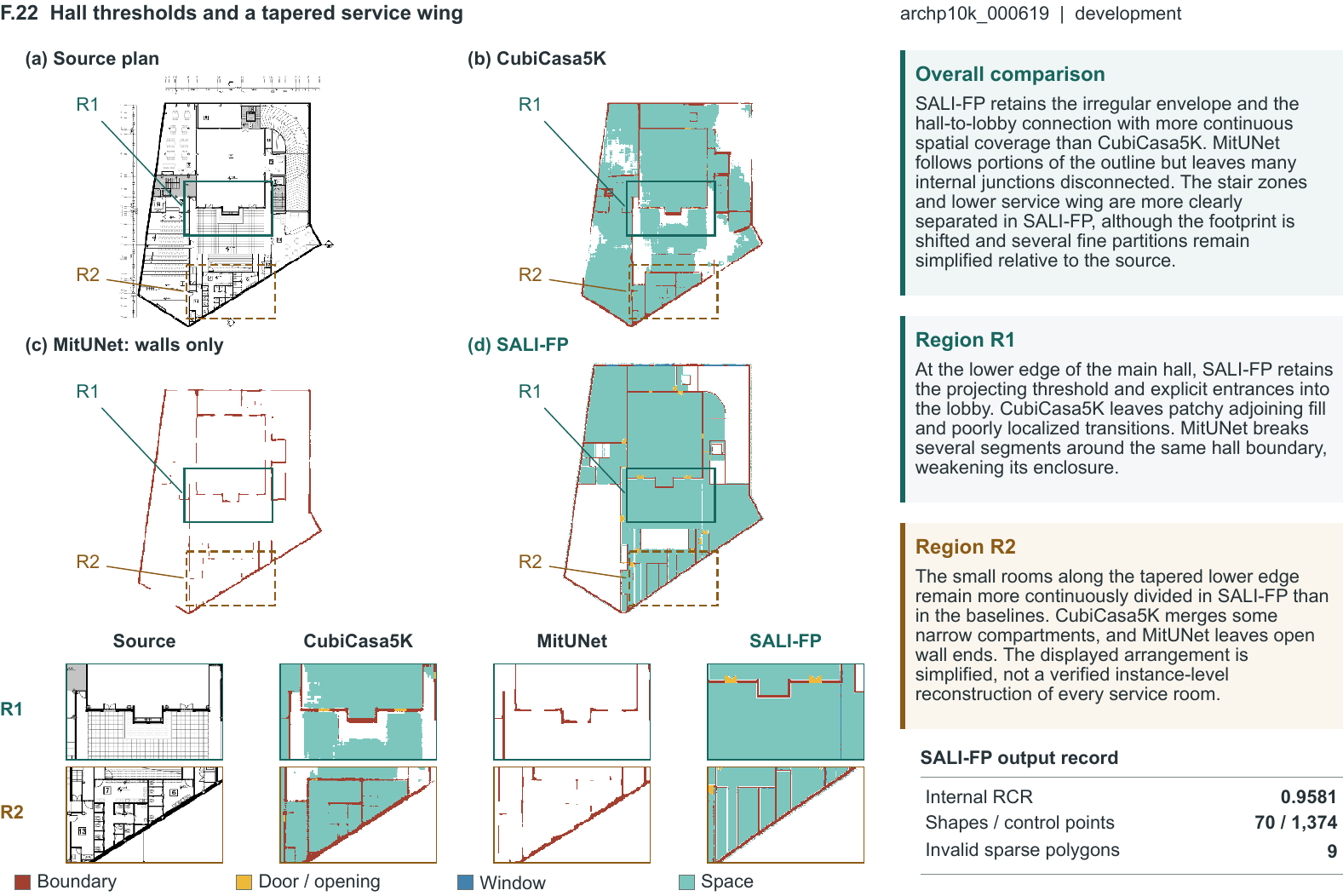}
\captionof{figure}{archp10k\_000619: hall thresholds and a tapered service wing.}\label{fig:F.22}
\end{landscape}
\restoregeometry

\clearpage
\newgeometry{margin=10mm}
\begin{landscape}
\thispagestyle{empty}
\centering
\includegraphics[width=267mm,height=178mm,keepaspectratio]{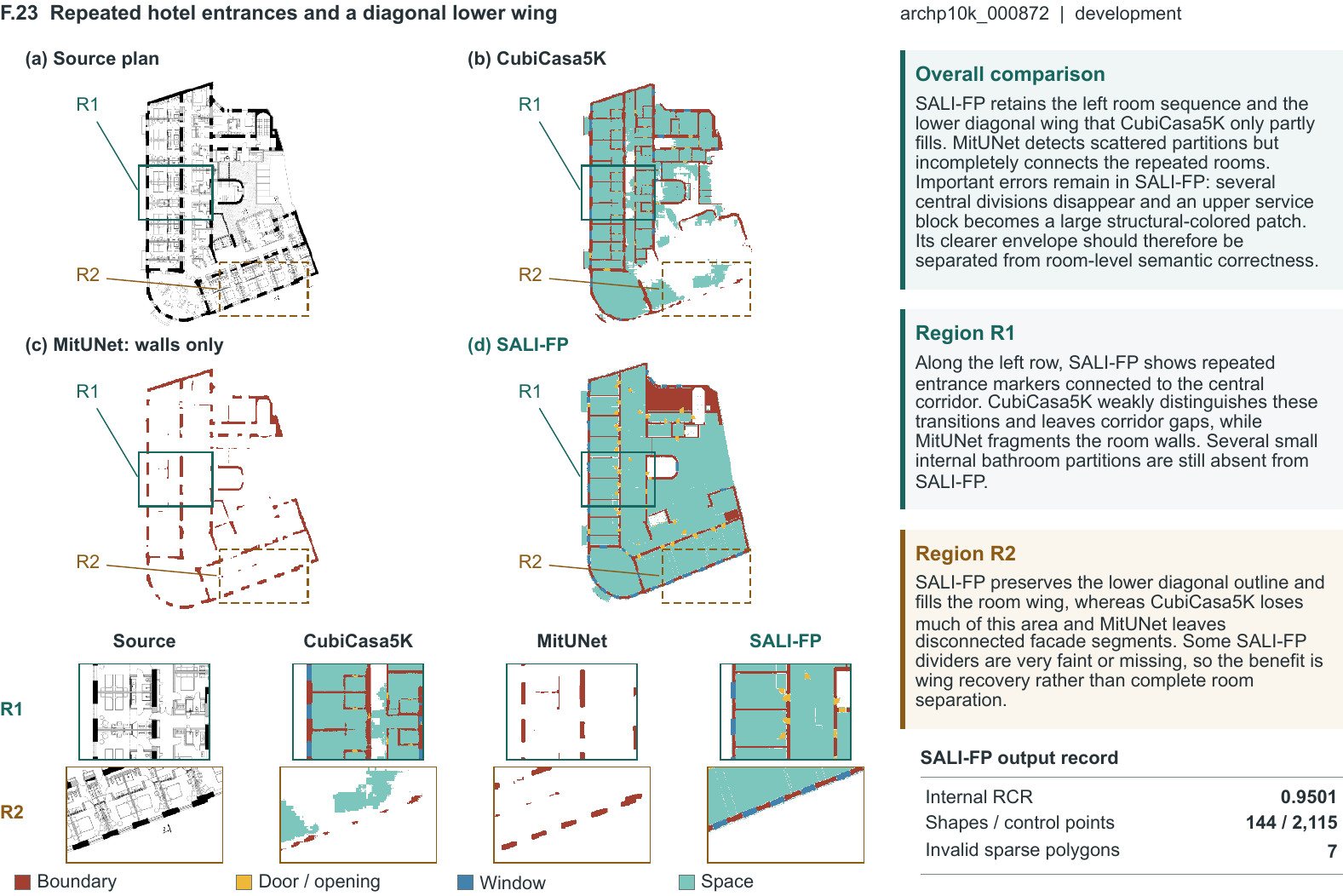}
\captionof{figure}{archp10k\_000872: repeated hotel entrances and a diagonal lower wing.}\label{fig:F.23}
\end{landscape}
\restoregeometry

\clearpage
\newgeometry{margin=10mm}
\begin{landscape}
\thispagestyle{empty}
\centering
\includegraphics[width=267mm,height=178mm,keepaspectratio]{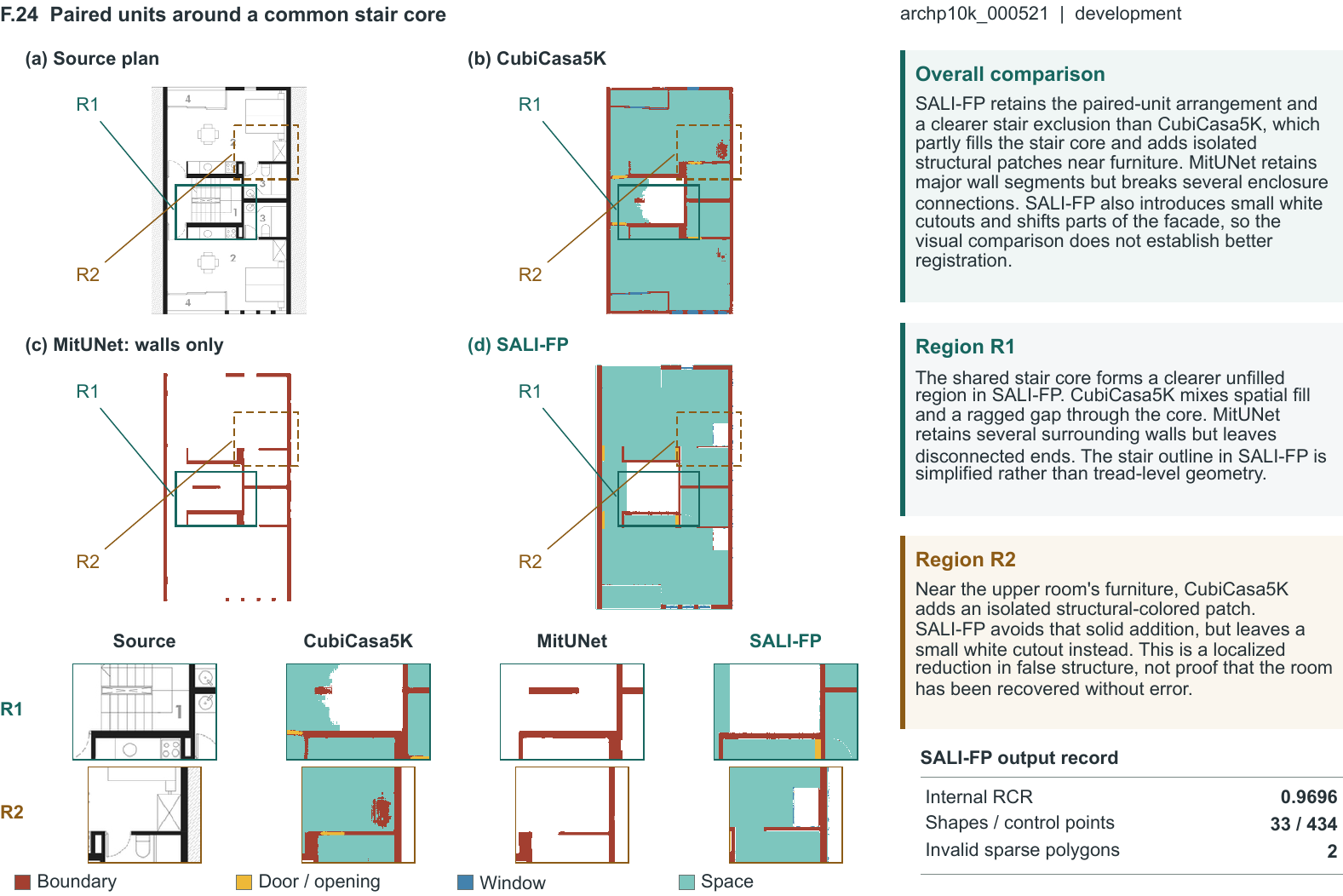}
\captionof{figure}{archp10k\_000521: paired units around a common stair core.}\label{fig:F.24}
\end{landscape}
\restoregeometry

\clearpage
\newgeometry{margin=10mm}
\begin{landscape}
\thispagestyle{empty}
\centering
\includegraphics[width=267mm,height=178mm,keepaspectratio]{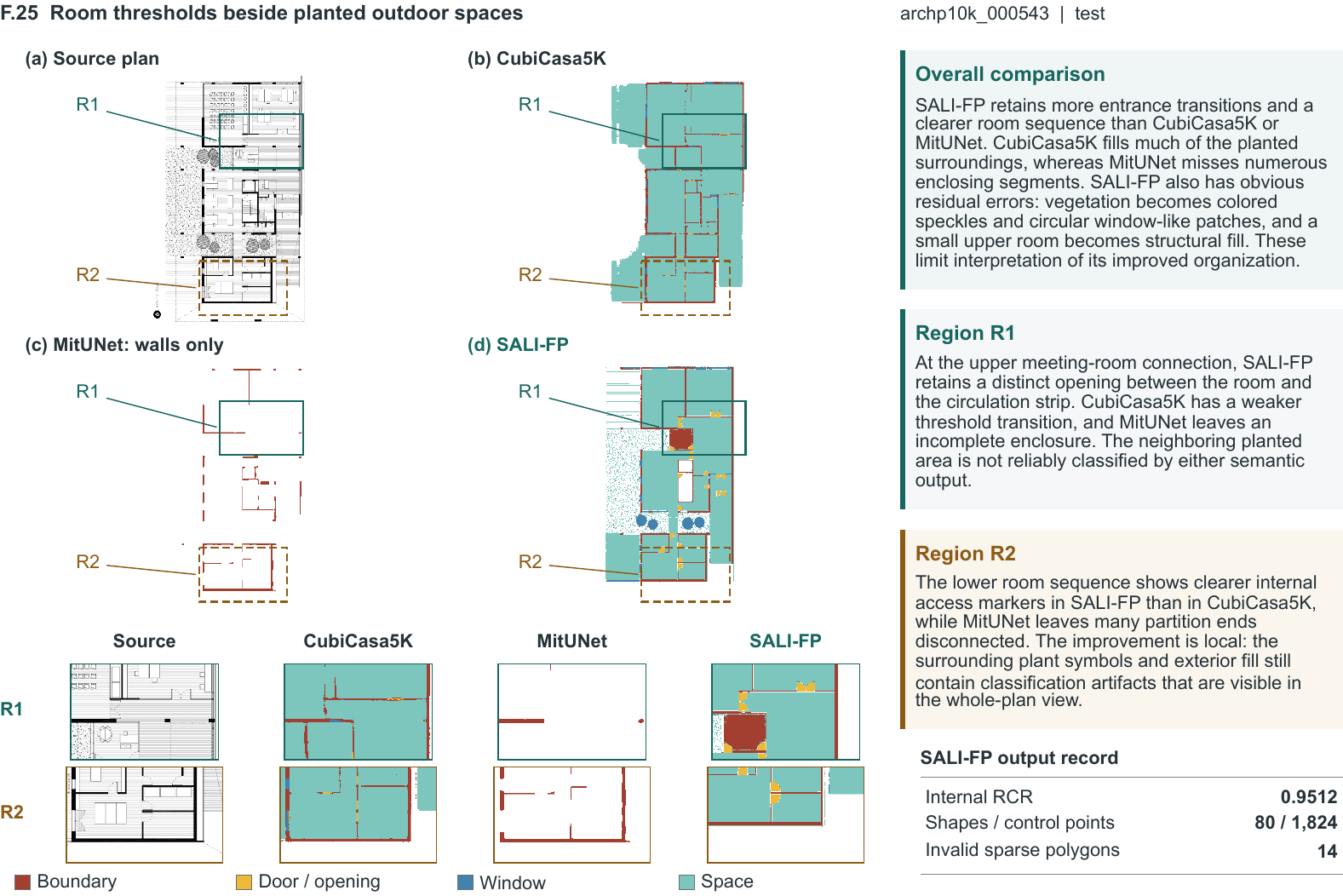}
\captionof{figure}{archp10k\_000543: room thresholds beside planted outdoor spaces.}\label{fig:F.25}
\end{landscape}
\restoregeometry

\clearpage
\newgeometry{margin=10mm}
\begin{landscape}
\thispagestyle{empty}
\centering
\includegraphics[width=267mm,height=178mm,keepaspectratio]{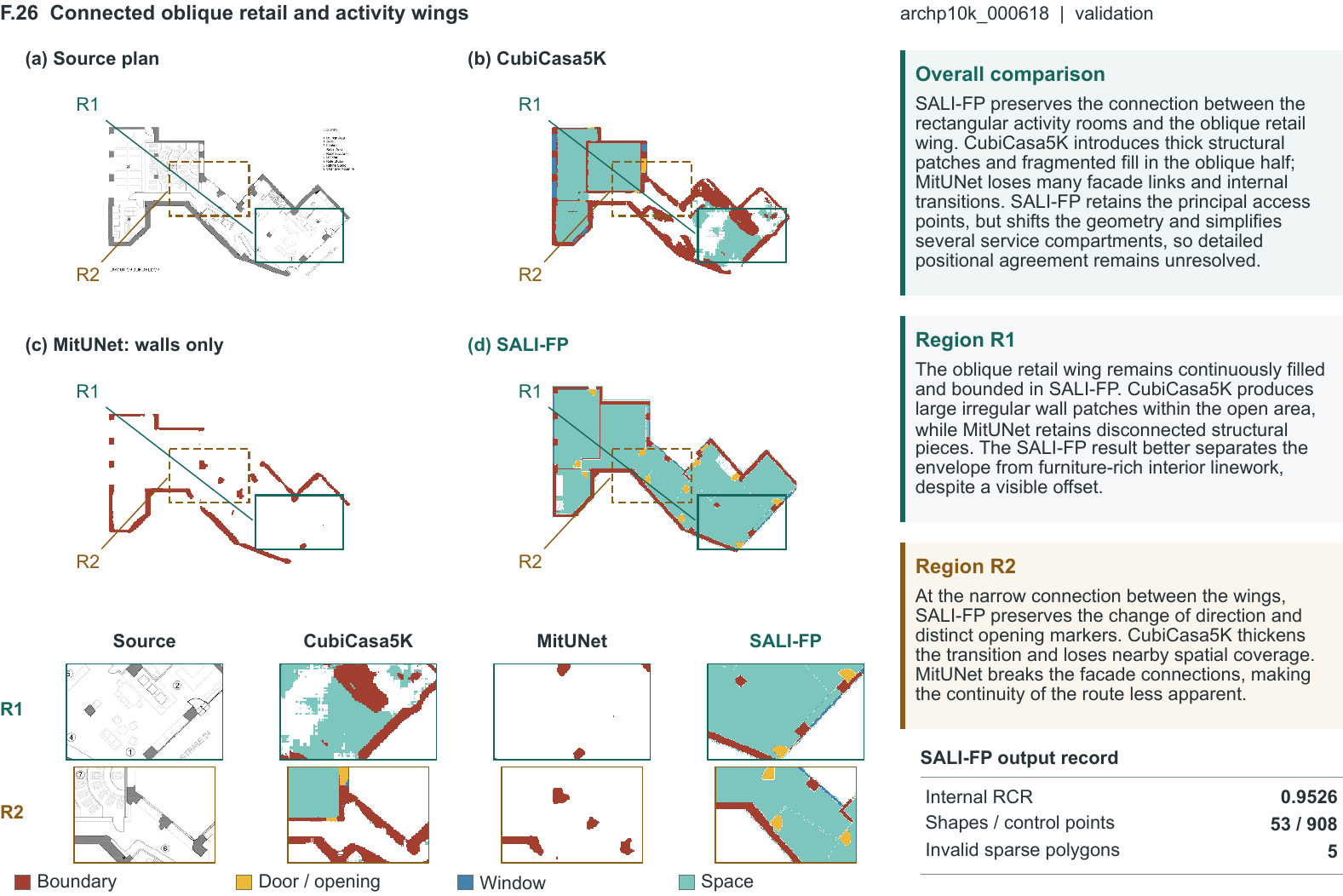}
\captionof{figure}{archp10k\_000618: connected oblique retail and activity wings.}\label{fig:F.26}
\end{landscape}
\restoregeometry

\clearpage
\newgeometry{margin=10mm}
\begin{landscape}
\thispagestyle{empty}
\centering
\includegraphics[width=267mm,height=178mm,keepaspectratio]{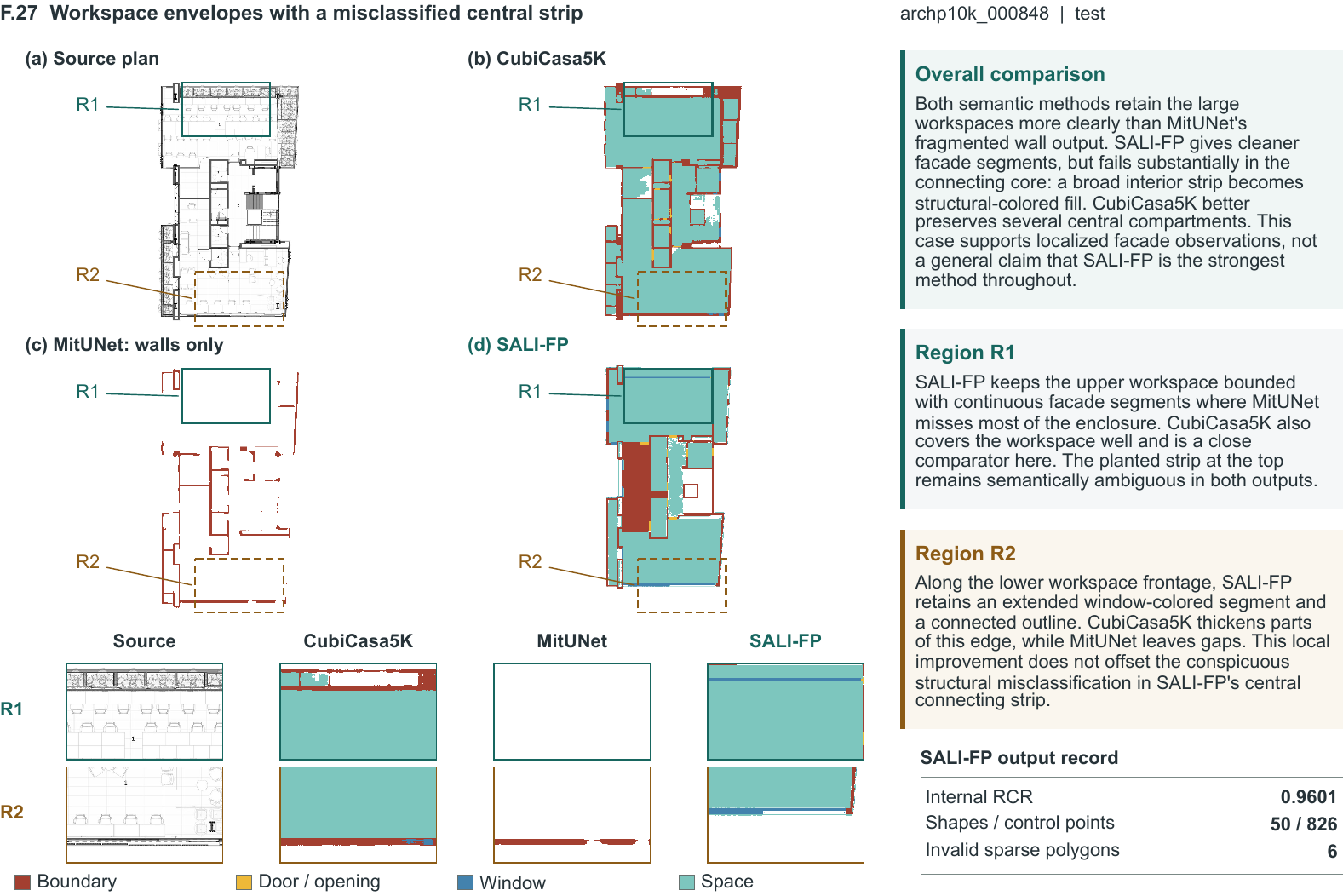}
\captionof{figure}{archp10k\_000848: workspace envelopes with a misclassified central strip.}\label{fig:F.27}
\end{landscape}
\restoregeometry

\clearpage
\newgeometry{margin=10mm}
\begin{landscape}
\thispagestyle{empty}
\centering
\includegraphics[width=267mm,height=178mm,keepaspectratio]{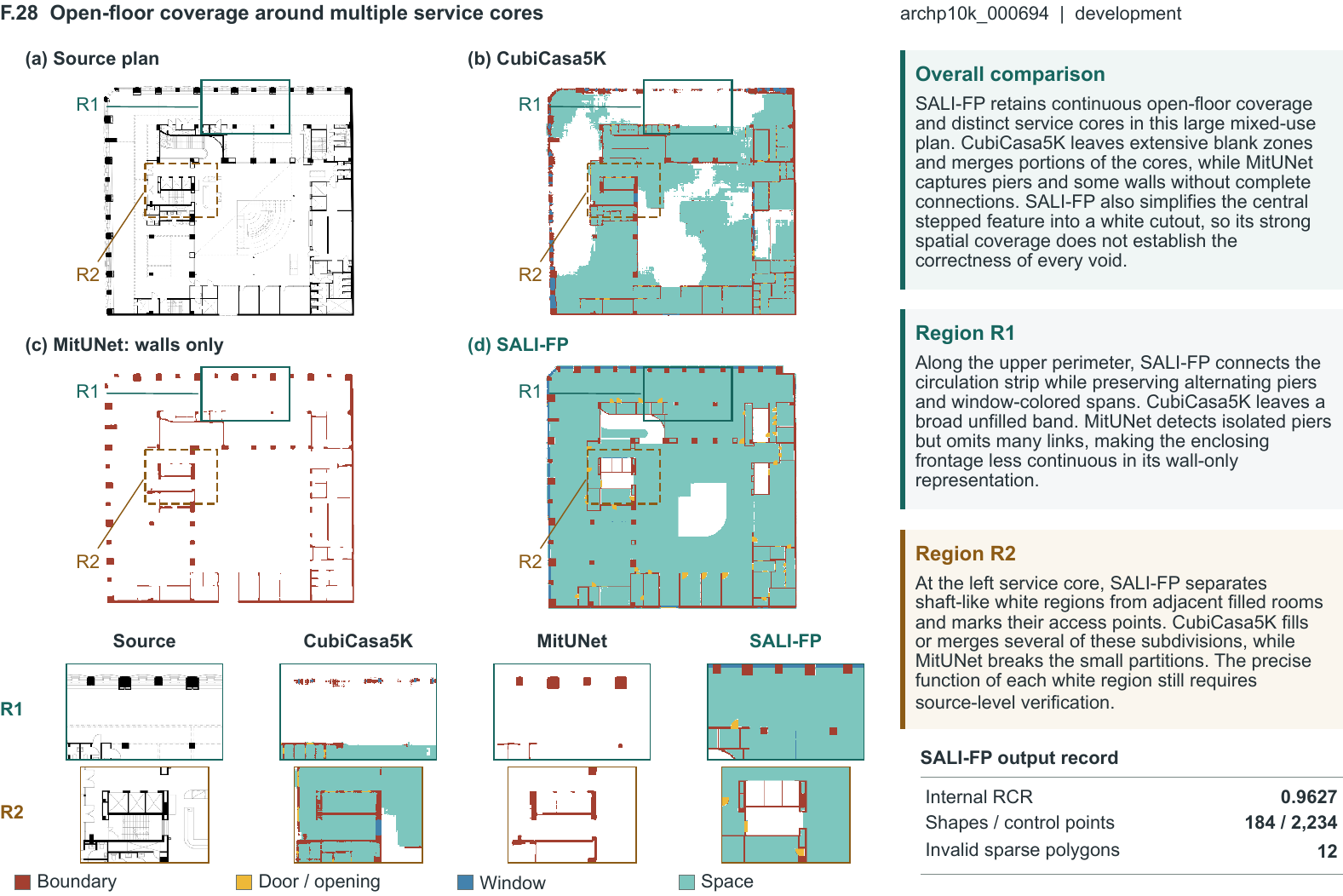}
\captionof{figure}{archp10k\_000694: open-floor coverage around multiple service cores.}\label{fig:F.28}
\end{landscape}
\restoregeometry

\clearpage
\newgeometry{margin=10mm}
\begin{landscape}
\thispagestyle{empty}
\centering
\includegraphics[width=267mm,height=178mm,keepaspectratio]{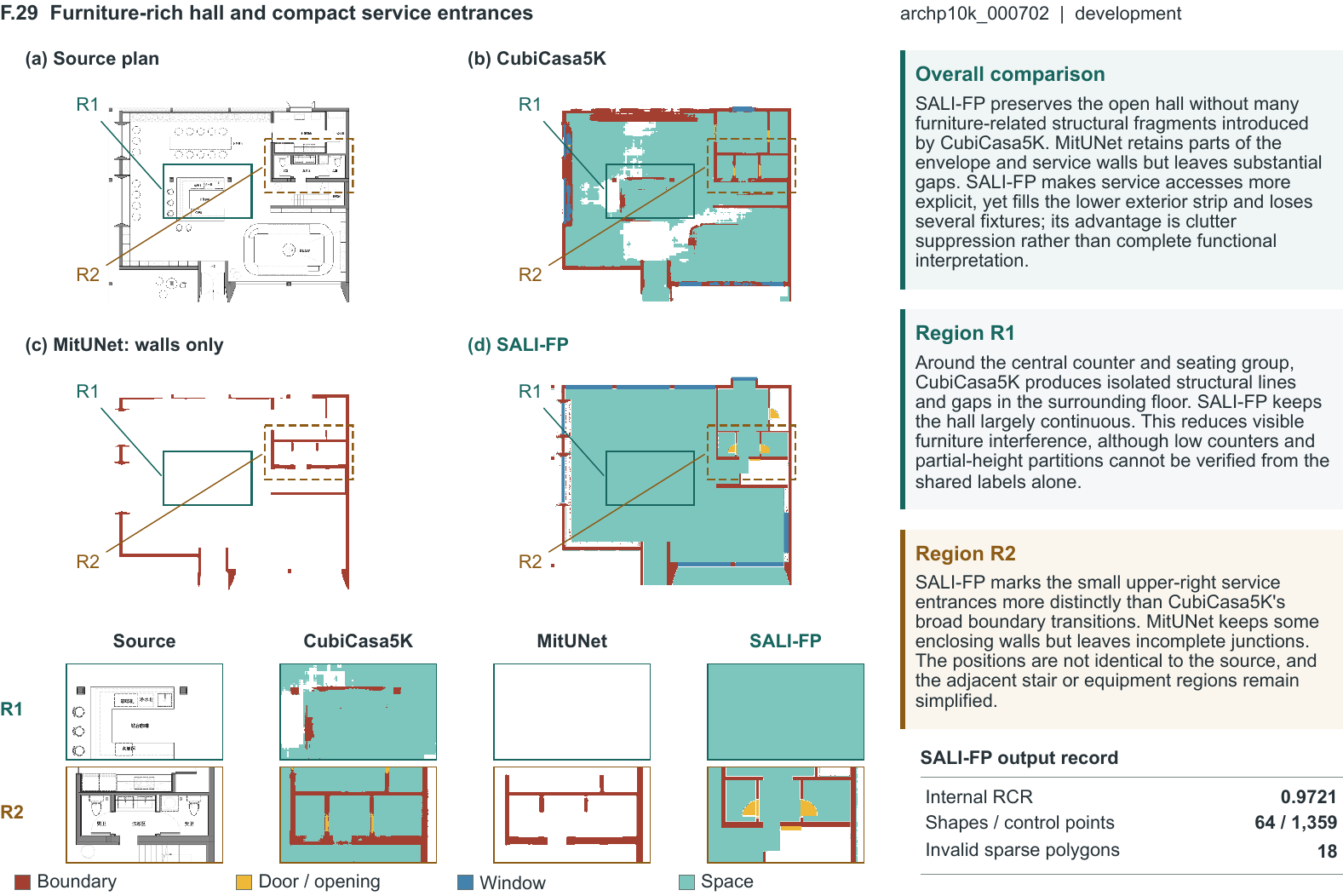}
\captionof{figure}{archp10k\_000702: furniture-rich hall and compact service entrances.}\label{fig:F.29}
\end{landscape}
\restoregeometry

\clearpage
\newgeometry{margin=10mm}
\begin{landscape}
\thispagestyle{empty}
\centering
\includegraphics[width=267mm,height=178mm,keepaspectratio]{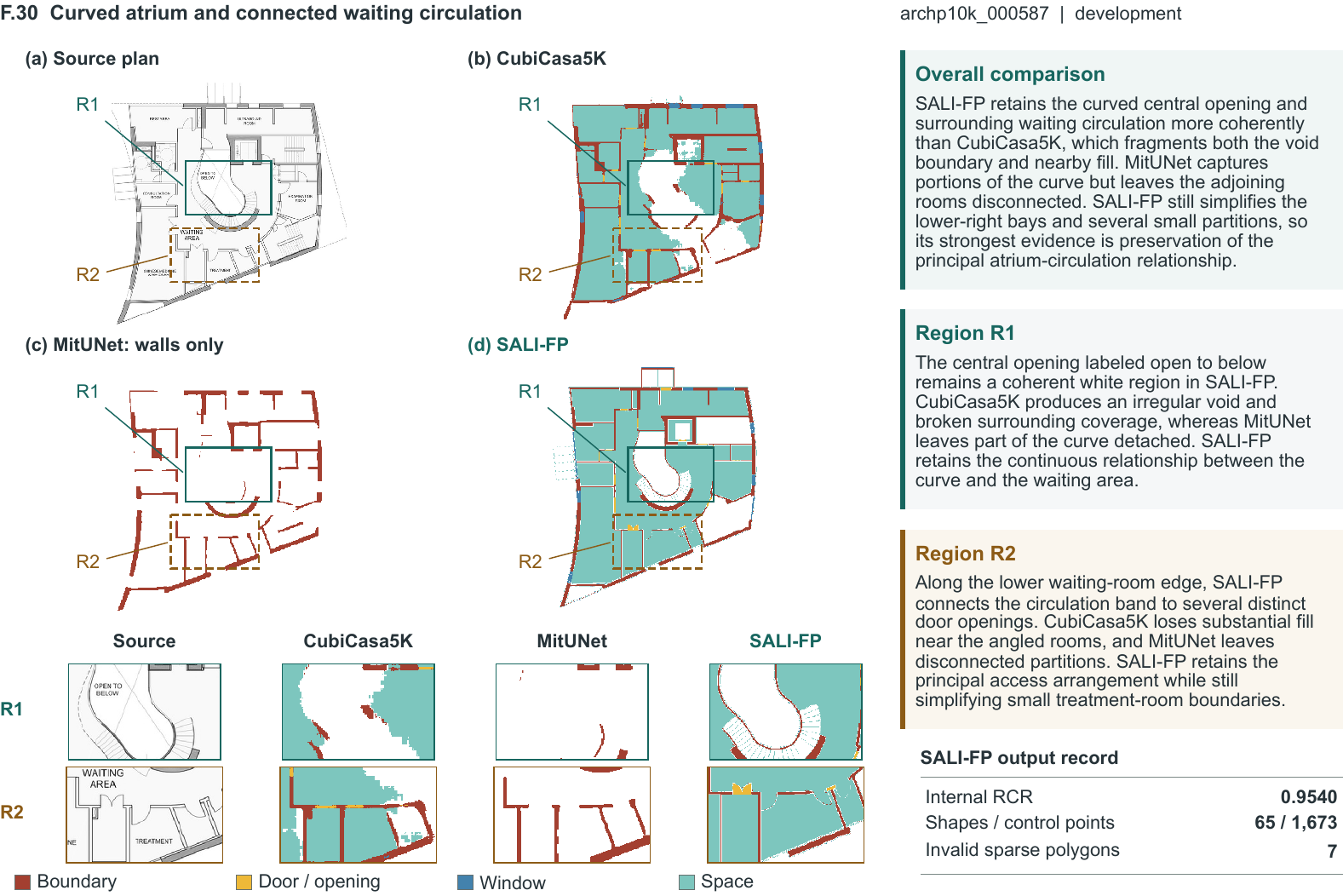}
\captionof{figure}{archp10k\_000587: curved atrium and connected waiting circulation.}\label{fig:F.30}
\end{landscape}
\restoregeometry

\addtocounter{page}{-1}
\end{document}

%% file: highlights.tex
\par\noindent\begin{minipage}{\linewidth}\vspace{7pt}
\noindent\textbf{Highlights}
\begin{itemize}\setlength{\itemsep}{1pt}\setlength{\parskip}{0pt}
  \item Evidence-gated multimodal parsing links floor-plan interpretation to geometry.
  \item Local evidence gates constrain semantic edits and retain recoverable states.
  \item Structured outputs are recovered for 11,534 heterogeneous architectural plans.
  \item Stage comparisons separate semantic revision from coordinate effects.
\end{itemize}
\end{minipage}\par